\documentclass[11pt]{article}

\usepackage{fullpage}
\usepackage[margin = 2.5cm]{geometry}

\usepackage{authblk}

\usepackage[round]{natbib}

\usepackage[ruled,vlined]{algorithm2e}

\usepackage{amsmath, amssymb, amsthm,bbm}
\usepackage{algorithmic}
\usepackage{graphicx}
\usepackage{color}
\usepackage[dvipsnames]{xcolor}
\usepackage{subcaption}
\usepackage{float}

\usepackage[hyphens]{url}
\usepackage[colorlinks=true, breaklinks]{hyperref}

\usepackage{nicefrac}

\usepackage{graphicx} 
\usepackage[font=footnotesize,labelfont=bf]{caption}

\usepackage{booktabs}
\newcommand{\opt}{\operatorname{OPT}}
\newcommand{\TV}{{\sf TV}}

\newcommand{\R}{\mathbb{R}}

\newcommand{\calC}{\mathcal{C}}

\newcommand{\calX}{\mathcal{X}}
\newcommand{\calY}{\mathcal{Y}}
\newcommand{\calZ}{\mathcal{Z}}

\newcommand{\vol}{\mathrm{vol}}

\newcommand{\eat}[1]{}

\newcommand{\Esymb}{\mathbb{E}}
\newcommand{\Psymb}{\mathbb{P}}

\DeclareMathOperator*{\E}{\Esymb}
 
 \DeclareMathOperator*{\ProbOp}{\Psymb}

 \DeclareMathOperator*{\argmin}{argmin}
 \DeclareMathOperator*{\argmax}{argmax}

\renewcommand{\Pr}{\ProbOp}

\renewcommand{\epsilon}{\varepsilon}

\newif\ifnotes\notestrue

\ifnotes
\usepackage{color}
\definecolor{mygrey}{gray}{0.50}
\newcommand{\notename}[2]{{\textcolor{blue}{\footnotesize{\bf (#1:} {#2}{\bf ) }}}}

\newcommand{\vnote}[1]{{\notename{Vaidehi}{#1}}}

\newcommand{\anote}[1]{{\notename{Aravindan}{#1}}}
\else

\newcommand{\notename}[2]{{}}

\newcommand{\enote}[1]{}
\newcommand{\vnote}[1]{}
\newcommand{\bnote}[1]{}
\newcommand{\anote}[1]{}

\fi

\newtheorem{theorem}{Theorem}[section]
\newtheorem*{theorem*}{Theorem}

\newtheorem{proposition}[theorem]{Proposition}
\newtheorem*{proposition*}{Proposition}
\newtheorem{lemma}[theorem]{Lemma}
\newtheorem*{lemma*}{Lemma}

\newtheorem*{conjecture*}{Conjecture}

\newtheorem*{fact*}{Fact}

\newtheorem*{hypothesis*}{Hypothesis}

\newtheorem*{claim*}{Claim}

\theoremstyle{definition}

\newtheorem{assumption}[theorem]{Assumption}

\theoremstyle{remark}
\newtheorem{remark}[theorem]{Remark}
\newtheorem*{remark*}{Remark}

\renewcommand{\and}{\hspace{1em}} 

\title{
Volume Optimality in Conformal Prediction with Structured Prediction Sets }

\author{    
    Chao Gao\thanks{Department of Statistics, University of Chicago, Chicago, USA} \and
    Liren Shan\thanks{Toyota Technological Institute at Chicago, Chicago, USA} \and Vaidehi Srinivas\thanks{Department of Computer Science, Northwestern University, Evanston, USA} \and Aravindan Vijayaraghavan\thanks{Department of Computer Science, Northwestern University, Evanston, USA}
    }
\date{}

\begin{document}

\maketitle

\begin{abstract}

Conformal Prediction is a widely studied technique to construct prediction sets of future observations. Most conformal prediction methods focus on achieving the necessary coverage guarantees, but do not provide formal guarantees on the size (volume) of the prediction sets. We first prove an impossibility of volume optimality where any distribution-free method can only find a trivial solution. We then introduce a new notion of volume optimality by restricting the prediction sets to belong to a set family (of finite VC-dimension), specifically a union of $k$-intervals. Our main contribution is an efficient distribution-free algorithm based on dynamic programming (DP) to find a union of $k$-intervals that is guaranteed for any distribution to have near-optimal volume among all unions of $k$-intervals satisfying the desired coverage property. 
By adopting the framework of distributional conformal prediction \citep{chernozhukov2021distributional}, the new DP based conformity score can also be applied to achieve approximate conditional coverage and conditional restricted volume optimality, as long as a reasonable estimator of the conditional CDF is available. 
While the theoretical results already establish volume-optimality guarantees, they are complemented by experiments that demonstrate that our method can significantly outperform existing methods in many settings.

\end{abstract}

\newpage

\tableofcontents

\newpage

\section{Introduction}

Conformal inference has emerged as a powerful black-box method for quantifying uncertainty in model predictions, providing confidence sets or prediction sets that contain the true value with a specified probability
~\citep{gammerman1998learning,Vovkbook}. Consider a prediction problem where $\mathcal{X}$ is the covariate space (feature space), and $\mathcal{Y}$ is the label space.  Given a dataset of $n$ labeled samples $(X_1, Y_1), \dots, (X_n, Y_n) \in \mathcal{X} \times \mathcal{Y}$, a conformal prediction algorithm uses these $n$ samples (often called calibration samples) to construct for a test $X_{n+1} \in \mathcal{X}$ with (unknown) true value $Y_{n+1} \in \mathcal{Y}$, a prediction set that we will denote by $\widehat{C}(X_{n+1}) \subset \mathcal{Y}$,\footnote{It may be more accurate to use $\widehat{C}(X_1, Y_1, \dots, X_n,Y_n, X_{n+1})$ instead of $\widehat{C}(X_{n+1})$ to reflect that $\widehat{C}$ is a function of the calibration samples and $X_{n+1}$.} 
satisfying the {\em coverage} requirement for some desired parameter $\alpha \in (0,1)$:
\begin{equation}\label{eq:intro:coverage}
 \mathbb{P}\left(Y_{n+1} \in \widehat{C}(X_{n+1})\right)\geq 1-\alpha.
\end{equation}

 Here the probability $\mathbb{P}$ refers to the joint distribution over all $n+1$ pairs of observations $(X_1, Y_1), \dots, (X_n, Y_n), (X_{n+1}, Y_{n+1})$ including the test sample $(X_{n+1}, Y_{n+1})$. 
 Unlike traditional approaches, conformal inference is distribution-free, relying only on the assumption of exchangeability of the joint distribution $\mathbb{P}$  over the $(n+1)$ samples. 

While most conformal methods provide guarantees on coverage, they do not provide any control on the size or volume of the prediction sets; in fact, the trivial choice of $\widehat{C}(X_{n+1})=\mathcal{Y}$ also satisfies the coverage requirement. Consequently, the size of these sets is often validated empirically, without formal guarantees. 
This raises the important question of {\em volume optimality}, which is the focus of this paper:\\

\vspace{-15pt}
\noindent {\bf Question:} {\em Given calibration samples $(X_1, Y_1),\dots,$ $(X_n, Y_n)$ drawn i.i.d. from a distribution $P$, can we find among all data-dependent sets $\widehat{C} \subset \mathcal{Y}$ satisfying the desired coverage requirement for $(X_{n+1}, Y_{n+1}) \sim P$, the one with the smallest volume, as quantified by the Lebesgue measure
$\vol(\widehat{C})=\lambda(\widehat{C})?$}

The volume of the prediction set in conformal prediction is also sometimes referred to as `efficiency' has been stated as an important consideration in many prior works \citep[see e.g.,][]{shafer2008tutorial, angelopoulos2023survey}. However, most works that we are aware of do not give theoretical guarantees of volume optimality, and mainly reason about volume control through empirical evaluations.

There are few works that provide guarantees of volume optimality.\footnote{Some works also guarantee that the coverage is not much more than $1-\alpha$, e.g., $\mathbb{P}\left(Y_{n+1}\in \widehat{C}(X_{n+1}) \mid X_{n+1}\right) \le  1-\alpha + o(1)$ to argue that the prediction set is not too big. However, smallness according to the measure $\mathbb{P}$ does not necessarily reflect a small volume (or Lebesgue measure) for the set. 
} Notable exceptions include \citet{Lei2013DistributionFreePS} in the unsupervised setting,  \citet{Sadinle2016LeastAS} in the set value setting, and recent works of \citet*{izbicki2020flexible,izbicki2022cd} and \citet{kiyani2024length} in the regression setting. As summarized by \cite{angelopoulos2024theoretical}, a sufficient condition that leads to volume optimality of conformal prediction is consistent estimation of the conditional density function of $Y$ given $X$. This is essentially the strategy adopted by previous work \citep{Lei2013DistributionFreePS,izbicki2020flexible,izbicki2022cd}. In comparison, our method, by incorporating a framework of \cite{chernozhukov2021distributional}, builds on the estimation of the conditional CDF via a new conformity score computed by dynamic programming, and thus also works in settings where good conditional density estimation is impossible or density does not even exist.



\subsection{Our Results}

\noindent {\bf An Impossibility Result.} We first prove an impossibility result in a one-dimensional setting where any distribution-free method that satisfies the coverage requirement can only find a trivial solution whose volume is sub-optimal. See Theorem~\ref{thm:impossibility} for a formal statement. This result provides an explanation for the lack of such volume-optimality guarantees in the conformal prediction literature, and also motivates our new notion of volume-optimality that we introduce in this work. 


\paragraph{Structured Prediction Sets and Restricted Volume Optimality.}


Motivated by the impossibility result, our goal is to find a prediction set $\widehat{C} \in \mathcal{C}\subset 2^{\mathcal{Y}}$ whose volume is competitive with the optimum volume of any set in the family $\mathcal{C}$ as given by
\begin{equation} \label{eq:intro:optC}
\opt_{\mathcal{C}}(P, 1-\alpha) = \inf_{C \in \mathcal{C}} \left\{\vol(C) : P(C) \geq 1-\alpha \right\}. 
\end{equation}
As long as $\mathcal{C}$ has bounded VC-dimension, for any distribution $P$ we can obtain good empirical estimates of the probability measure of any set $C \in \mathcal{C}$ via a standard uniform concentration inequality, which allows us to overcome the impossibility result in Theorem~\ref{thm:impossibility}.
In the rest of the paper, we focus on the setting when $\mathcal{Y}=\mathbb{R}$ and $\mathcal{C}= \mathcal{C}_k$ which is the collection of unions of $k$ intervals.






\paragraph{Conformalized Dynamic Programming.}

Equipped with our new notion of volume optimality, we propose a new conformity score based on dynamic programming, the proposed method is shown to not only achieve approximate conditional coverage as in \citep{chernozhukov2021distributional} and \citep{romano2019conformalized}, but also conditional volume optimality with respect to unions of $k$ intervals, as long as a reasonable estimator of the conditional CDF is available. Our method of learning a predictive set via CDF can be regarded an extension of the framework of \cite{izbicki2020flexible,chernozhukov2021distributional}.

\subsection{Paper Organization}

We will start with the unsupervised setting with label-only data in Section \ref{sec:unlabeled}. The extension of the theory and algorithm to the supervised setting is given in Section \ref{sec:labeled}. The numerical comparisons between our proposed methodology and existing methods in the literature are presented in Section \ref{sec:numerical_experiments}. All technical proofs and additional numerical experiments will be presented in the appendix.

\section{Unsupervised Setting} \label{sec:unlabeled}

\subsection{Approximate Volume Optimality}

Suppose $Y_1,\cdots,Y_n,Y_{n+1}$ are independently drawn from a distribution $P$ on $\mathbb{R}$. The goal is to predict $Y_{n+1}$ based on the first $n$ samples $Y_1,\cdots,Y_n$. To be specific, we would like to construct a data-dependent set $\widehat{C}=\widehat{C}(Y_1,\cdots,Y_n)$ such that 
\begin{equation}
\mathbb{P}(Y_{n+1}\in \widehat{C})\geq 1-\alpha. \label{eq:un-cov}
\end{equation}
Among all data-dependent set that satisfies (\ref{eq:un-cov}), our goal is to find the one with the smallest volume, quantified by the Lebesgue measure $\vol(\widehat{C})=\lambda(\widehat{C})$.
When the distribution $P$ is known, one can directly minimize $\lambda(C)$, subject to $P(C)\geq 1-\alpha$ without even using the data. In particular, when $P\ll \lambda$, an optimal solution is given by the density level set
$$C_{\rm opt}=\left\{\frac{dP}{d\lambda}>t\right\}\cup D,$$
for some $t>0$ and $D$ is some subset of $\{dP/d\lambda =t\}$.

In general, $P$ may not be absolutely continuous and the density need not exist. Nonetheless, we can still define the optimal volume by
$$\opt(P,1-\alpha)=\inf\{\vol(C):P(C)\geq 1-\alpha\}.$$
Note that without any assumption on $P$, the above optimization problem may not have a unique solution. Moreover, it is  possible that the infimum cannot be achieved by any measurable set. Therefore, a natural relaxation is to consider approximate volume optimality. For some $\epsilon\in (0,\alpha)$, a prediction set $\widehat{C}$ is called $\epsilon$-optimal if
\begin{equation}
\vol(\widehat{C})\leq \opt(P,1-\alpha+\epsilon), \label{eq:un-vol}
\end{equation}
either in expectation or with high probability.

The notion of volume optimality defined by (\ref{eq:un-vol}) is quite different from those considered in the literature. A popular quantity that has already been studied is the volume of set difference $\vol\left(\widehat{C}\Delta C_{\rm opt}\right)$ \citep{Lei2013DistributionFreePS,izbicki2020flexible,chernozhukov2021distributional}. However, this much stronger notion requires that the optimal solution $C_{\rm opt}$ must not only exist but also be unique. Usually additional assumptions need to be imposed in the neighborhood of the boundary of $C_{\rm opt}$ in order that the set difference vanishes in the large sample limit. In comparison, the definition (\ref{eq:un-vol}) only requires the volume to be controlled, which can be achieved even if $\widehat{C}$ is not close to $C_{\rm opt}$, or when $C_{\rm opt}$ does not even exist. Indeed, from a practical point of view, any set with coverage and volume control would serve the purpose of valid prediction. Insisting the closeness to a questionable target $C_{\rm opt}$ comes at the cost of unnecessary assumptions on the data generating process.

Another notion considered in the literature is close to our formulation (\ref{eq:un-vol}). Instead of relaxing the coverage probability level from $1-\alpha$ to $1-\alpha+\epsilon$, one can consider the following approximate volume optimality,
\begin{equation}
\vol(\widehat{C})\leq \opt(P,1-\alpha) + \epsilon. \label{eq:additive}
\end{equation}
Results of interval length optimality in the sense of (\ref{eq:additive}) have been studied by \citep{chernozhukov2021distributional,kiyani2024length}. However, the $\epsilon$ in (\ref{eq:additive}) is usually proportional to the scale of the distribution $P$, or may depend on $P$ in some other ways. In comparison, the $\epsilon$ in (\ref{eq:un-vol}) has the unit of probability, and as we will show later, can be made independent of the distribution $P$, which leads to more natural and cleaner theoretical results with fewer assumptions.

\subsection{Impossibility of Distribution-Free Volume Optimality}

It is known that conformal prediction achieves the coverage property (\ref{eq:un-cov}) in a distribution-free sense, meaning that (\ref{eq:un-cov}) holds uniformly for all distributions $P$. One naturally hopes that the approximate volume optimality (\ref{eq:un-vol}) can also be established in a distribution-free way. Perhaps not surprisingly, this goal is too ambitious. The theorem below rigorously proves the impossibility of the task.

\begin{theorem}\label{thm:impossibility}
Consider observations $Y_1$, $Y_2$, $\dots$, $Y_n$, $Y_{n+1}$ sampled $i.i.d.$ from a distribution $P$ on $\R$. 
Suppose $\widehat{C}=\widehat{C}(Y_1,\cdots,Y_n)$ satisfies $\mathbb{P}(Y_{n+1}\in \widehat{C})\geq 1-\alpha$
for all distribution $P$. 
Then, for any $\epsilon \in (0,\alpha)$, there exists some distribution $P$ on $\R$, such that the expected volume of the prediction set is at least
$$\E\vol(\widehat{C})\geq \opt(P,1-\alpha+\epsilon).$$
\end{theorem}

The above impossibility result can be regarded as a consequence of a nonparametric testing lower bound. Consider the following hypothesis testing problem,
\begin{eqnarray*}
H_0:&& P=P_0 \\
H_1:&& P\in\left\{P:\TV(P,P_0)>1-\delta\right\}.
\end{eqnarray*}
It is well known that a testing procedure with both vanishing Type-1 and Type-2 errors does not exist without further constraining the alternative hypothesis, even when $\delta$ is arbitrarily close to $0$ \citep{lecam1960necessary,barron1989uniformly}. In the setting of distribution-free inference with simultaneous coverage and volume guarantees, the coverage property involves the measure $P^{n+1}$, while the expected volume is defined by another measure $P^n\otimes\lambda$. When restricting the support of $P$ to the unit interval $[0,1]$, $\lambda$ becomes the uniform probability, and thus both $P^{n+1}$ and $P^n\otimes\lambda$ are probability distributions. It turns out achieving approximate volume optimality is related to hypothesis testing between $P^{n+1}$ and $P^n\otimes\lambda$ with total variation separation.

\subsection{Distribution-Free Restricted Volume Optimality}

The impossibility result implies a volume lower bound $\opt(P,1-\alpha+\epsilon)$, where the coverage level $1-\alpha+\epsilon$ can be arbitrarily close to $1$. This means that, at least in the worst case, the volume cannot be smaller than that of the support of $P$.

To avoid this triviality, in this section, we consider a weaker notion of volume optimality by only considering prediction sets that are unions of $k$ intervals. We use $\mathcal{C}_k$ to denote the collection of all sets that are unions of $k$ intervals. The restricted optimal volume with respect to the class $\mathcal{C}_k$ is defined by
\begin{equation}
\opt_k(P, 1-\alpha) = \inf_{C \in \mathcal{C}_k} \left\{\vol(C) : P(C) \geq 1-\alpha \right\}. \label{eq:res-opt}
\end{equation}

\begin{remark}\label{rem:noloss:smooth}
We remark that we are still in a distribution-free setting, since no assumption is imposed on $P$. Instead, the restriction only constrains the shape of the prediction set. 
From a practical point of view, it is reasonable to require that
$\widehat{C}\in \mathcal{C}_k$,
since a more complicated prediction set would be hard to interpret. Moreover, as long as $P$ admits a density function with at most $k$ modes, the two notions match,
$$\opt_k(P,1-\alpha)=\opt(P,1-\alpha).$$
More generally, it can be shown that
$$\opt_k(P,1-\alpha)\leq\opt(P,1-\alpha+\epsilon),$$
for some $\epsilon\in (0,\alpha)$, whenever $P$ can be approximated by a distribution with at most $k$ modes. This, in particular, includes the situation where the density of $P$ can be well estimated by a kernel density estimator. A rigorous statement will be given in Appendix \ref{sec:DPvsKDE}. 
\end{remark}

Given the observations $Y_1,\cdots,Y_n$, we define the empirical distribution $\mathbb{P}_n=\frac{1}{n}\sum_{i=1}^n\delta_{Y_i}$. To achieve restricted volume optimality, one can use
\begin{equation}
\widehat{C}=\argmin_{C \in \calC_k}\left\{\vol(C): \mathbb{P}_n(C)\geq 1-\alpha \right\}. \label{eq:erm}
\end{equation}
According to its definition, the prediction set (\ref{eq:erm}) satisfies both $\mathbb{P}_n(\widehat{C})\geq 1-\alpha$ and $\vol(\widehat{C})=\opt_k(\mathbb{P}_n,1-\alpha)$. The coverage and volume guarantees under $P$ can be obtained via
\begin{equation}
\sup_{C \in \calC_k}|\mathbb{P}_n(C)-P(C)| = O_P\left(\sqrt{{\rm VC}(\calC)/n}\right),\label{eq:un-con}
\end{equation}
with ${\rm VC}(\calC)=O(k)$. Therefore, approximate optimality can be achieved by (\ref{eq:erm}) whenever (\ref{eq:un-con}) holds.

A naive exhaustive search to find (\ref{eq:erm}) requires exponential computational time. We show that an efficient dynamic programming algorithm (Algorithm \ref{alg:dp}) can solve (\ref{eq:erm}) approximately with some additional slack $\gamma$, which determines the computational complexity. Theoretical guarantees of Algorithm \ref{alg:dp} are given in the following proposition.

\begin{algorithm}[tb]
\caption{Dynamic Programming Solving (\ref{eq:erm})}\label{alg:dp}
\begin{algorithmic}
\STATE {\bfseries Input:} data points $Y_1, \dots, Y_n \in \R$, coverage level $1-\alpha \in (0,1)$ and slack $\gamma \in (0,\alpha)$, number of intervals $k$ 
\STATE {\bfseries Output:} $k$ intervals that cover $\lceil (1-\alpha) n \rceil$ points with minimum volume
\STATE Sort data in non-decreasing order $Y_{(1)} \leq \dots \leq Y_{(n)}$\;
\STATE Set $DP(i,j,l) = \infty$ for all $i \in [k], j \in [n], l \in [1/\gamma]$\;
\FOR{$i = 1$ {\bfseries to} $k$, $j = 1$ {\bfseries to} $n$, $l = 1$ {\bfseries to} $\lceil 1/\gamma \rceil$}
    \FOR{$j' = i$ {\bfseries to} $j$}
        \FOR{$j'' = i-1$ {\bfseries to} $j'-1$}
            \STATE Set $l' = l - \lfloor (j - j' + 1)/(\gamma n) \rfloor$\;
            \IF{$l' < 0$ \textbf{and}  $i=1$}
                \STATE $DP(i,j,l) = \min\{DP(i,j,l), Y_{(j)} - Y_{(j')}\}$\;
            \ENDIF
            \IF{$DP(i-1, j'', l') \neq \infty$}
                \STATE $DP(i,j,l) = \min\{DP(i,j,l), Y_{(j)} - Y_{(j')} + DP(i-1, j'', l')\}$\;
            \ENDIF
        \ENDFOR
    \ENDFOR
\ENDFOR
\STATE Return the solution with minimum volume among all $DP(k,j, \lceil (1-\alpha)/\gamma \rceil )$ for $1 \leq j \leq n$\;
\end{algorithmic}
\end{algorithm}

\begin{proposition}\label{thm:DP}
For any $\gamma\in(0,\alpha)$, Algorithm \ref{alg:dp} computes a prediction set $\widehat{C}_{\rm DP}\in \calC_k$ by dynamic programming with time complexity $O(n^3k/\gamma)$ such that 
    \begin{enumerate}
        \item $\mathbb{P}_n(\widehat{C}_{\rm DP})\geq 1-\alpha$;
        \item $\vol(\widehat{C}_{\rm DP})\leq \opt_k(\mathbb{P}_n,1-\alpha+\gamma)$.
    \end{enumerate}
\end{proposition}

Together with (\ref{eq:un-con}), the coverage and volume guarantees of the dynamic programming can also be generalized from $\mathbb{P}_n$ to $P$.

\subsection{Conformalizing Dynamic Programming}\label{sec:cdp}

Having understood the generalization ability of dynamic programming, we are ready to conformalize the procedure to achieve a finite-sample coverage property. For simplicity, we will adopt the framework of split conformal prediction, though in principle full conformal prediction can also be applied here.

In the split conformal predicition framework, the data set is split into two halves. The first half is used to compute a conformity score, and the second half determines the quantile level. For convenience of notation, let us assume, from now on, that the sample size is $2n$. The split conformal procedure is outlined below.
\begin{enumerate}
\item Compute a score function $q(\cdot)$ using $Y_1,\cdots,Y_n$.
\item Evaluate $q(Y_{n+1}),\cdots, q(Y_{2n})$, and order them as $q_1\leq \cdots\leq q_n$.
\item Output the prediction set
\begin{equation}
\widehat{C}=\left\{y: q(y)\geq q_{\lfloor(n+1)\alpha\rfloor}\right\}. \label{eq:pred-set-un}
\end{equation}
\end{enumerate}
By the exchangeability of $Y_1,\cdots,Y_{2n},Y_{2n+1}$, the prediction set $\widehat{C}$ satisfies
$$\mathbb{P}\left(Y_{2n+1}\in \widehat{C}\right)\geq 1-\alpha,$$
where the above probability is over the randomness of $(Y_1,\cdots,Y_n)$, that of $(Y_{n+1},\cdots,Y_{2n})$, and that of $Y_{2n+1}$.

To conformalize the dynamic programming that approximately computes (\ref{eq:erm}), we will first compute a nested system $S_1\subset\cdots\subset S_m\subset \mathbb{R}$ using the data $Y_1,\cdots,Y_n$. The nested system is required to satisfy the following assumption.

\begin{assumption}\label{as:ne-un}
The sets $S_1\subset\cdots\subset S_m\subset\mathbb{R}$ are measurable with respect to the $\sigma$-field generated by $Y_1,\cdots,Y_n$. Moreover, for some positive integer $k$, some $\alpha\in(0,1)$ and some $\delta,\gamma$ such that $3\delta+\gamma+n^{-1}\leq \alpha$, we have
\begin{enumerate}
\item $\mathbb{P}_n(S_j)=\frac{j}{m}$ and $S_j\in\calC_k$ for all $j\in[m]$.
\item There exists some $j^*\in[m]$, such that\\ $\mathbb{P}_n(S_{j^*})\geq 1-\alpha+n^{-1}+3\delta$ and \\$\vol(S_{j^*})\leq \opt_k(\mathbb{P}_n,1-\alpha+\frac{1}{n}+3\delta+\gamma)$.
\end{enumerate}
Here, $\mathbb{P}_n$ denotes the empirical distribution $\frac{1}{n}\sum_{i=1}^n\delta_{Y_i}$ of the first half of the data.
\end{assumption}
To construct a nested system $\{S_j\}_{j\in[m]}$ that satisfies the above assumption, one only needs to make sure that there exists one subset $S_{j^*}$ in the system that is computed by the dynamic programming (Algorithm \ref{alg:dp}) with confidence level $1-\alpha+n^{-1}+3\delta$ and slack parameter $\gamma$. The rest of the sets in the system can be constructed just to satisfy $\mathbb{P}_n(S_j)=\frac{j}{m}$. In Section \ref{sec:con-nest-m}, we will present a greedy expansion/contraction algorithm that satisfies Assumption \ref{as:ne-un}.

With a nested system $\{S_j\}_{j\in[m]}$ satisfying Assumption \ref{as:ne-un}, we can define the conformity score as
\begin{equation}
q(y) = \sum_{j=1}^m\mathbb{I}\{y\in S_j\}. \label{eq:score-un}
\end{equation}
The equivalence between nested system and conformity score was advocated by \cite{gupta2022nested}.
Intuitively, $q(y)$ quantifies the depth of the location $y$. A higher score implies that $y$ is covered by more sets in the nested system, and thus the location should be more likely to be included in the prediction set. Applying the standard split conformal framework, our prediction set based on conformalized dynamic programming is defined by (\ref{eq:pred-set-un}) with the conformity score (\ref{eq:score-un}).

\begin{theorem}\label{thm:unsupervised}
Consider i.i.d. observations $Y_1,\cdots,Y_{2n},Y_{2n+1}$ generated by some distribution $P$ on $\mathbb{R}$. Let $\widehat{C}_{\rm CP-DP}$ be the split conformal prediction set defined by the score (\ref{eq:score-un}) based on a nested system $\{S_j\}_{j\in[m]}$ satisfying Assumption \ref{as:ne-un}. Suppose the parameter $\delta$ in Assumption \ref{as:ne-un} satisfies $\delta\gg \sqrt{\frac{k+\log n}{n}}$. Then the following properties hold.
\begin{enumerate}
\item Coverage: $\mathbb{P}\left(Y_{2n+1}\in\widehat{C}_{\rm CP-DP}\right)\geq 1-\alpha$.
\item Restricted volume optimality: \\$\vol(\widehat{C}_{\rm CP-DP})\leq \opt_k\left(P,1-\alpha+\frac{1}{n}+4\delta+\gamma\right)$ with probability at least $1-2\delta$.
\end{enumerate}
\end{theorem}

We emphasize that Theorem \ref{thm:unsupervised} guarantees both distribution-free coverage and distribution-free volume optimality properties. In practice, $k$ is usually chosen to be a constant for prediction interpretability. By setting $\gamma=O\left(\sqrt{\frac{\log n}{n}}\right)$, the volume sub-optimality is at most $\frac{1}{n}+4\delta+\gamma=O\left(\sqrt{\frac{\log n}{n}}\right)$.

\section{Supervised Setting}\label{sec:labeled}

\subsection{Problem Setting}

In this section, we consider conformal prediction with labeled data. Suppose data points $(X_1,Y_1)$,$\cdots$, $(X_{2n},Y_{2n})$,$(X_{2n+1},Y_{2n+1})$ are $i.i.d.$ drawn from a distribution $P$ on $\calX \times \calY$ with $\calY = \R$. Using the first $2n$ samples, our goal is to compute a prediction set $\widehat{C}(x)$ for each $x \in \calX$. We will study the following properties for the prediction set.
\begin{enumerate}
\item \textit{Marginal Coverage:}
$$
\mathbb{P}\left(Y_{2n+1} \in \widehat{C}(X_{2n+1})\right)\geq 1-\alpha,
$$
where the probability $\mathbb{P}$ is jointly over all $2n+1$ pairs of observations.
\item \textit{Conditional Coverage:}
\begin{equation}
\mathbb{P}\left(Y_{2n+1}\in \widehat{C}(X_{2n+1}) \mid X_{2n+1}\right)\geq 1-\alpha, \label{eq:con-co}
\end{equation}
with high probability.
\end{enumerate}
It is well known that the conditional coverage property cannot be achieved without additional assumptions on $P$ \citep{vovk2012conditional,lei2014distribution,foygel2021limits}. Therefore, some form of relaxation of (\ref{eq:con-co}) is necessary.

In addition to the coverage properties listed above, we will also extend the notion of restricted volume optimality (\ref{eq:res-opt}) from the unsupervised setting to the supervised setting. Define the conditional CDF by
$$F(y\mid x)=\mathbb{P}\left(Y_{2n+1}\leq y \mid X_{2n+1}=x\right).$$
The conditional restricted optimal volume is given by
\begin{eqnarray}
\nonumber && \opt_k\left(F(\cdot\mid x),1-\alpha\right) \\
\nonumber  &=& \inf\left\{\vol(C): \int_C\mathrm{d} F(\cdot\mid x)\geq 1-\alpha, C\in\mathcal{C}_k\right\}.
\end{eqnarray}
With this definition, we can list the following volume requirement.
\begin{enumerate}
  \setcounter{enumi}{2}
  \item \textit{Conditional Restricted Volume Optimality:}
\begin{equation}
\hspace{-2.5em}
\vol(\widehat{C}(X_{2n+1}))\leq \opt_k\left(F(\cdot|X_{2n+1}),1-\alpha+\epsilon\right), \label{eq:con-vo}
\end{equation}
with high probability, for some $\epsilon\in(0,\alpha)$.
\end{enumerate}
Similar to the conditional coverage property (\ref{eq:con-co}), the conditional restricted volume optimality (\ref{eq:con-vo}) is only required for a typical value of the design point. We will show that based on an extension of distributional conformal prediction \citep{chernozhukov2021distributional}, these two properties can be achieved under the same assumption.

\subsection{Distributional Conformal Prediction}
Conformal prediction based on estimating the conditional CDF has been considered independently by \cite{izbicki2020flexible,chernozhukov2021distributional}. 
We will briefly review the version by~\citet{chernozhukov2021distributional}, and then extend it to serve our purpose. Suppose $\widehat{F}(y\mid x)$ is an estimator of the conditional CDF, which is computed from the first half of the data $(X_1,Y_1),\cdots,(X_n,Y_n)$.
The prediction set proposed by~\citet{chernozhukov2021distributional} is
$$\widehat{C}_{\rm DCP}(X_{2n+1})=\left\{y\in\mathbb{R}:\left|\widehat{F}(y\mid X_{2n+1})-\frac{1}{2}\right|\leq \widehat{t}\right\},$$
where $\widehat{t}$ is an appropriate quantile of $$\left\{\left|\widehat{F}(Y_{n+1}\mid X_{n+1})-\frac{1}{2}\right|,\cdots,\left|\widehat{F}(Y_{2n}\mid X_{2n})-\frac{1}{2}\right|\right\}.$$
Since $\widehat{C}_{\rm DCP}(X_{2n+1})$ is in the form of split conformal prediction, the marginal coverage property is automatically satisfied. When $\widehat{F}(y\mid x)$ is close to $F(y\mid x)$ in some appropriate sense, it was proved by \cite{chernozhukov2021distributional} that asymptotic conditional coverage also holds. However, in general, $\widehat{C}_{\rm DCP}(X_{2n+1})$ is not optimal in terms of its volume. A modification was also proposed in \cite{chernozhukov2021distributional} to achieve volume optimality within the class of intervals. Though not explicitly stated in \cite{chernozhukov2021distributional}, we believe that the DCP procedure essentially achieves (\ref{eq:con-vo}) for $k=1$. Our goal is to achieve the conditional restricted volume optimality for a general $k$ by combining the ideas of DCP and dynamic programming (DP).

\subsection{DCP meets DP}\label{sec:DCP-DP}

To achieve (\ref{eq:con-vo}) for a general $k$, we will modify the DCP procedure by considering a different conformity score that generalizes (\ref{eq:score-un}) to the supervised setting. Recall that $\widehat{F}(y\mid x)$ is an estimator of the conditional CDF, and it is computed from the first half of the data $(X_1,Y_1),\cdots,(X_n,Y_n)$. Our first step is to construct a nested system for each $x\in\calX$. To be specific, for each $x\in\calX$, we will construct a collection of sets $\{S_j(x)\}_{j\in[m]}$ based on the function $\widehat{F}(\cdot\mid x)$. The requirement of the nested system is summarized as the following assumption.

\begin{assumption}\label{as:ne-su}
The sets $S_1(x)\subset\cdots\subset S_m(x)\subset\mathbb{R}$ are measurable with respect to the $\sigma$-field generated by $\widehat{F}(\cdot\mid x)$. Moreover, for some positive integer $k$, some $\alpha\in(0,1)$ and some $\delta,\gamma$ such that $3\delta+\gamma+n^{-1}\leq \alpha$, we have
\begin{enumerate}
\item $\int_{S_j(x)}\mathrm{d} \widehat{F}(\cdot\mid x)=\frac{j}{m}$ and $S_j\in\calC_k$ for all $j\in[m]$.
\item There exists some $j^*\in[m]$, such that $\int_{S_{j^*}(x)}\mathrm{d} \widehat{F}(\cdot\mid x)\geq 1-\alpha+n^{-1}+3\delta$ and $\vol(S_{j^*})\leq \opt_k(\widehat{F}(\cdot\mid x),1-\alpha+\frac{1}{n}+3\delta+\gamma)$.
\end{enumerate}
\end{assumption}
The construction of nested systems satisfying Assumption \ref{as:ne-su} is similar to that in the unsupervised setting. That is, one can apply dynamic programming (Algorithm \ref{alg:dp}) and obtain $S_{j^*}(x)$, and the rest of the sets can be constructed via the greedy expansion/contraction procedure described in Section \ref{sec:con-nest-m} to satisfy $\int_{S_j(x)}\mathrm{d} \widehat{F}(\cdot\mid x)=\frac{j}{m}$.  The main difference here is that Algorithm \ref{alg:dp} is directly applied to the data in the unsupervised setting, while we only have access to $\widehat{F}(\cdot\mid x)$ in the supervised setting. This issue can be easily addressed by computing quantiles $Y_1(x),\cdots,Y_L(x)$ from $\widehat{F}(\cdot\mid x)$ on a grid, and then apply Algorithm \ref{alg:dp} with $Y_1(x),\cdots,Y_L(x)$ as input. Indeed, since the distance between $\widetilde{F}(\cdot\mid x)$ and $\widehat{F}(\cdot\mid x)$ can be controlled by the size of the grid with $\widetilde{F}(y\mid x)=\frac{1}{L}\sum_{l=1}^L\mathbb{I}\{Y_l(x)\leq y\}$, Assumption \ref{ass:F}, which will be stated later in Section \ref{sec:theory}, is also satisfied by $\widetilde{F}(y\mid x)$ (with a slightly larger value of $\delta$) by triangle inequality.

The computational cost of constructing $\{S_j(x)\}_{j\in[m]}$ for a single $x\in\calX$ is $O(L^3k/\gamma)$. Note that there is no need to repeat the construction for each individual $x\in\calX$. Since the split conformal framework only requires evaluating the conformity score at $(X_{n+1},Y_{n+1}),\cdots,(X_{2n},Y_{2n}),(X_{2n+1},y)$, it is sufficient to compute $\{S_j(X_i)\}_{j\in[m]}$ for $i=n+1,\cdots,2n+1$, which leads to the total computational cost $O(nL^3k/\gamma)$.

With nested systems satisfying Assumption \ref{as:ne-su}, the conformity score in the supervised setting is defined as
\begin{equation*}
q(y,x) = \sum_{j=1}^m\mathbb{I}\{y\in S_j(x)\}.
\end{equation*}
Let $q_1\leq \cdots\leq q_n$ be the order statistics computed from the set
\begin{equation*}
\left\{q(X_{n+1},Y_{n+1}),\cdots, q(X_{2n},Y_{2n})\right\}.
\end{equation*}
The prediction set for $Y_{2n+1}$ is constructed as
\begin{equation*}
\widehat{C}_{\rm DCP-DP}(X_{2n+1}) = \left\{y: q(y,X_{2n+1})\geq q_{\lfloor(n+1)\alpha\rfloor}\right\}. \label{eq:dcp-dp}
\end{equation*}

\subsection{Theoretical Guarantees}\label{sec:theory}

We will show in this section that $\widehat{C}_{\rm DCP-DP}(X_{2n+1})$ satisfies marginal coverage. Moreover, when $\widehat{F}(y\mid x)$ is close to $F(y\mid x)$ in some appropriate sense, it also satisfies approximate conditional coverage and conditional restricted volume optimality. Given two CDFs $\widehat{F}$ and $F$, we define the $(k,\infty)$ norm of the difference by
$$\|\widehat{F}-F\|_{k,\infty}=\sup_{C\in\calC_k}\left|\int_C\mathrm{d}\widehat{F}-\int_C\mathrm{d}F\right|.$$

\begin{assumption}\label{ass:F}
The estimated conditional CDF $\widehat{F}(y \mid x)$ satisfies 
$$\mathbb{P}\left(\|\widehat{F}(\cdot \mid X_{2n+1})-F(\cdot \mid X_{2n+1})\|_{k,\infty}\leq\delta\right)\geq 1-\delta,$$
where $\delta$ takes the same value as the one in Assumption \ref{as:ne-su}.
\end{assumption}

The theoretical properties of $\widehat{C}_{\rm DCP-DP}(X_{2n+1})$ are given by the theorem below.

\begin{theorem}\label{thm:supervised}
Consider i.i.d. observations $(X_1,Y_1)$, $\dots$, $(X_{2n},Y_{2n})$, $(X_{2n+1},Y_{2n+1})$ generated by some distribution $P$ on $\calX \times \R$. The conformal prediction set $\widehat{C}_{\rm DCP-DP}(X_{2n+1})$ is computed from nested systems $\{S_j(\cdot)\}_{j\in[m]}$ and $\widehat{F}(\cdot\mid\cdot)$ satisfying Assumption \ref{as:ne-su} and Assumption \ref{ass:F}. Suppose the parameter $\delta$ in the two assumptions satisfies $\delta^2\geq \frac{\log(2\sqrt{n})}{2n}$. Then the following properties hold. 
    \begin{enumerate}
        \item Marginal coverage, 
        $$\mathbb{P}\left(Y_{2n+1} \in \widehat{C}_{\rm DCP-DP}(X_{2n+1})\right)\geq 1-\alpha.$$
        \item Approximate conditional coverage,
        $$\mathbb{P}\left(Y_{2n+1}\in \widehat{C}_{\rm DCP-DP}(X_{2n+1})| X_{2n+1}\right)\!\geq\! 1-\alpha-3\delta,$$
        with probability at least $1-\delta$.
        \item Conditional restricted volume optimality,
        \begin{align*}
        &\vol\left(\widehat{C}_{\rm DCP-DP}(X_{2n+1})\right) \\
        \leq& \opt_k\left(F(\cdot\mid X_{2n+1}),1-\alpha+\frac{1}{n}+4\delta+\gamma\right),
        \end{align*}
        with probability at least $1-2\delta$.
    \end{enumerate} 
\end{theorem}

Theorem \ref{thm:supervised} can be regarded as a generalization of Theorem \ref{thm:unsupervised}. Indeed, when $F(\cdot\mid x)$ does not depend on $x$ and $\widehat{F}(\cdot\mid x)$ is defined as the empirical CDF of $Y_1,\cdots,Y_n$, Theorem \ref{thm:supervised} recovers Theorem \ref{thm:unsupervised}. Moreover, since the volume optimality is over all sets that are unions of $k$ intervals, it also covers the length optimality of intervals considered by \cite{chernozhukov2021distributional}. The case $k\geq 2$ will be important if the conditional density of $Y$ given $X$ has multiple modes; Gaussian mixture is a leading example.

\section{Numerical Experiments}\label{sec:numerical_experiments}

We complement our theoretical guarantees with an evaluation of our methods for both the unsupervised setting of Section~\ref{sec:unlabeled} and the supervised setting of Section~\ref{sec:labeled}. 

\subsection{Construction of Nested Systems}\label{sec:con-nest-m}

We first describe a procedure that generates a nested system $\{S_j\}_{j\in[m]}$ that satisfies Assumption \ref{as:ne-un}. The construction involves the following steps:
\begin{enumerate}
\item \textit{Generate $S_{j^*}$ by Dynamic Programming.} For $j^* = \lceil(1-\alpha + n^{-1} + 3\delta) m \rceil$, we generate $S_{j^*}$ by applying Algorithm \ref{alg:dp} with coverage level $1-j^*/m$ and slack $\gamma=1/m$.
\item \textit{Generate $S_{j^*+1},\cdots,S_m$ by Greedy Expansion.} For each $j>j^*$, we iteratively identify the closest uncovered data point to the boundary of the current $k$ intervals and expand the nearest interval to cover it. Once the intervals cover $\lceil jn/m \rceil$ data points, we define the union as $S_j$ and move on to the construction of $S_{j+1}$.
\item \textit{Generate $S_1,\cdots, S_{j^*-1}$ by Greedy Contraction.} For each $j<j^*$, we iteratively remove a boundary point of the current $k$ intervals that results in the maximum volume reduction. Once the intervals after contraction cover exactly $\lceil jn/m \rceil$ data points, we define the union as $S_j$ and move on to the construction of $S_{j-1}$.
\end{enumerate}

In the supervised setting, the above procedure will be applied to quantiles $Y_1(x),\cdots,Y_L(x)$ computed from $\widehat{F}(\cdot\mid x)$ with $L=m$ for all $x\in\{X_{n+1},\cdots,X_{2n+1}\}$.

\subsection{Comparison in Unsupervised Settings}

The algorithm in Section~\ref{sec:unlabeled} is compared against the method based on kernel density estimation due to \citet{Lei2013DistributionFreePS} and evaluated on several different  distributions. 
Though the original conformalized KDE was proposed in the full conformal framework, we will consider its split conformal version for a direct comparison. We believe the comparison between the full conformal versions of the two methods will lead to the same conclusion.
For the conformalized DP method, the conformity score is constructed based on the nested system described in Section \ref{sec:con-nest-m} with $m=50$ and $\delta = \sqrt{(k + \log n)/n}$. 
The conformalized KDE is also in the form of (\ref{eq:pred-set-un}), with the conformity score given by
$q_{\rm KDE}(x)=\frac{1}{n\rho}\sum_{i=1}^nK\left(\frac{y-Y_i}{\rho}\right)$,
where $K(\cdot)$ is the standard Gaussian kernel and $\rho$ is the bandwidth parameter.
Both methods involve a single tuning parameter, $k$ for conformalized DP and $\rho$ for conformalized KDE.
\begin{figure}[htbp]
    \centering
    \begin{subfigure}[b]{0.48\columnwidth}
        \includegraphics[width=\textwidth]{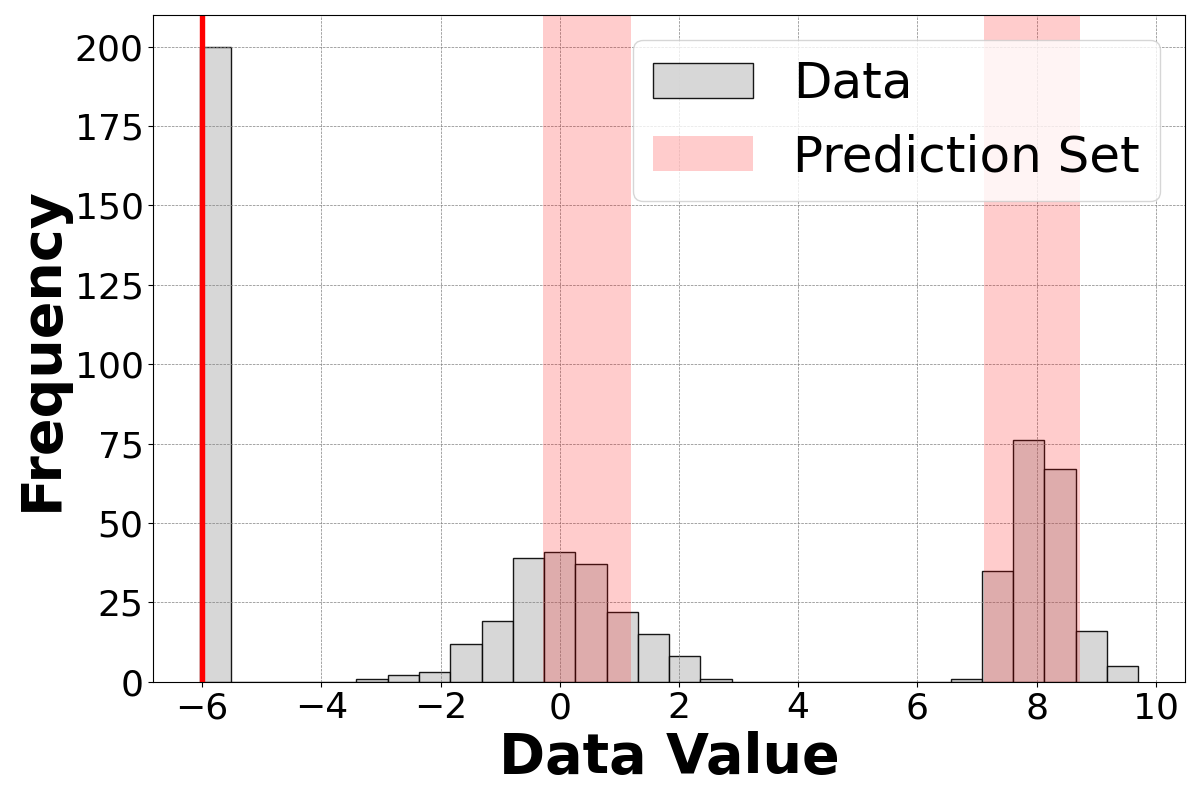}
        \caption{The histogram of the dataset and the prediction set given by conformalized DP with $k=3$ intervals (The first interval is at $[-6.03,-5.97]$.). The volume of the prediction sets is $3.1438$.}
    \end{subfigure}
    \hfill
    \begin{subfigure}[b]{0.48\columnwidth}
        \includegraphics[width=\textwidth]{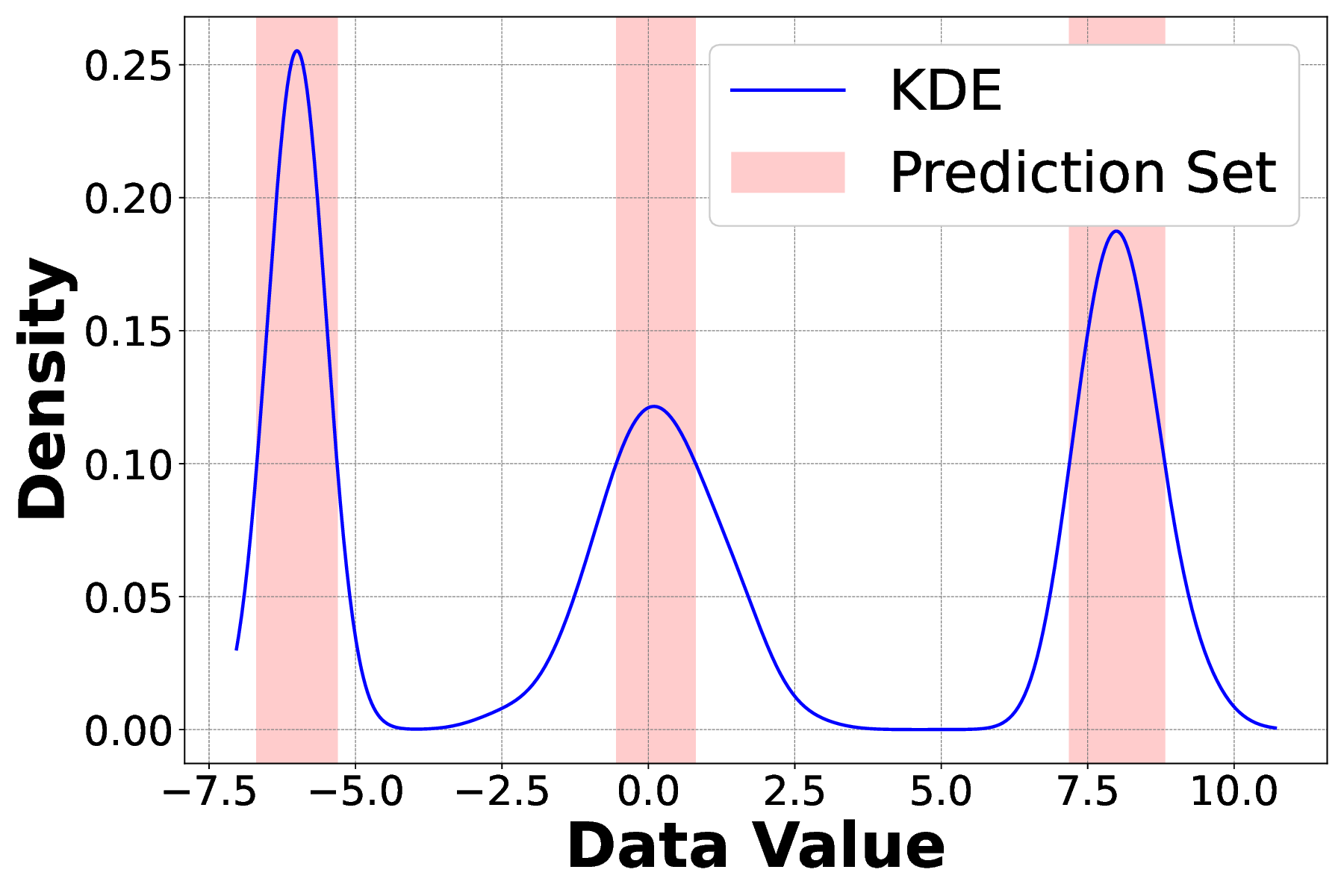}
        \caption{The kernel density estimation with bandwidth $\rho=0.5$ and the prediction set given by conformalized KDE \citep{Lei2013DistributionFreePS}. The volume of the prediction sets is $4.4775$.}
    \end{subfigure}
    \hfill
    \begin{subfigure}[b]{0.48\columnwidth}
        \includegraphics[width=\textwidth]{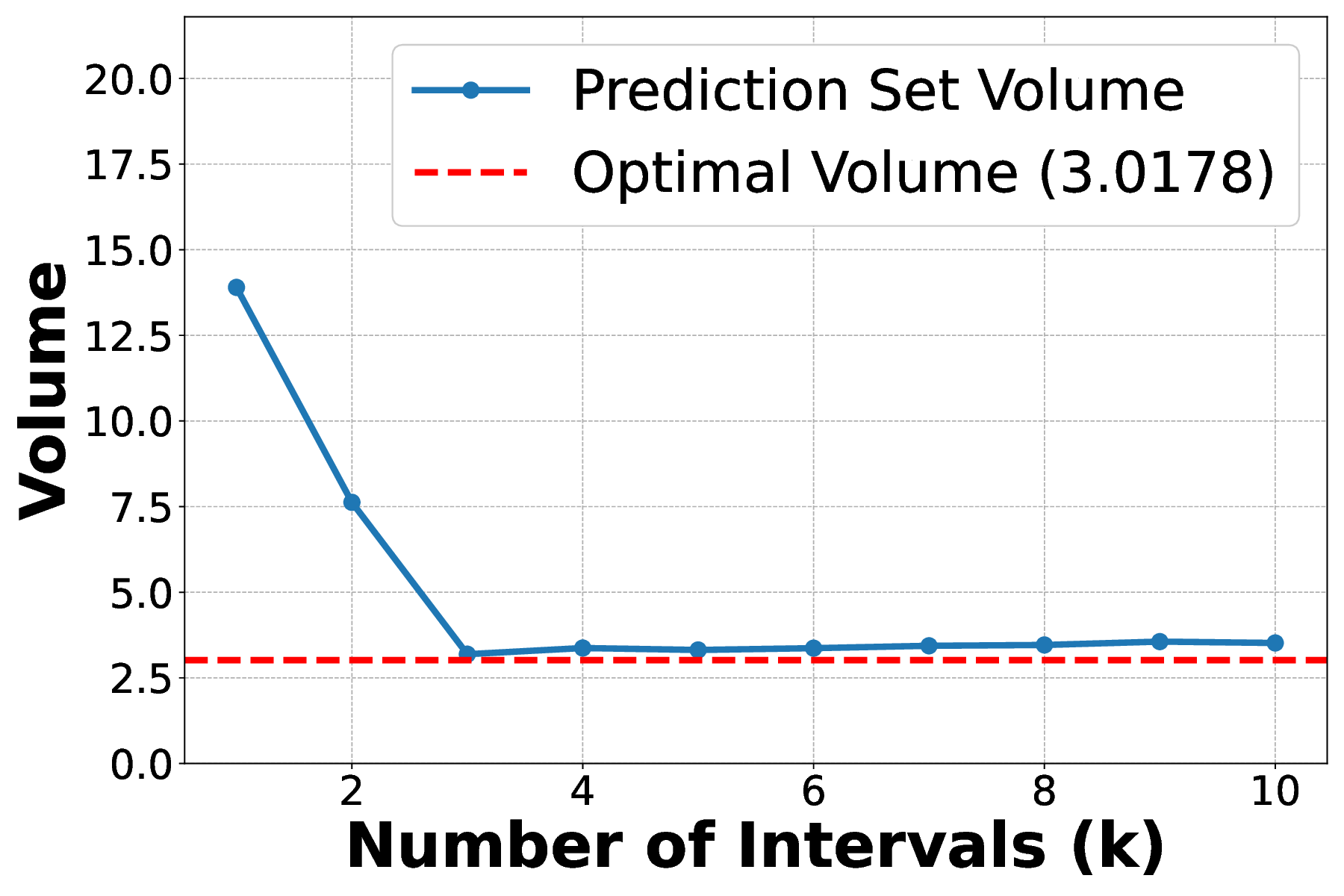}
        \caption{Volumes of prediction sets by conformalized DP is not sensitive to the choice of $k$.}
    \end{subfigure}
    \hfill
    \begin{subfigure}[b]{0.48\columnwidth}
        \includegraphics[width=\textwidth]{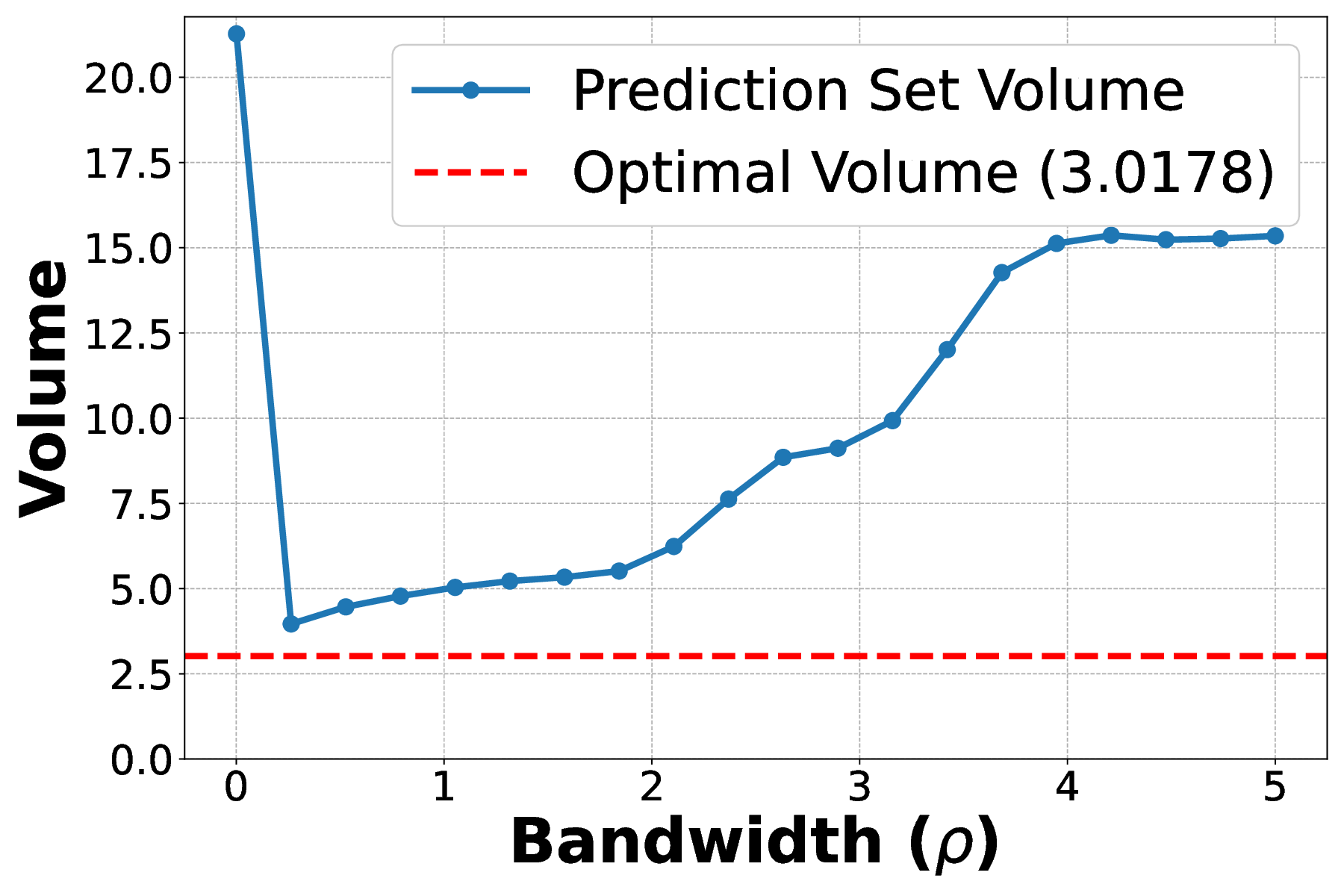}
        \caption{Volumes of prediction sets by conformalized KDE is highly densitive to the choice of $\rho$.}
    \end{subfigure}        
    \caption{Conformal prediction sets on the mixture of Gaussians data from $P = \frac{1}{3}N(-6,0.0001)+\frac{1}{3}N(0,1)+\frac{1}{3}N(8,0.25)$. The coverage probability is $80\%$.
    The theoretically optimal volume is $3.0178$.}
    \label{fig:intro:unlabeled}
\end{figure}
Figure~\ref{fig:intro:unlabeled} summarizes the results using data generated from a mixture of Gaussians $\frac{1}{3}N(-6,0.0001)+\frac{1}{3}N(0,1)+\frac{1}{3}N(8,0.25)$. Additional experiments on other distributions including standard Gaussian, censored Gaussian and ReLU-transformed Gaussian will be presented in Appendix~\ref{sec:numerical}.

\subsection{Comparison in Supervised Settings}

The algorithms for the supervised setting are compared against conformalized quantile regression (CQR)  \citep{romano2019conformalized},  distributional conformal prediction methods (DCP-QR and DCP-QR*) of \citet{chernozhukov2021distributional}, and CD-Split and HPD-Split methods~\citep{izbicki2022cd} against benchmark simulated datasets in \citet{romano2019conformalized, izbicki2020flexible} (Figures \ref{fig:intro:labeled} and~\ref{fig:exp:bimodal}). The implementation details of all methods are given in Appendix~\ref{sec:numerical}. 
\begin{figure}[H]
    \centering
    \begin{subfigure}{0.48\columnwidth}
        \includegraphics[width=\textwidth]{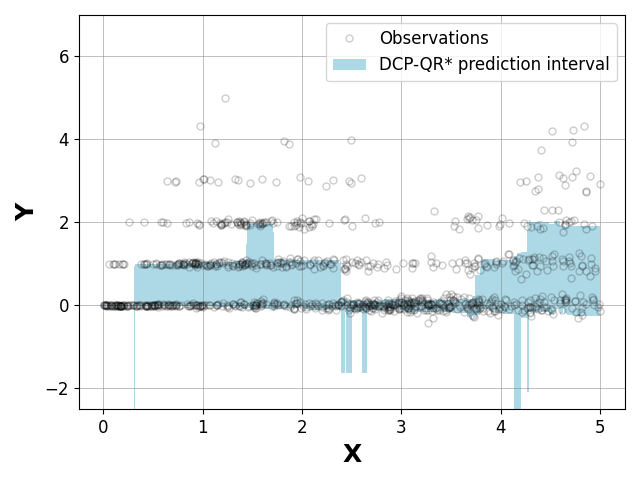}
    \end{subfigure}
    \hfill
    \begin{subfigure}{0.48\columnwidth}
        \includegraphics[width=\textwidth]{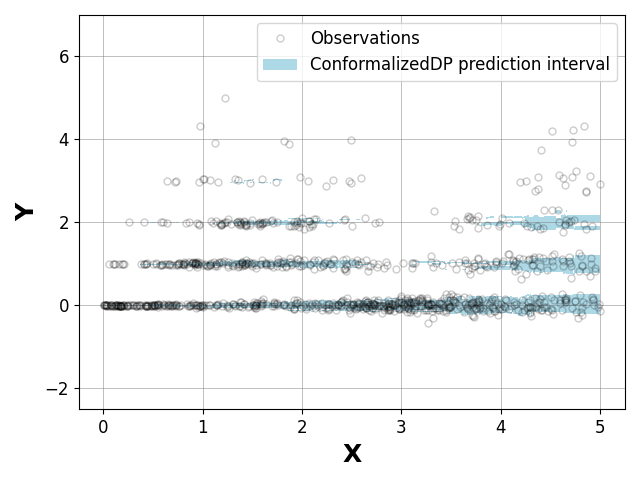}
    \end{subfigure}
    \caption{Results in the supervised setting on a synthetic data from \citet{romano2019conformalized} for target coverage 0.7.  The left plot shows the output of DCP-QR*, the state of the art method by \citet{chernozhukov2021distributional}, which outputs prediction sets with average volume 1.29.  The right plot shows the output of our method with \(k = 5\) intervals, which achieves a significantly improved average volume of 0.45. }
    \label{fig:intro:labeled}
\end{figure}

\begin{figure}[H]
    \centering
    \begin{subfigure}{0.48\columnwidth}
        \includegraphics[width=\textwidth]{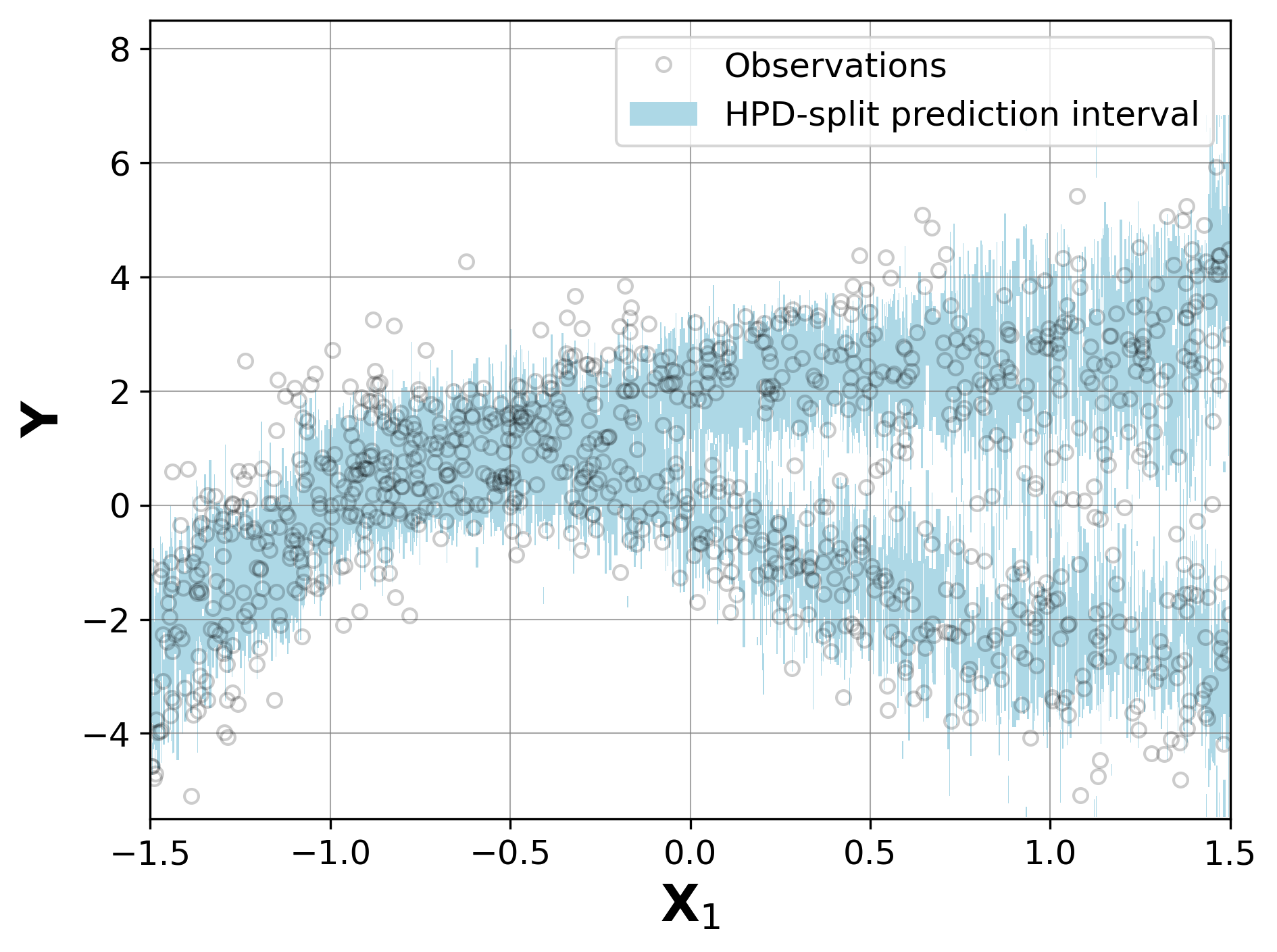}
    \end{subfigure}
    \hfill
    \begin{subfigure}{0.48\columnwidth}
        \includegraphics[width=\textwidth]{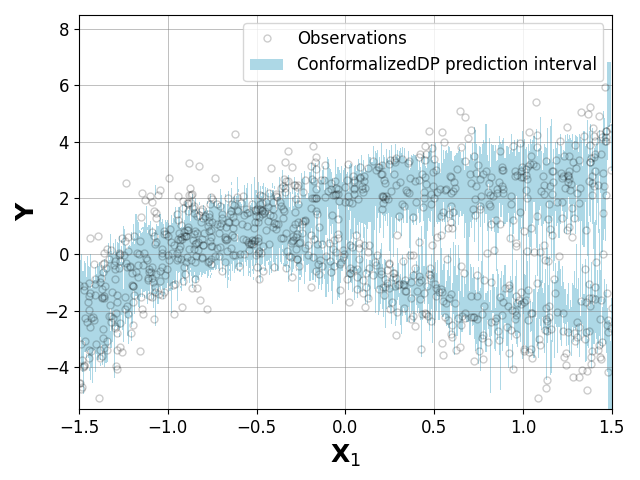}
    \end{subfigure}
    \caption{Results in the supervised setting on a synthetic data with $20$ dimensional feature from \citet{izbicki2020flexible} for target coverage 0.7. The left plot shows the output of HPD-Split method by~\citet{izbicki2022cd}, with average volume $3.60$. The right plot shows the output of our method with $k=2$ intervals, which has an average volume $3.55$.}
    \label{fig:exp:bimodal}
\end{figure}

\begin{table}[H]
    \centering
    \caption{Comparison on simulated data in~\citet{romano2019conformalized}.}
    \label{tab:conformal_results1}
    \begin{tabular}{p{5cm} p{5cm} p{5cm}}
        \toprule
        \textbf{Method} & \textbf{Average Volume} & \textbf{Empirical Coverage (\%)} \\
        \midrule
        CQR & 1.42 & 70.62 \\
        DCP-QR & 1.48 & 71.60 \\
        DCP-QR* & 1.29 & 71.06 \\
        CD-split & 1.83 & 69.94 \\
        HPD-split & 1.75 & 69.44 \\
        Conformalized-DP ($k$=1) & 1.14 & 74.04 \\
        Conformalized-DP ($k$=5) & \textbf{0.45} & 72.36 \\
        \bottomrule
    \end{tabular}
\end{table}

\begin{table}[H]
    \centering
    \caption{Comparison on simulated data in~\citet{izbicki2020flexible}.}
    \label{tab:conformal_results2}
    \begin{tabular}{p{5cm} p{5cm} p{5cm}}
        \toprule
        \textbf{Method} & \textbf{Average Volume} & \textbf{Empirical Coverage (\%)} \\
        \midrule
        CQR & 4.10 & 71.54 \\
        DCP-QR & 4.04 & 70.85 \\
        DCP-QR* & 4.05 & 69.66 \\
        CD-split & 3.69 & 69.86 \\
        HPD-split & 3.60 & 69.64 \\
        Conformalized-DP ($k$=1) & 4.00 & 68.98 \\
        Conformalized-DP ($k$=2) & \textbf{3.55} & 69.42 \\
        \bottomrule
    \end{tabular}
\end{table}

As shown in Tables~\ref{tab:conformal_results1} and~\ref{tab:conformal_results2}, our method  outperforms previous methods by outputting prediction sets that are unions of intervals. Among all other methods, the CD-split and HDP-split \citep{izbicki2020flexible} are also able to produce unions of intervals. However, since these methods rely on consistent estimation of the conditional density, our conformalized DP still produces prediction sets with smaller volumes. The comparison is more pronounced on the first dataset (see Figure \ref{fig:intro:labeled} and Table~\ref{tab:conformal_results1}), where it would be inappropriate to assume a smooth conditional density, but our method is based on conformalizing the estimation of conditional CDF, and thus works in much more general settings.

These experiments together demonstrate that our methods give a clear advantage over competitive procedures under these different conditions. More crucially, our methods also come with theoretical guarantees for both coverage and volume-optimality with respect to unions of $k$ intervals. 

Finally, we remark that our focus on the family of structured prediction sets conceptually differs from the predominant approach of designing new conformity score functions in existing approaches. 
The prediction sets that are output e.g., unions of intervals, are often more interpretable.
Apart from being more interpretable, the shift in focus to structured prediction sets, which is in the same spirit of \cite{gupta2022nested}, is what allows us to overcome any sample efficiency concerns, to obtain {\em distribution-free volume-optimality}, through {computationally efficient algorithms}.

\section{Conclusion}

We study conformal prediction with volume optimality 
in both the unsupervised setting and the supervised setting, 
by proposing a new conformity score computed via dynamic programming. In the supervised setting, when consistent estimation of the conditional CDF is available, we prove that the proposed method not only achieves conditional coverage, but the output of prediction set also has approximate conditional volume optimality with respect to the class of unions of $k$ intervals.

Our method is especially suitable to settings where the data generating process is multi-modal or  has a mixture structure. The numerical experiments show that the performance of the method is quite insensitive to the choice of $k$, whenever it is not smaller than the number of modes of the distribution. For future work, it would be interesting to study restricted volume optimality in more general response settings and under other notions of coverage in conformal prediction.

\section*{Acknowledgments}
This research project was supported by NSF-funded Institute for Data, Econometrics, Algorithms and Learning (IDEAL) through the grants NSF ECCS-2216970 and ECCS-2216912. The research started as part of the IDEAL special program on Reliable and Robust Data Science. Vaidehi Srinivas was supported by the Northwestern Presidential Fellowship.  We also acknowledge the support of the NSF-Simons SkAI institute and the NSF-Simons NITMB institute, and thank Rina Barber and Jing Lei for helpful discussions.

\bibliographystyle{plainnat}
\bibliography{ref}

\begin{thebibliography}{31}
\providecommand{\natexlab}[1]{#1}
\providecommand{\url}[1]{\texttt{#1}}
\expandafter\ifx\csname urlstyle\endcsname\relax
  \providecommand{\doi}[1]{doi: #1}\else
  \providecommand{\doi}{doi: \begingroup \urlstyle{rm}\Url}\fi

\bibitem[Angelopoulos and Bates(2023)]{angelopoulos2023survey}
Anastasios~N. Angelopoulos and Stephen Bates.
\newblock Conformal prediction: A gentle introduction.
\newblock \emph{Found. Trends Mach. Learn.}, 16\penalty0 (4):\penalty0
  494–591, March 2023.
\newblock ISSN 1935-8237.
\newblock \doi{10.1561/2200000101}.
\newblock URL \url{https://doi.org/10.1561/2200000101}.

\bibitem[Angelopoulos et~al.(2024)Angelopoulos, Barber, and
  Bates]{angelopoulos2024theoretical}
Anastasios~N Angelopoulos, Rina~Foygel Barber, and Stephen Bates.
\newblock Theoretical foundations of conformal prediction.
\newblock \emph{arXiv preprint arXiv:2411.11824}, 2024.

\bibitem[Barber et~al.(2023)Barber, Candès, Ramdas, and
  Tibshirani]{barber2023beyond}
Rina Barber, Emmanuel Candès, Aaditya Ramdas, and Ryan Tibshirani.
\newblock Conformal prediction beyond exchangeability.
\newblock \emph{The Annals of Statistics}, 51, 04 2023.
\newblock \doi{10.1214/23-AOS2276}.

\bibitem[Barber et~al.(2021)Barber, Cand{\`e}s, Ramdas, and
  Tibshirani]{barber2021predictive}
Rina~Foygel Barber, Emmanuel~J Cand{\`e}s, Aaditya Ramdas, and Ryan~J
  Tibshirani.
\newblock Predictive inference with the jackknife+.
\newblock \emph{The Annals of Statistics}, 49\penalty0 (1), 2021.

\bibitem[Barron(1989)]{barron1989uniformly}
Andrew~R Barron.
\newblock Uniformly powerful goodness of fit tests.
\newblock \emph{The Annals of Statistics}, 17\penalty0 (1):\penalty0 107--124,
  1989.

\bibitem[Carreira-Perpin{\'a}n and Williams(2003)]{carreira2003number}
Miguel~A Carreira-Perpin{\'a}n and Christopher~KI Williams.
\newblock On the number of modes of a gaussian mixture.
\newblock In \emph{International Conference on Scale-Space Theories in Computer
  Vision}, pages 625--640. Springer, 2003.

\bibitem[Chernozhukov et~al.(2021)Chernozhukov, W{\"u}thrich, and
  Zhu]{chernozhukov2021distributional}
Victor Chernozhukov, Kaspar W{\"u}thrich, and Yinchu Zhu.
\newblock Distributional conformal prediction.
\newblock \emph{Proceedings of the National Academy of Sciences}, 118\penalty0
  (48):\penalty0 e2107794118, 2021.

\bibitem[Devroye and Lugosi(2001)]{devroye2001combinatorial}
Luc Devroye and G{\'a}bor Lugosi.
\newblock \emph{Combinatorial methods in density estimation}.
\newblock Springer Science \& Business Media, 2001.

\bibitem[Foygel~Barber et~al.(2021)Foygel~Barber, Candes, Ramdas, and
  Tibshirani]{foygel2021limits}
Rina Foygel~Barber, Emmanuel~J Candes, Aaditya Ramdas, and Ryan~J Tibshirani.
\newblock The limits of distribution-free conditional predictive inference.
\newblock \emph{Information and Inference: A Journal of the IMA}, 10\penalty0
  (2):\penalty0 455--482, 2021.

\bibitem[Gammerman et~al.(1998)Gammerman, Vovk, and
  Vapnik]{gammerman1998learning}
A.~Gammerman, V.~Vovk, and V.~Vapnik.
\newblock Learning by transduction.
\newblock In \emph{Proceedings of the Fourteenth Conference on Uncertainty in
  Artificial Intelligence}, UAI'98, page 148–155, San Francisco, CA, USA,
  1998. Morgan Kaufmann Publishers Inc.
\newblock ISBN 155860555X.

\bibitem[Ghosal and van~der Vaart(2007)]{ghosal2007posterior}
S~Ghosal and AW~van~der Vaart.
\newblock Posterior convergence rates of dirichlet mixtures at smooth
  densities.
\newblock \emph{Annals of Statistics}, 35\penalty0 (2):\penalty0 697--723,
  2007.

\bibitem[Gupta et~al.(2022)Gupta, Kuchibhotla, and Ramdas]{gupta2022nested}
Chirag Gupta, Arun~K Kuchibhotla, and Aaditya Ramdas.
\newblock Nested conformal prediction and quantile out-of-bag ensemble methods.
\newblock \emph{Pattern Recognition}, 127:\penalty0 108496, 2022.

\bibitem[Izbicki and Lee(2017)]{izbicki2017converting}
Rafael Izbicki and Ann~B Lee.
\newblock Converting high-dimensional regression to high-dimensional
  conditional density estimation.
\newblock \emph{Electronic Journal of Statistics}, 11:\penalty0 2800--2831,
  2017.

\bibitem[Izbicki et~al.(2020)Izbicki, Shimizu, and Stern]{izbicki2020flexible}
Rafael Izbicki, Gilson Shimizu, and Rafael Stern.
\newblock Flexible distribution-free conditional predictive bands using density
  estimators.
\newblock In \emph{International Conference on Artificial Intelligence and
  Statistics}, pages 3068--3077. PMLR, 2020.

\bibitem[Izbicki et~al.(2022)Izbicki, Shimizu, and Stern]{izbicki2022cd}
Rafael Izbicki, Gilson Shimizu, and Rafael~B Stern.
\newblock Cd-split and hpd-split: Efficient conformal regions in high
  dimensions.
\newblock \emph{Journal of Machine Learning Research}, 23\penalty0
  (87):\penalty0 1--32, 2022.

\bibitem[Kiyani et~al.(2024)Kiyani, Pappas, and Hassani]{kiyani2024length}
Shayan Kiyani, George Pappas, and Hamed Hassani.
\newblock Length optimization in conformal prediction, 2024.
\newblock URL \url{https://arxiv.org/abs/2406.18814}.

\bibitem[Kruijer et~al.(2010)Kruijer, Rousseau, and van~der
  Vaart]{kruijer2010adaptive}
Willem Kruijer, Judith Rousseau, and Aad van~der Vaart.
\newblock Adaptive bayesian density estimation with location-scale mixtures.
\newblock \emph{Electronic Journal of Statistics}, 4:\penalty0 1225--1257,
  2010.

\bibitem[Kumar et~al.(2023)Kumar, Lu, Gupta, Palepu, Bellamy, Raskar, and
  Beam]{Kumar2023ConformalPW}
Bhawesh Kumar, Cha-Chen Lu, Gauri Gupta, Anil Palepu, David~R. Bellamy, Ramesh
  Raskar, and Andrew~L. Beam.
\newblock Conformal prediction with large language models for multi-choice
  question answering.
\newblock \emph{ArXiv}, abs/2305.18404, 2023.
\newblock URL \url{https://api.semanticscholar.org/CorpusID:258967849}.

\bibitem[LeCam and Schwartz(1960)]{lecam1960necessary}
Lucien LeCam and Lorraine Schwartz.
\newblock A necessary and sufficient condition for the existence of consistent
  estimates.
\newblock \emph{The Annals of Mathematical Statistics}, 31\penalty0
  (1):\penalty0 140--150, 1960.

\bibitem[Lei and Wasserman(2014)]{lei2014distribution}
Jing Lei and Larry Wasserman.
\newblock Distribution-free prediction bands for non-parametric regression.
\newblock \emph{Journal of the Royal Statistical Society Series B: Statistical
  Methodology}, 76\penalty0 (1):\penalty0 71--96, 2014.

\bibitem[Lei et~al.(2013)Lei, Robins, and Wasserman]{Lei2013DistributionFreePS}
Jing Lei, James~M. Robins, and Larry~A. Wasserman.
\newblock Distribution-free prediction sets.
\newblock \emph{Journal of the American Statistical Association}, 108:\penalty0
  278 -- 287, 2013.
\newblock URL \url{https://api.semanticscholar.org/CorpusID:17499892}.

\bibitem[Meinshausen(2006)]{Meinshausen06Quantile}
Nicolai Meinshausen.
\newblock Quantile regression forests.
\newblock \emph{Journal of Machine Learning Research}, 7\penalty0
  (35):\penalty0 983--999, 2006.
\newblock URL \url{http://jmlr.org/papers/v7/meinshausen06a.html}.

\bibitem[Roebroek(2023)]{sklearn-quantile}
Jasper Roebroek.
\newblock sklearn-quantile, 2023.
\newblock URL
  \url{https://sklearn-quantile.readthedocs.io/en/latest/index.html}.

\bibitem[Romano et~al.(2019)Romano, Patterson, and
  Cand\`{e}s]{romano2019conformalized}
Yaniv Romano, Evan Patterson, and Emmanuel~J. Cand\`{e}s.
\newblock Conformalized quantile regression.
\newblock \emph{Advances in neural information processing systems}, 32, 2019.

\bibitem[Sadinle et~al.(2016)Sadinle, Lei, and Wasserman]{Sadinle2016LeastAS}
Mauricio Sadinle, Jing Lei, and Larry~A. Wasserman.
\newblock Least ambiguous set-valued classifiers with bounded error levels.
\newblock \emph{Journal of the American Statistical Association}, 114:\penalty0
  223 -- 234, 2016.
\newblock URL \url{https://api.semanticscholar.org/CorpusID:622583}.

\bibitem[Scott and Nowak(2006)]{scott2006learning}
Clayton~D. Scott and Robert~D. Nowak.
\newblock Learning minimum volume sets.
\newblock \emph{J. Mach. Learn. Res.}, 7:\penalty0 665–704, December 2006.
\newblock ISSN 1532-4435.

\bibitem[Shafer and Vovk(2008)]{shafer2008tutorial}
Glenn Shafer and Vladimir Vovk.
\newblock A tutorial on conformal prediction.
\newblock \emph{Journal of Machine Learning Research}, 9\penalty0
  (12):\penalty0 371--421, 2008.
\newblock URL \url{http://jmlr.org/papers/v9/shafer08a.html}.

\bibitem[Stutz et~al.(2022)Stutz, Dvijotham, Cemgil, and
  Doucet]{stutz2022learning}
David Stutz, Krishnamurthy~Dj Dvijotham, Ali~Taylan Cemgil, and Arnaud Doucet.
\newblock Learning optimal conformal classifiers.
\newblock In \emph{International Conference on Learning Representations}, 2022.
\newblock URL \url{https://openreview.net/forum?id=t8O-4LKFVx}.

\bibitem[Vovk(2012)]{vovk2012conditional}
Vladimir Vovk.
\newblock Conditional validity of inductive conformal predictors.
\newblock In \emph{Asian conference on machine learning}, pages 475--490. PMLR,
  2012.

\bibitem[Vovk et~al.(2005)Vovk, Gammerman, and Shafer]{Vovkbook}
Vladimir Vovk, Alex Gammerman, and Glenn Shafer.
\newblock \emph{Algorithmic Learning in a Random World}.
\newblock Springer-Verlag, Berlin, Heidelberg, 2005.
\newblock ISBN 0387001522.

\bibitem[Xie et~al.(2024)Xie, Barber, and Candès]{xie2024boosted}
Ran Xie, Rina~Foygel Barber, and Emmanuel~J. Candès.
\newblock Boosted conformal prediction intervals, 2024.
\newblock URL \url{https://arxiv.org/abs/2406.07449}.

\end{thebibliography}

\newpage

\appendix 

%
%
\section{Related Work} \label{sec:related}

\paragraph{Comparison to \citet{Lei2013DistributionFreePS}} 
The influential work of \citet{Lei2013DistributionFreePS}  gave the first theoretical guarantees of volume control or optimality to the best of our knowledge. 
In fact \citet{Lei2013DistributionFreePS} and subsequent follow-up works including \citet{Sadinle2016LeastAS, chernozhukov2021distributional} with theoretical guarantees on volume control study a stricter quantity that corresponds to the volume of set difference $\vol\left(\widehat{C}\Delta C_{\rm opt}\right)$ \citep{Lei2013DistributionFreePS,Sadinle2016LeastAS, chernozhukov2021distributional}. However, this much stronger notion requires that the optimal solution $C_{\rm opt}$ must not only exist but also be unique. Usually additional assumptions need to be imposed in the neighborhood of the boundary of $C_{\rm opt}$ in order that the set difference vanishes in the large sample limit. 
Specifically, the work of \citep{Lei2013DistributionFreePS} assumes that the density is smooth, and in addition is strictly increasing or decreasing significantly. 
In comparison, our notion of volume optimality only requires the volume to be controlled, which can be achieved even if $\widehat{C}$ is not close to $C_{\rm opt}$, or when $C_{\rm opt}$ does not even exist. 
Moreover, we do not need to make any assumptions on the smoothness of the density. In fact, the density may not even exist, and can have discrete point masses or $\delta$ functions as shown in the experiments. Indeed, from a practical point of view, any set with coverage and volume control would serve the purpose of valid prediction. Insisting the closeness to a questionable target $C_{\rm opt}$ comes at the cost of unnecessary assumptions on the data generating process.

\paragraph{Comparison to \citet{izbicki2020flexible,izbicki2022cd}}

The work of \citet{izbicki2020flexible,izbicki2022cd} provided conformal prediction methods that can produce a union of intervals in a supervised setting. Specifically, their methods, CD-split and HPD-split, are designed to leverage level sets of an estimated conditional density function. CD-split achieves local and marginal validity by partitioning the feature space adaptively but does not guarantee conditional coverage in general. 
In contrast, HPD-split simplifies tuning by using a conformity score based on the cumulative distribution function of the conditional density. Under certain assumptions of density estimation accuracy and the uniqueness of the optimal solution, HPD-split achieves asymptotic conditional coverage and converges to the highest predictive density
set which is the smallest volume set with the desired coverage.
In comparison, our method outputs a union of intervals with the smallest length from a direct estimator of the conditional CDF, which only requires the accuracy of conditional CDF estimation. Estimating the conditional CDF is statistically simpler than estimating the conditional density, which usually requires additional smoothness or regularity conditions.

\paragraph{Comparison to \citet{kiyani2024length}} In very recent independent work, \citet{kiyani2024length} considered a min-max approach for conformal prediction in the covariate shift setting with a view towards length optimality of their intervals. They proposed a new method based on minimax optimization to optimize the average volume of prediction sets in the context of covariate shift, which generalizes the marginal or group-conditional coverage setting. 
Their method uses a given (predefined) conformity score, and optimizes the choice of the thresholds $h(X)$ for different covariates $X \in \mathcal{X}$ 
to minimize the average prediction interval length, while maintaining the marginal or group-conditional coverage. Under certain assumptions that the conformity score is consistent with a volume optimal prediction set, they show that solving their minimax optimization will give a volume-optimal solution. However the problem of finding the best threshold function $h(X)$ is a non-convex problem that may be computational inefficient in theory; but they use SGD to find a good heuristic solution in practice. This work is incomparable to this paper in multiple ways. While \citet{kiyani2024length} considers the covariate shift setting with a specific focus on marginal coverage and group coverage, we focus more on the unlabeled setting, and the conditional coverage setting of \citet{chernozhukov2021distributional}. 
In contrast to their method that uses an off-the-shelf conformity score (and optimizes the thresholds $h(X)$), our method introduces a new conformity score function based on dynamic programming to find volume-optimal unions of intervals. This also suggests that our methods and the methods of \citet{kiyani2024length} may potentially be complementary. Finally, by restricting the prediction sets to unions of $k$ intervals, we got theoretical guarantees of volume optimality and get polynomial time algorithms based on dynamic programming to achieve them. Hence, while both their work and our work try to address the important consideration of volume optimality, they are incomparable in terms of the setting, the results and the techniques.

\paragraph{Other Related work}

In the non-conformal setting, the work of \citep{scott2006learning} studied the problem of finding minimal volume sets from a certain set family given samples drawn i.i.d from a distribution, with at least $1-\alpha$ fraction of probability mass. However this work mostly focused on statistical efficiency, and did not consider the conformal inference setting. 
In the past few years, there has been an explosion of literature in conformal inference that develops new conformal methods for various settings \citep[see e.g., ][and references therein]{barber2021predictive, stutz2022learning,Kumar2023ConformalPW, barber2023beyond, xie2024boosted}. To the best of our knowledge these works do not provide theoretical guarantees on volume optimality.

\section{Additional Numerical Experiments} \label{sec:numerical}

\subsection{Construction of Nested Systems}\label{sec:con-nest}

Recall the construction of the nested system described in Section \ref{sec:con-nest-m}.
It immediately follows Proposition \ref{thm:DP} that the construction satisfies Assumption \ref{as:ne-un} in the unsupervised setting.


In the supervised setting, the construction of the nested system is based on $\widehat{F}(y\mid x)$. For each $x\in\{X_{n+1},\cdots,X_{2n+1}\}$, we generate $Y_1(x),\cdots,Y_L(x)$ according to the quantile level
$$Y_l(x)=\argmax\{y:\widehat{F}(y\mid x)\leq l/L\}.$$
Then, the greedy expansion and contraction procedure described in Section \ref{sec:con-nest-m} is applied on $Y_1(x),\cdots,Y_L(x)$. Effectively, this is equivalent to using $\widetilde{F}(y\mid x)=\frac{1}{L}\sum_{l=1}^L\mathbb{I}\{Y_l(x)\leq y\}$ as input. By its definition, $\widetilde{F}(y\mid x)$ is a uniform approximation to $\widehat{F}(y\mid x)$ with error $1/L$. Thus, Assumption \ref{as:ne-su} is still satisfied for $\widetilde{F}(y\mid x)$. In all of our experiment, we set $L=m$.

\begin{remark}
It is clear that the details of the greedy expansion step and the greedy contraction step do not matter much for Assumption \ref{as:ne-un} or \ref{as:ne-su} to be satisfied. However, different choices will indeed affect practical performance, especially in the supervised setting when $\widehat{F}(y\mid x)$ is not close to $F(y\mid x)$. To be more specific, sensible choices of expansion and contraction sets from the $S_{j^*}$ generated by DP will serve as a safety net against model misspecification.  We discuss this in more detail in Section \ref{sec:app-labeled-data}, see e.g. Figure \ref{fig:volume-aware-example}.
\end{remark}

\subsection{Unsupervised Setting}

Given i.i.d. observations $Y_1, Y_2, \dots, Y_{2n} \in \R$ drawn from some distribution $P$, the goal is to find a prediction set $\widehat{C}=\widehat{C}(Y_1, \dots, Y_{2n})$ such that $\Pr(Y_{2n+1} \in \widehat{C}) \geq 1-\alpha$ for an independent future observation $Y_{2n+1}$ drawn from the same $P$. 
We implement the proposed conformalized dynamic programming (DP) method $\widehat{C}_{\rm CP-DP}$, and compare it with the conformalized kernel density estimation (KDE) proposed by~\cite{Lei2013DistributionFreePS} on the following synthetic datasets: (1) Gaussian; (2) Censored Gaussian; (3) Mixture of Gaussians; (4) ReLU-Transformed Gaussian.

Though the original conformalized KDE was proposed in the full conformal framework, we will consider its split conformal version for a direct comparison. We believe the comparison between the full conformal versions of the two methods will lead to the same conclusion.
For the conformalized DP method, the conformity score is constructed based on the nested system described in Section \ref{sec:con-nest} with $m=50$ and $\delta = \sqrt{(k + \log n)/n}$. 
The conformalized KDE is also in the form of (\ref{eq:pred-set-un}), with the conformity score given by
$$q_{\rm KDE}(x)=\frac{1}{n\rho}\sum_{i=1}^nK\left(\frac{y-Y_i}{\rho}\right),$$
where $K(\cdot)$ is the standard Gaussian kernel and $\rho$ is the bandwidth parameter.
Both methods involve a single tuning parameter, $k$ for conformalized DP and $\rho$ for conformalized KDE.

\begin{figure}[H]
    \centering
        \includegraphics[width=0.45\textwidth]{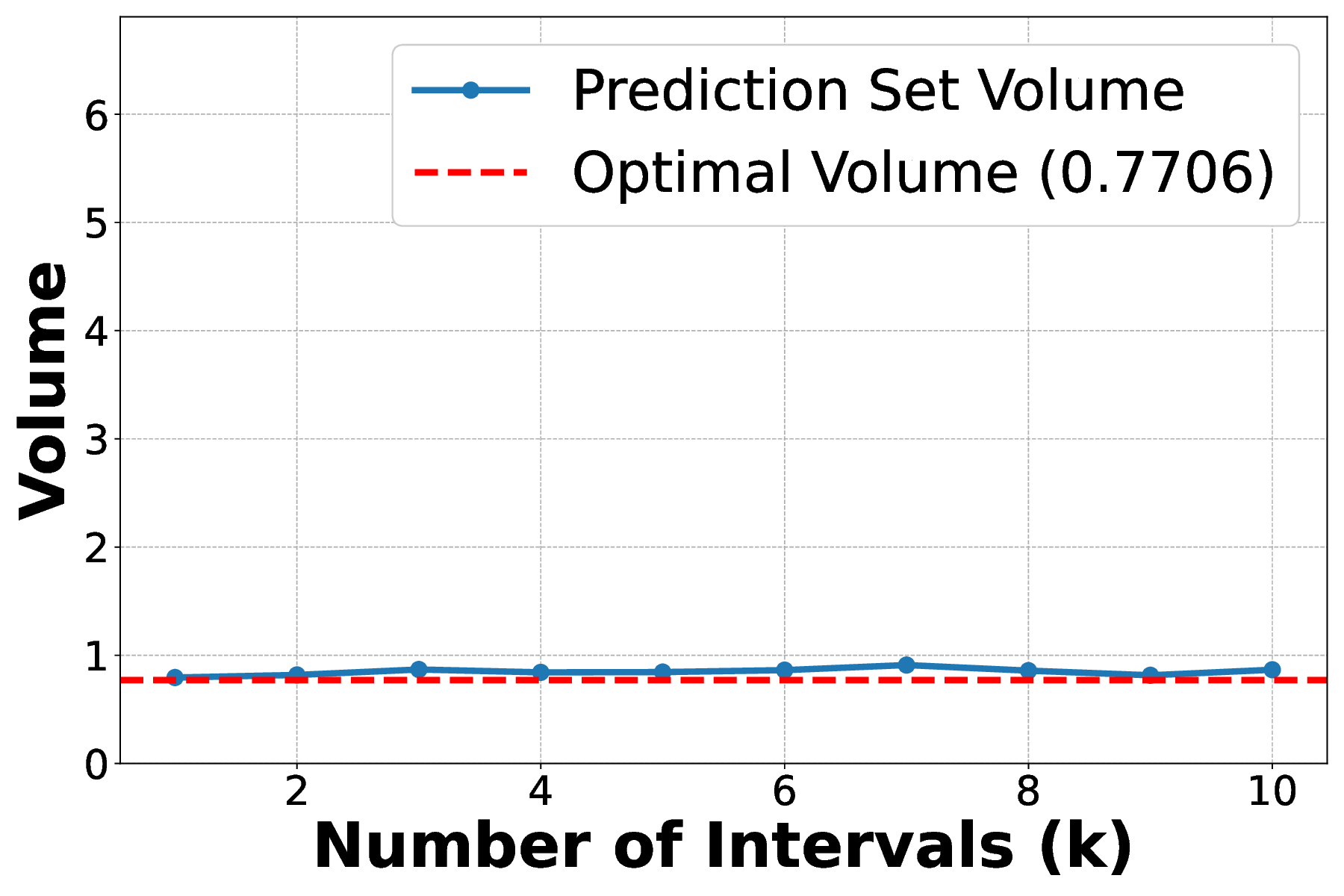}
    \hfill
        \includegraphics[width=0.45\textwidth]{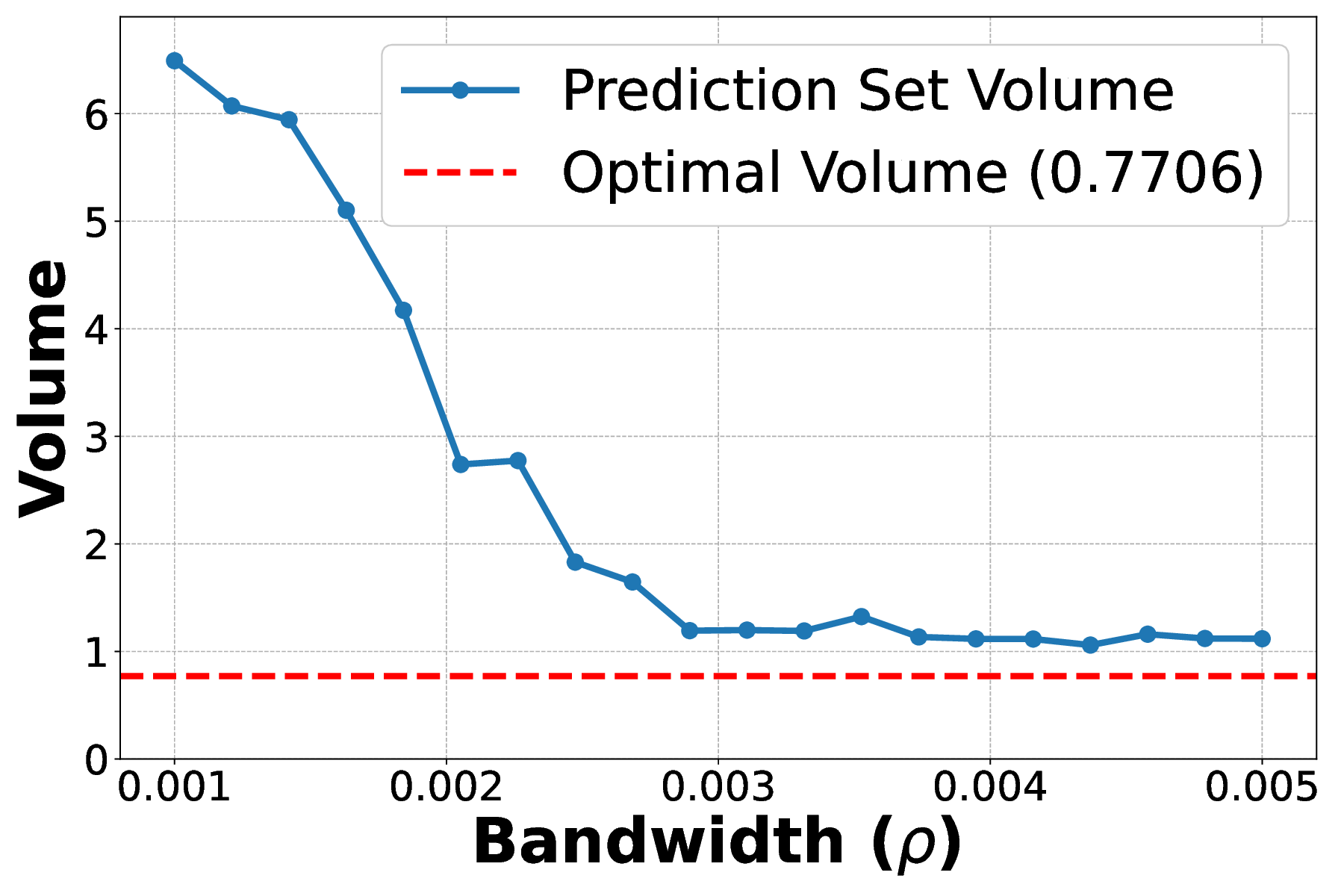}
    \caption{Volumes of prediction sets of the two methods on the Gaussian dataset (blue) and the benchmark $\opt_1(N(0,1),0.3)=0.7706$ (red). The blue curves are computed by averaging $100$ independent experiments. }
    \label{fig:gaussian_length}
\end{figure}

\paragraph{Gaussian:} Our first distribution is $P=N(0,1)$, which is a benign example for sanity check. We consider sample size being $100$, and set the coverage probability $1-\alpha=30\%$ for a more transparent comparison between the two methods. The conformalized DP is computed with number of intervals $k$ ranging from $1$ to $10$. It turns out that the output of the prediction set is quite stable when $k$ varies (Figure~\ref{fig:gaussian_length}). Even for $k=10$, our method still produces a single interval in this unimodal distribution.

\begin{figure}[H]
    \centering
        \includegraphics[width=0.45\textwidth]{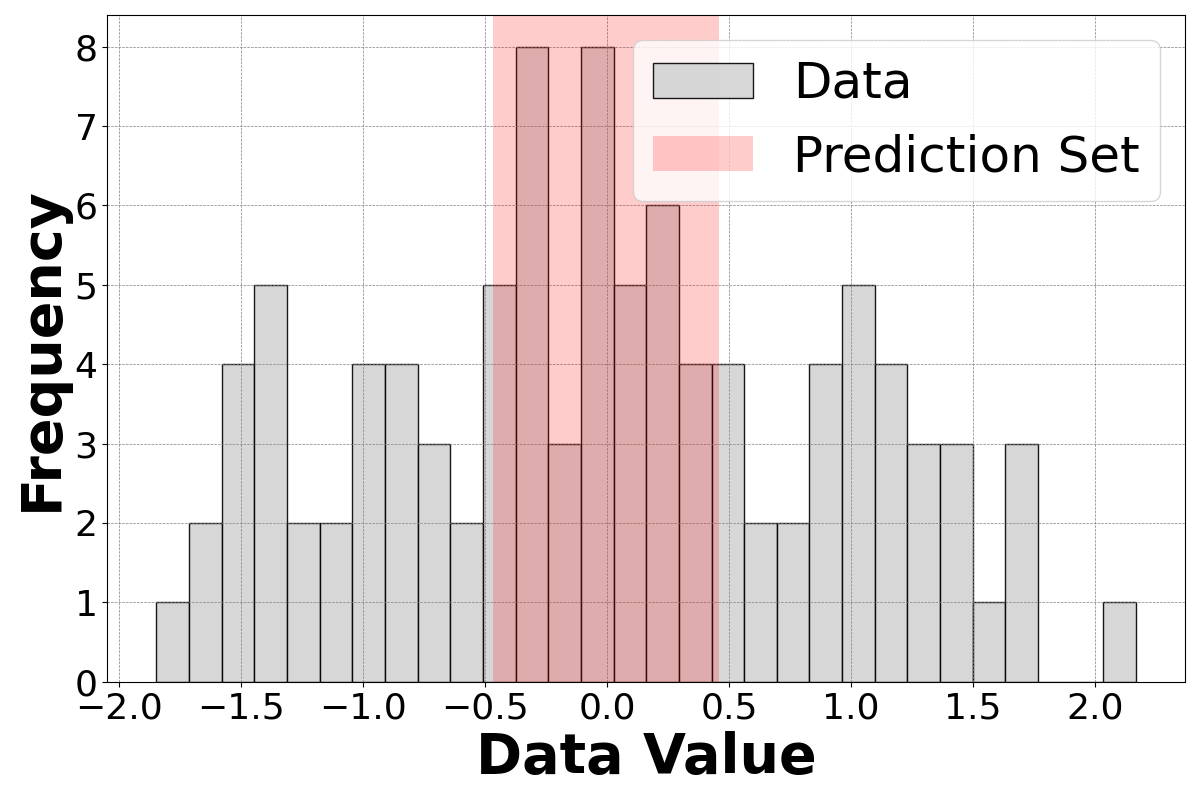}
    \hfill
        \includegraphics[width=0.45\textwidth]{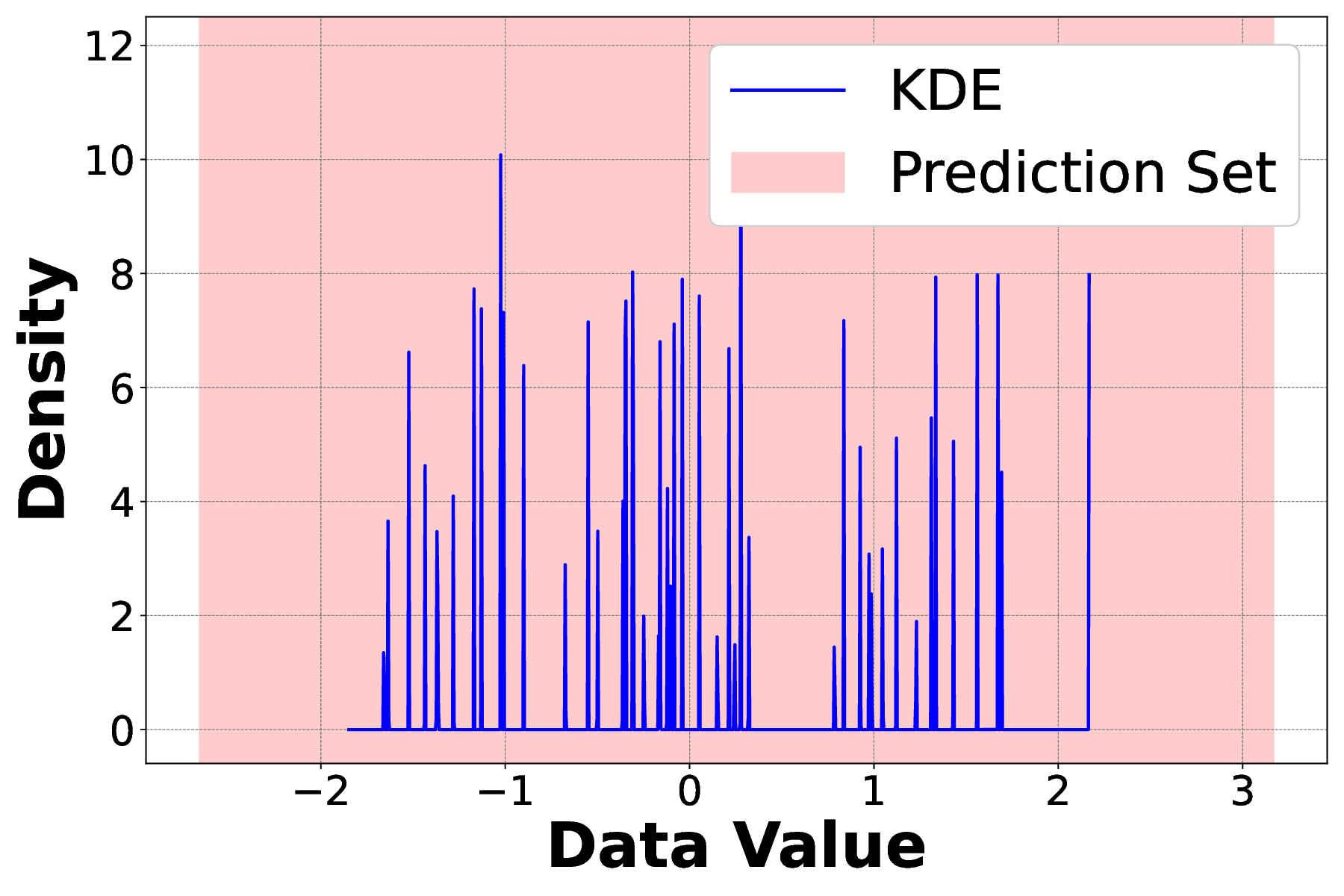}
    \caption{Conformal prediction sets on the Gaussian dataset. The left plot shows the histogram of the dataset and the prediction set produced by conformalized DP with $k=1$; the right plot shows the kernel density estimation with bandwidth $\rho=0.001$ and the prediction set given by the conformalized KDE. }
    \label{fig:gaussian_interval}
\end{figure}

The conformalized KDE is implemented with bandwidth $\rho$ ranging from $0.001$ to $0.005$. We observe that the quality of the prediction set is quite sensitive to the choice of the bandwidth. As is shown by Figure~\ref{fig:gaussian_interval}, if the bandwidth of KDE is too small, the conformal prediction will output almost the entire support of the data set. This is because if the KDE overfits the training samples, the level set of the KDE will likely not cover the future observation. Therefore, a conformal procedure, which guarantees finite sample coverage, has to be conservative by outputting the entire support. Figure~\ref{fig:gaussian_length} shows that this issue will be alleviated as the bandwidth gets larger.

\paragraph{Censored Gaussian:} We next consider $P$ being a censored Gaussian distribution. We take the sample size to be $100$, and each sample can be generated according to $Y_i = \sigma(Z_i+1) - \sigma(Z_i-1)$ with $Z_i\sim N(0,1)$ and $\sigma(t)=\max(t,1)$ being the ReLU transform. This is equivalently a truncated Gaussian distribution, which has a standard Gaussian density on $(0,2)$ and a point mass at $0$ with probability $\Pr(Z_i \leq -1)$ and another point mass at $2$ with probability $\Pr(Z_i \geq 1)$. Again, for the sake of comparison, we set the coverage probability to be $1-\alpha=30\%$.

\begin{figure}[H]
    \centering
        \includegraphics[width=0.45\textwidth]{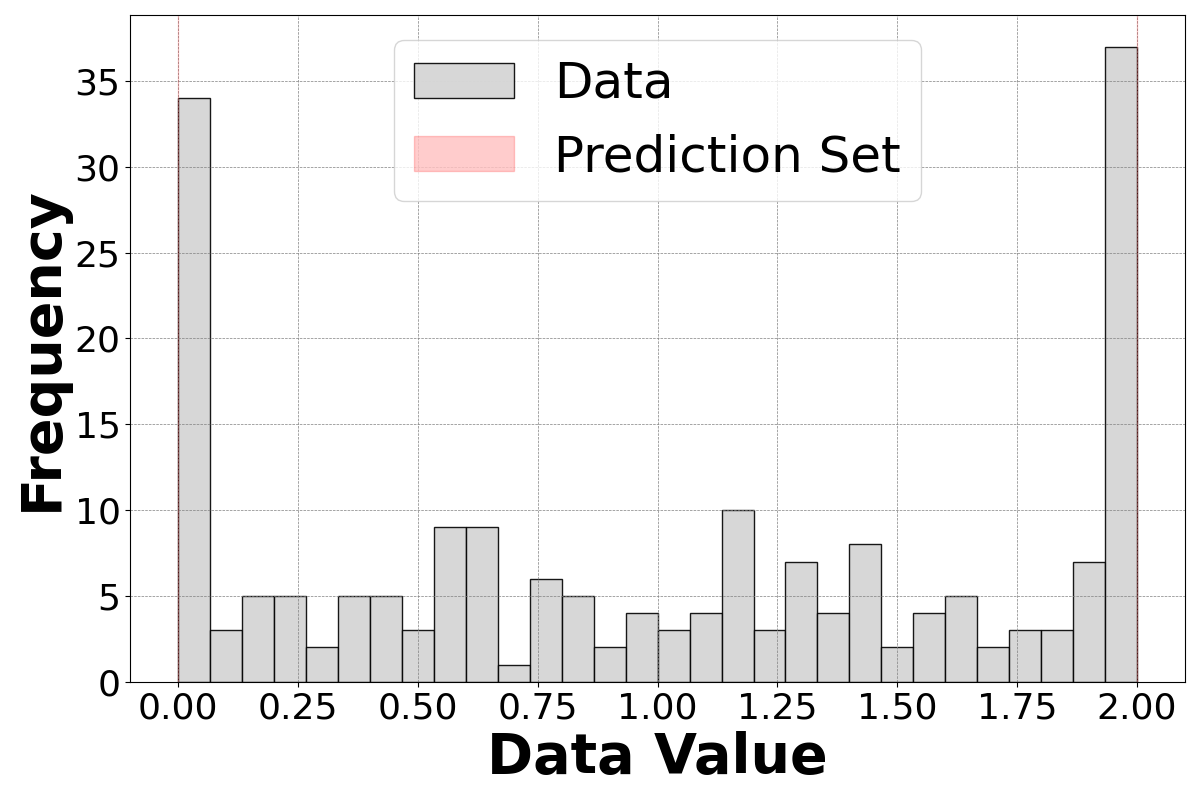}
    \hfill
        \includegraphics[width=0.45\textwidth]{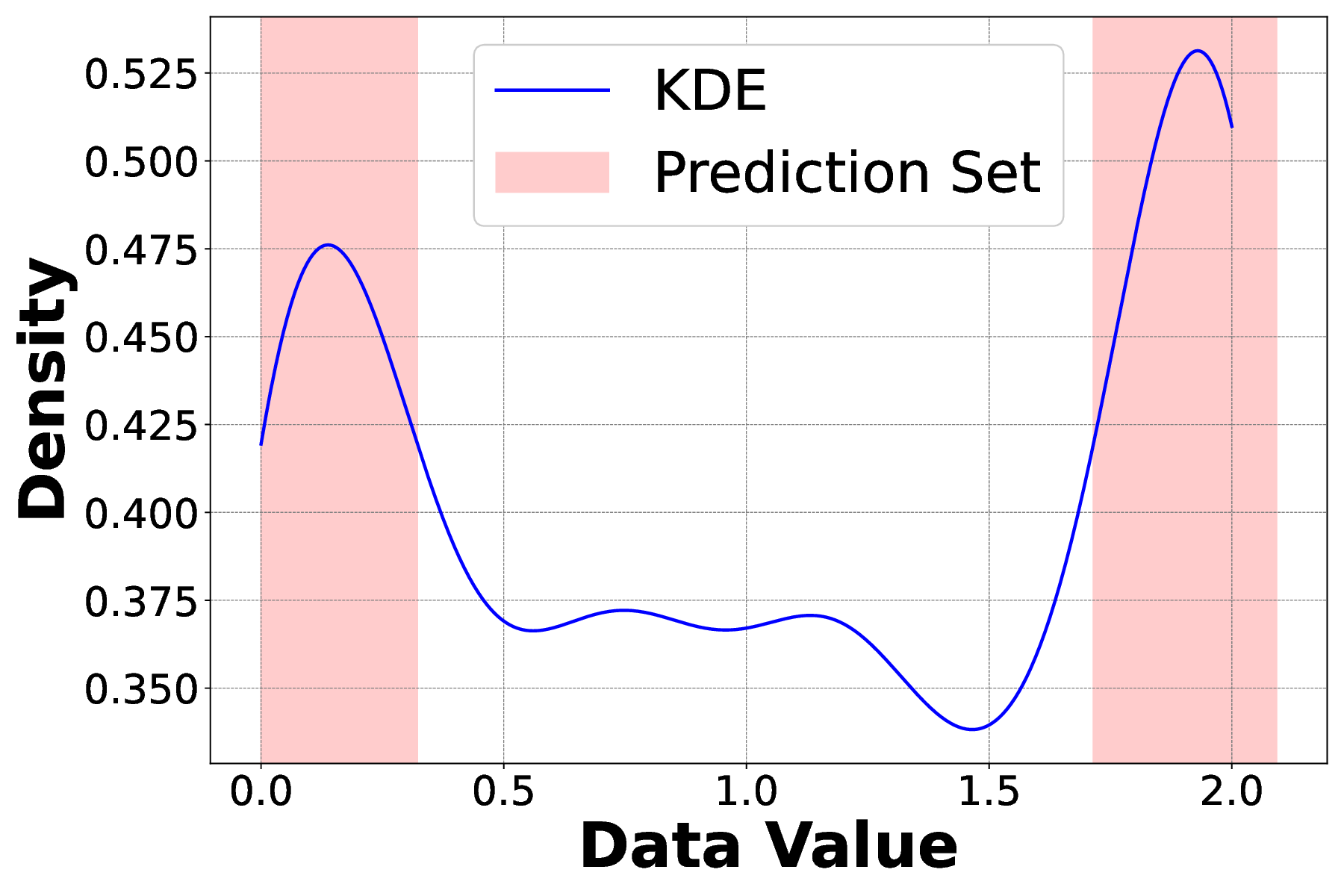}
    \caption{Conformal prediction sets on the censored Gaussian dataset. The left plot shows the histogram of the dataset and the prediction set given by conformalized DP with $k=2$ intervals (The prediction set is two zero-length intervals at $0.0$ and $2.0$). The right plot shows the kernel density estimation with bandwidth $\rho=0.2$ and the prediction set by conformalized KDE.}
    \label{fig:trunc_gaussian_interval}
\end{figure}

\begin{figure}[H]
    \centering
        \includegraphics[width=0.45\textwidth]{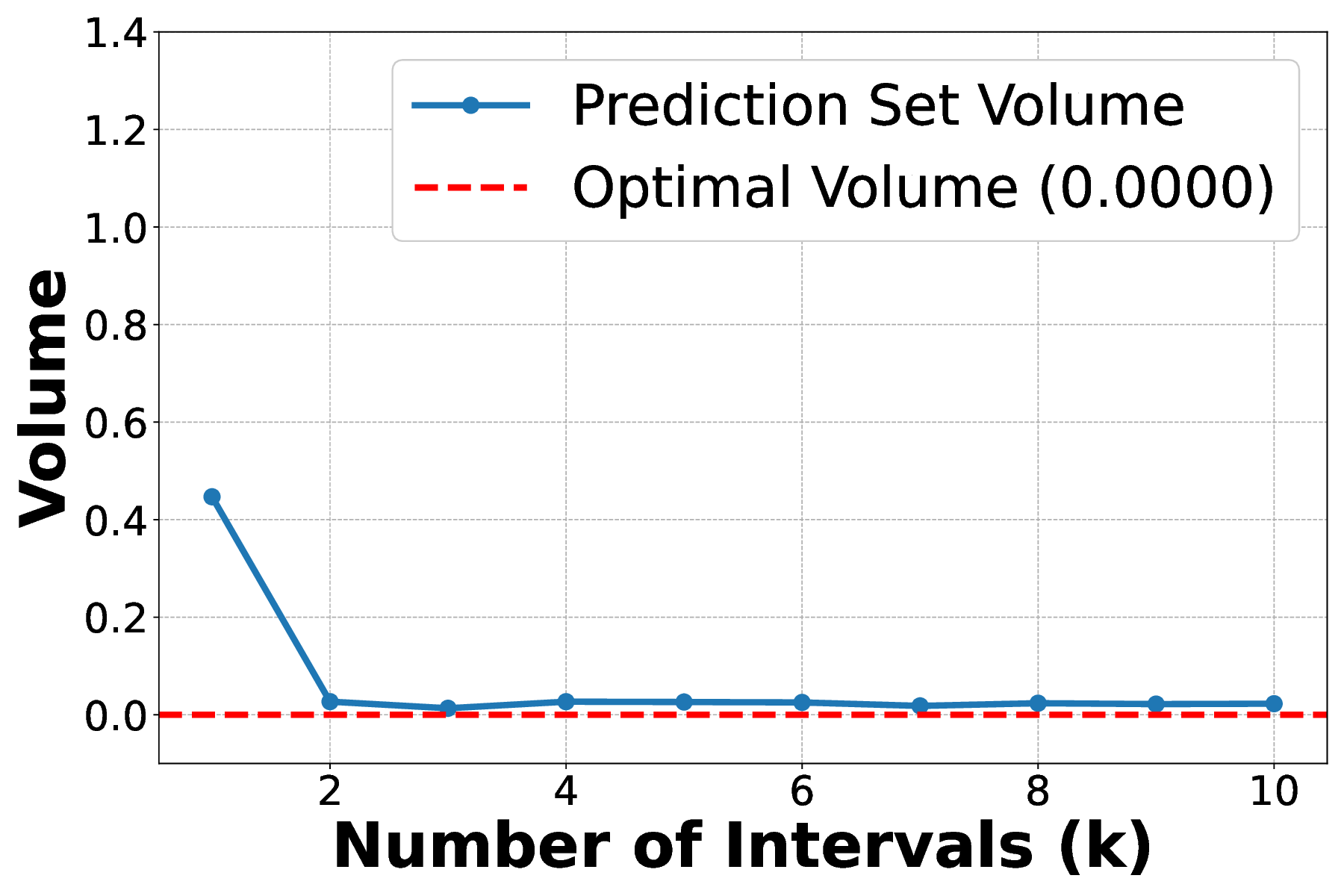}
    \hfill
        \includegraphics[width=0.45\textwidth]{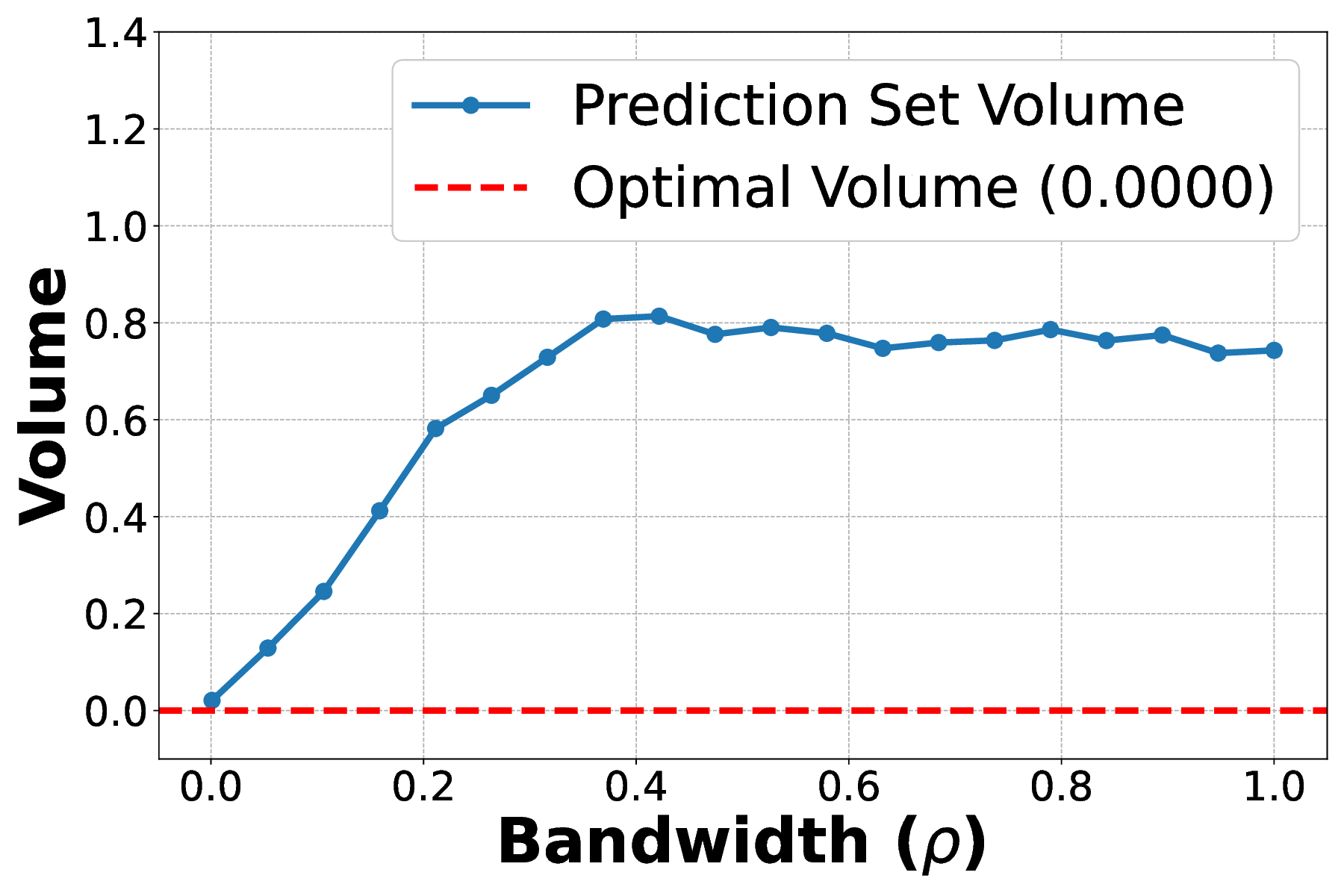}
    \caption{Volumes of prediction sets of the two methods on the censored Gaussian dataset (blue) and the optimal volume (red). The blue curves are computed by averaging $100$ independent experiments.}
    \label{fig:trunc_gaussian_length}
\end{figure}

Since $\Pr(|Z_i| \leq -1) \geq 1-\alpha$, the population optimal volume is $\opt(P,0.3)=\opt_2(P,0.3)=0$ due to the point masses at $\{0,2\}$. By setting $k=2$ for the conformalized DP procedure, the prediction set concentrates on the two point masses (Figure~\ref{fig:trunc_gaussian_interval}). Moreover, it produces very similar results as we increase $k$ up to $10$. Figure~\ref{fig:trunc_gaussian_length} shows that the only exception is $k=1$, since one short interval obviosly cannot cover two points that are far away from each other.

We also run conformalized KDE with bandwidth $\rho$ ranging from $0.001$ to $1$. Since the distribution does not even have a density function on the entire support, KDE is not really suitable for this setting. Not surprisingly, for a typical choice of bandwidth that is not too small, the conformalized KDE will not identify the two point masses due to smoothing (Figure~\ref{fig:trunc_gaussian_interval}). Figure~\ref{fig:trunc_gaussian_length} reports the volume of the prediction set as we vary bandwidth, and the volume of the prediction set is close to optimal only when the bandwidth is extremely close to $0$.

\paragraph{Mixture of Gaussians:}
In this experiment, we consider $P=\frac{1}{3}N(-6,0.0001)+\frac{1}{3}N(0,1)+\frac{1}{3}N(8,0.25)$. The sample size and coverage probability are set as $600$ and $1-\alpha=80\%$, respectively. The two methods are compared with $k$ ranging from $1$ to $10$ in conformalized DP and bandwidth $\rho$ ranging from $0.001$ to $5$ in conformalized KDE.

\begin{figure}[H]
    \centering
    \begin{subfigure}{0.32\textwidth}
        \includegraphics[width=\textwidth]{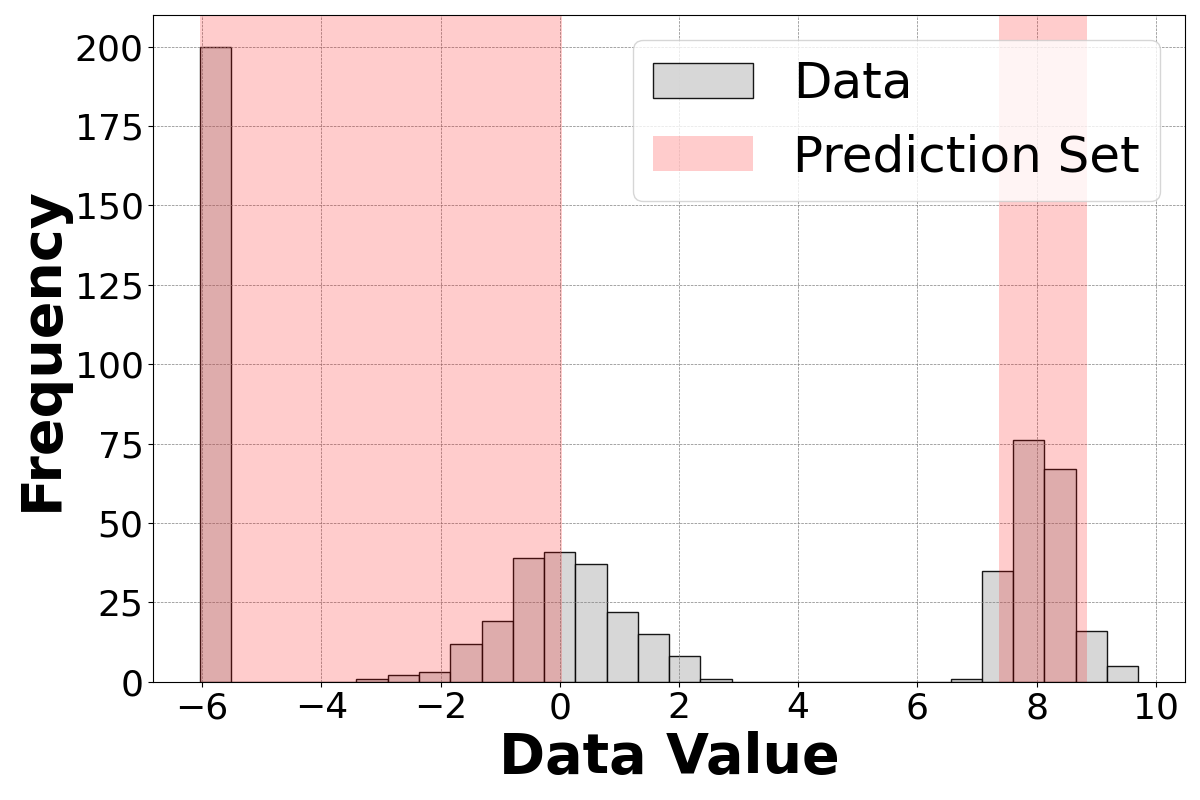}
        \caption{$k=2$}
    \end{subfigure}
    \hfill
    \begin{subfigure}{0.32\textwidth}
        \includegraphics[width=\textwidth]{figures/mix_gaussian3_DP_k3.png}
        \caption{$k=3$}
    \end{subfigure}
    \hfill
    \begin{subfigure}{0.32\textwidth}
        \includegraphics[width=\textwidth]{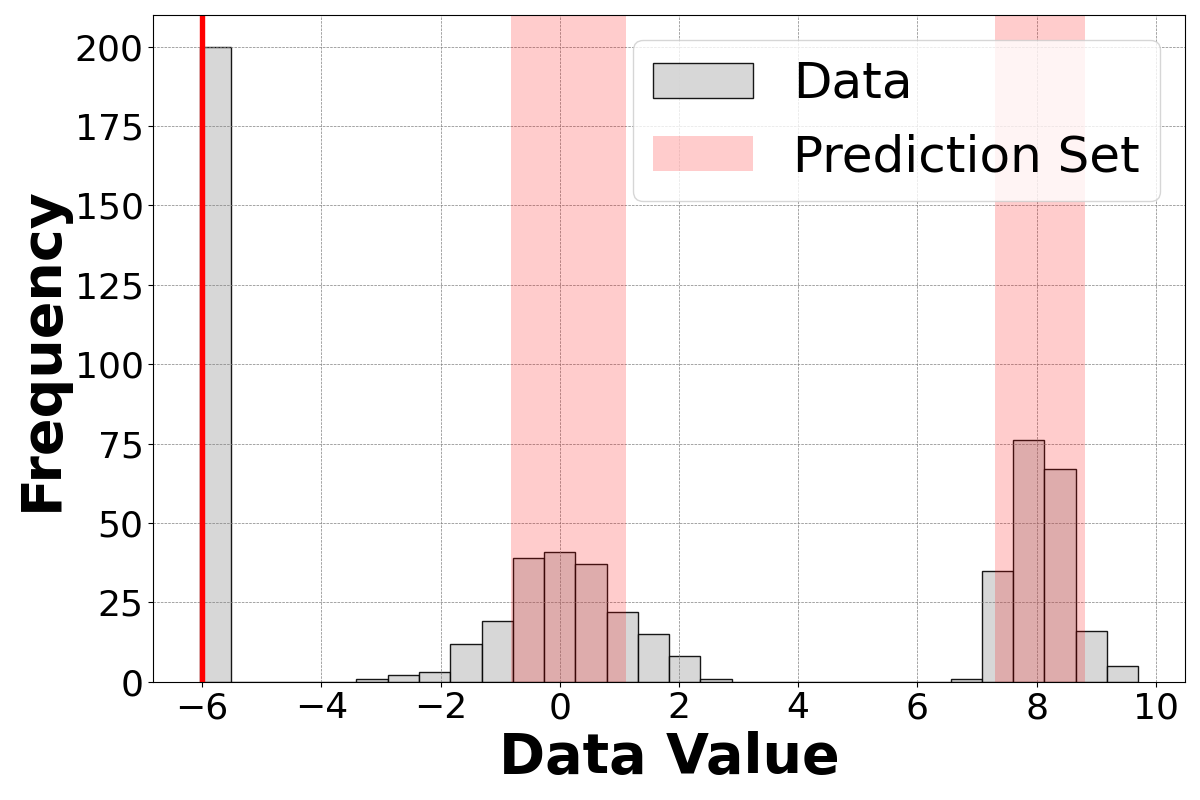}
        \caption{$k=6$}
    \end{subfigure}
    \caption{Prediction sets provided by the conformalized DP method with the number of intervals $k=2,3,6$ on the mixture of Gaussians dataset.}
    \label{fig:mix_gaussian_intervals_DP}
\end{figure}

\begin{figure}[H]
    \centering
    \begin{subfigure}{0.32\textwidth}
        \includegraphics[width=\textwidth]{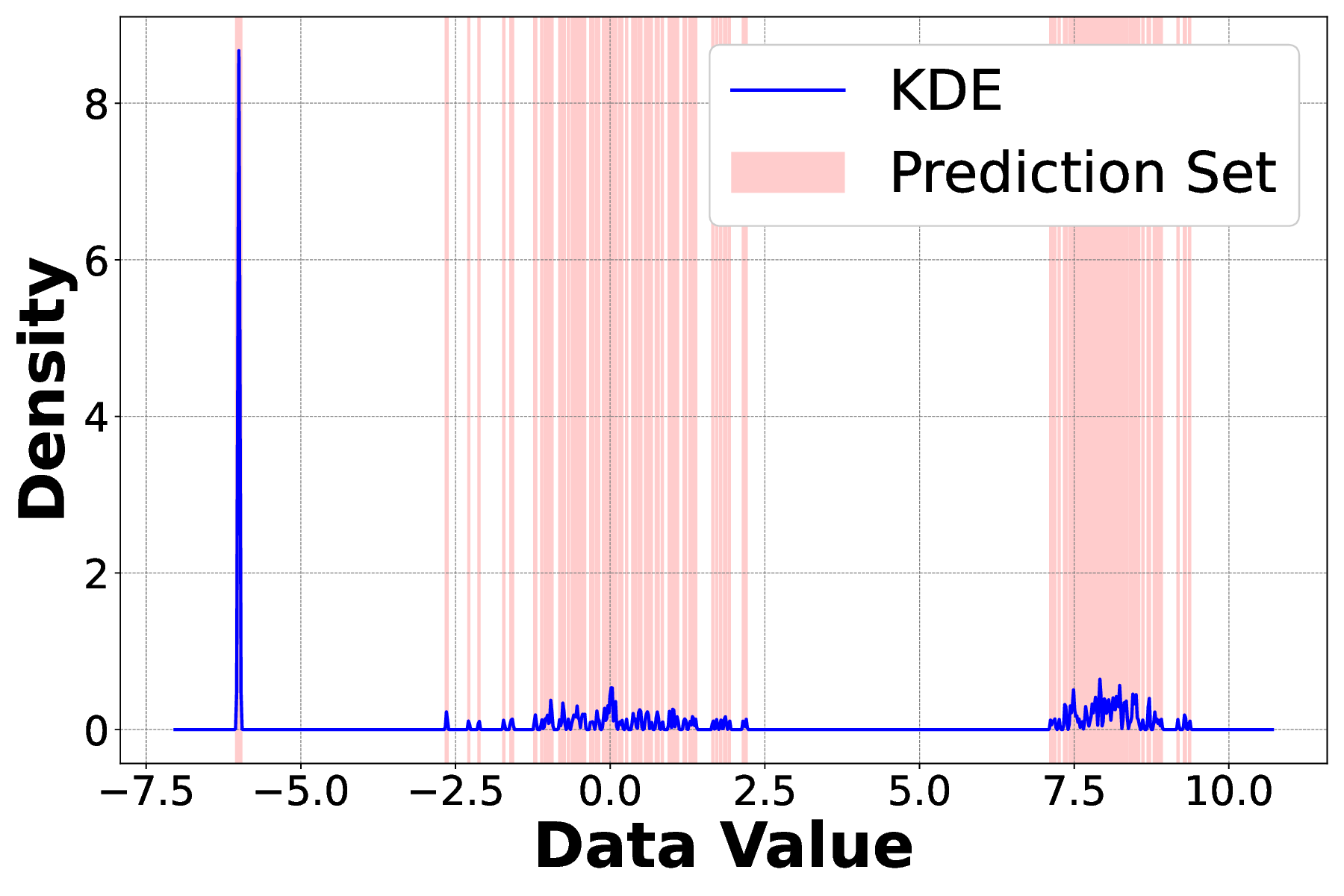}
        \caption{$\rho=0.01$}
    \end{subfigure}
    \hfill
    \begin{subfigure}{0.32\textwidth}
        \includegraphics[width=\textwidth]{figures/mix_gaussian3_KDE_bandwidth0.5.eps}
        \caption{$\rho=0.5$}
    \end{subfigure}
    \hfill
    \begin{subfigure}{0.32\textwidth}
        \includegraphics[width=\textwidth]{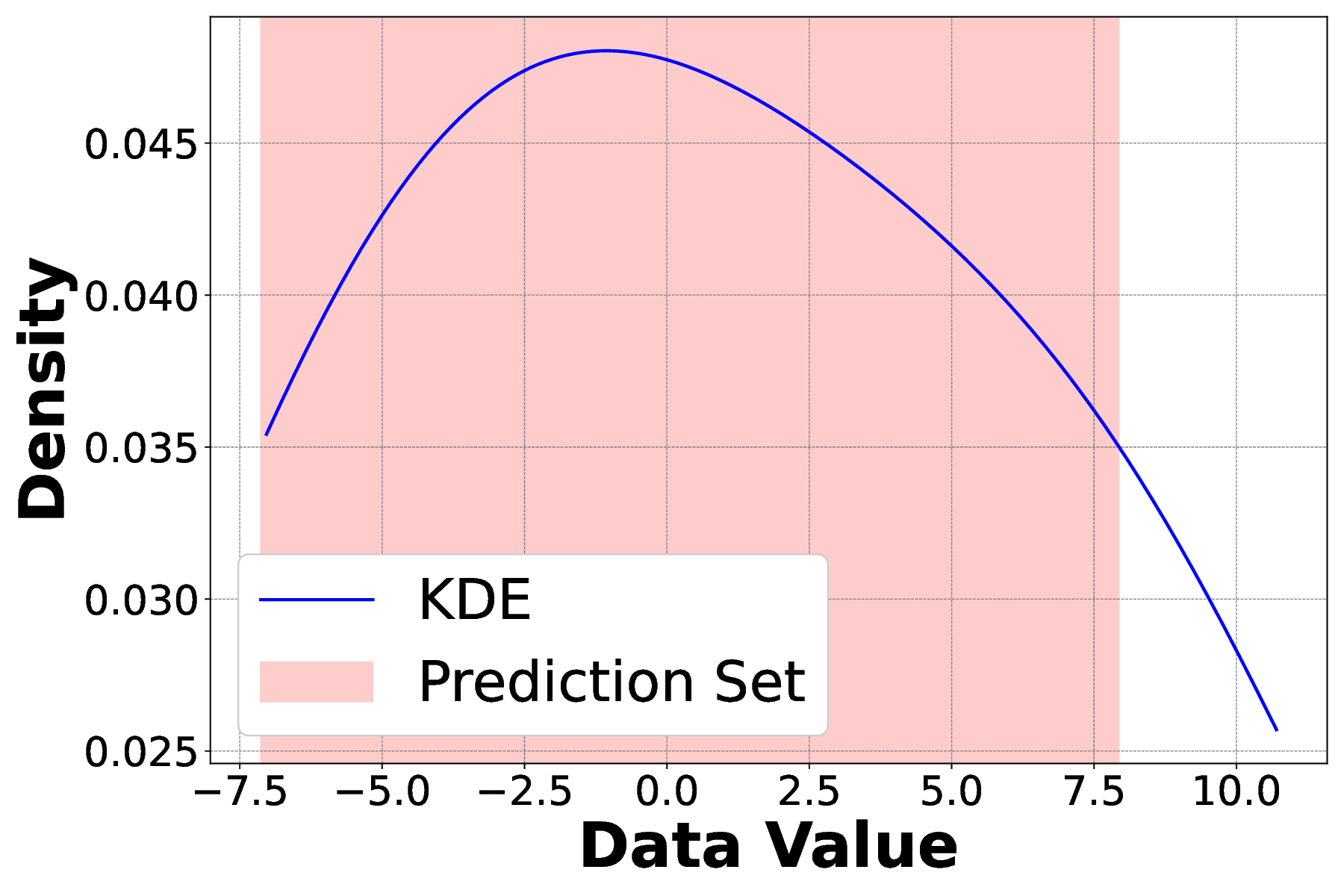}
        \caption{$\rho=5.0$}
    \end{subfigure}
    \caption{Prediction sets provided by the conformalized KDE using bandwidth $\rho=0.01,0.5,5.0$ on the mixture of Gaussians dataset.}
    \label{fig:mix_gaussian_intervals_KDE}
\end{figure}

\begin{figure}[H]
    \centering
        \includegraphics[width=0.45\textwidth]{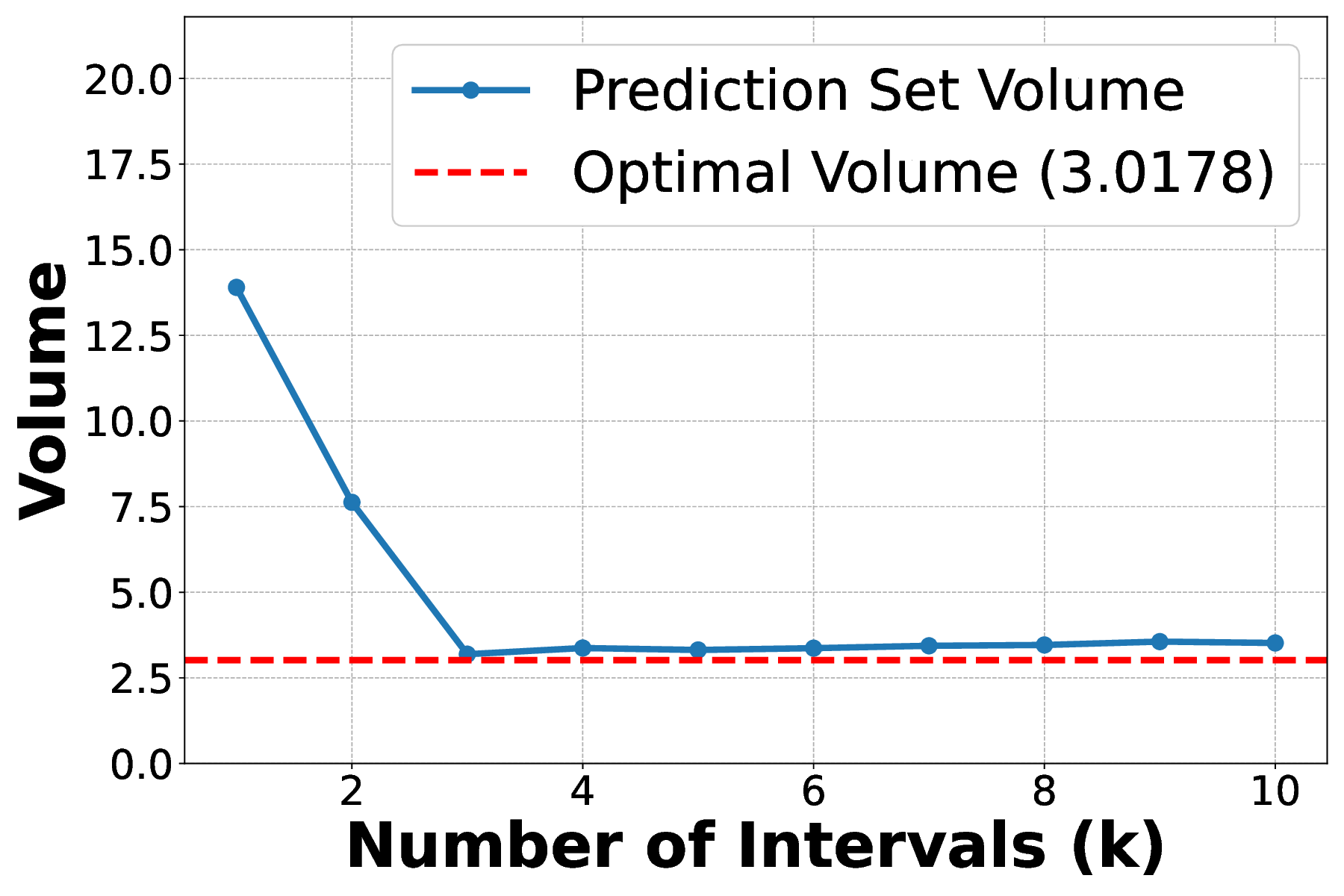}
    \hfill
        \includegraphics[width=0.45\textwidth]{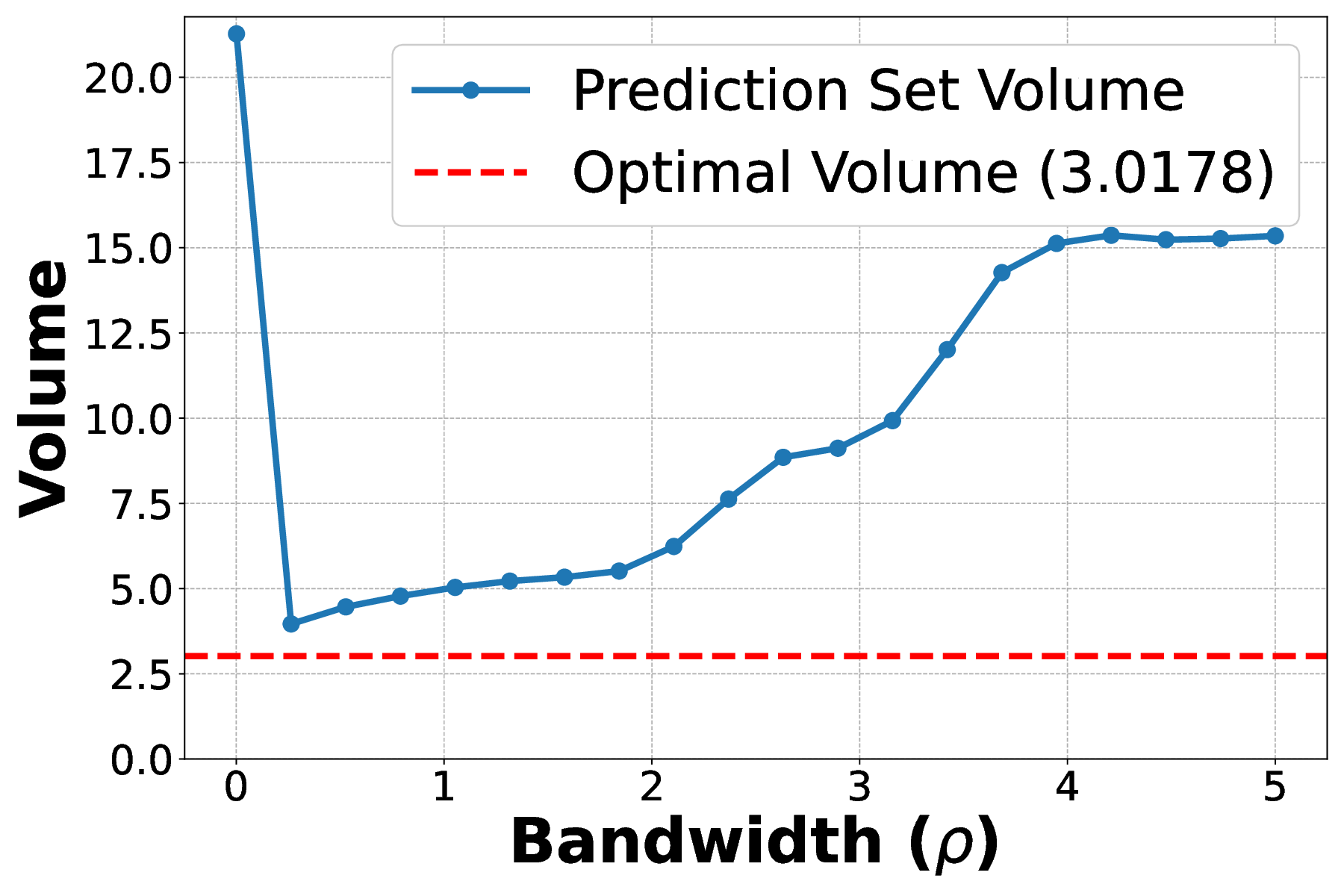}
    \caption{Volumes of prediction sets of the two methods on the mixture of Gaussians dataset (blue) and the optimal volume $\opt(P,0.8)=3.0178$ (red). The blue curves are computed by averaging $100$ independent experiments.}
    \label{fig:mix_gaussian_length}
\end{figure}

We report typical results of conformalized DP with $k\in\{2,3,6\}$ in Figure~\ref{fig:mix_gaussian_intervals_DP} and report those of conformalized KDE with $\rho\in\{0.01,0.5,5.0\}$ in Figure~\ref{fig:mix_gaussian_intervals_KDE}. The proposed method based on DP produces similar prediction sets close to optimal as long as $k\geq 3$ (Figure~\ref{fig:mix_gaussian_length}). This is because $\opt(P,0.8)=\opt_3(P,0.8)$ with $P$ being a Gaussian mixture of three components. In comparison, the results based on KDE are quite sensitive to the bandwidth choice, since different bandwidths lead to kernel density estimators with completely different numbers of modes. Figure~\ref{fig:mix_gaussian_length} shows that for the optimal choice of bandwidth around $0.5$, the KDE successfully identifies the three modes of the Gaussian mixture. However, even with the optimal bandwidth, the volume of the prediction set is in general still greater than that of the conformalized DP. This is partly because the three components of the Gaussian mixture do not have the same variance parameters, and thus cannot be optimally estimated by KDE with a single bandwidth.

\begin{figure}[H]
    \centering
    \begin{subfigure}{0.32\textwidth}
        \includegraphics[width=\textwidth]{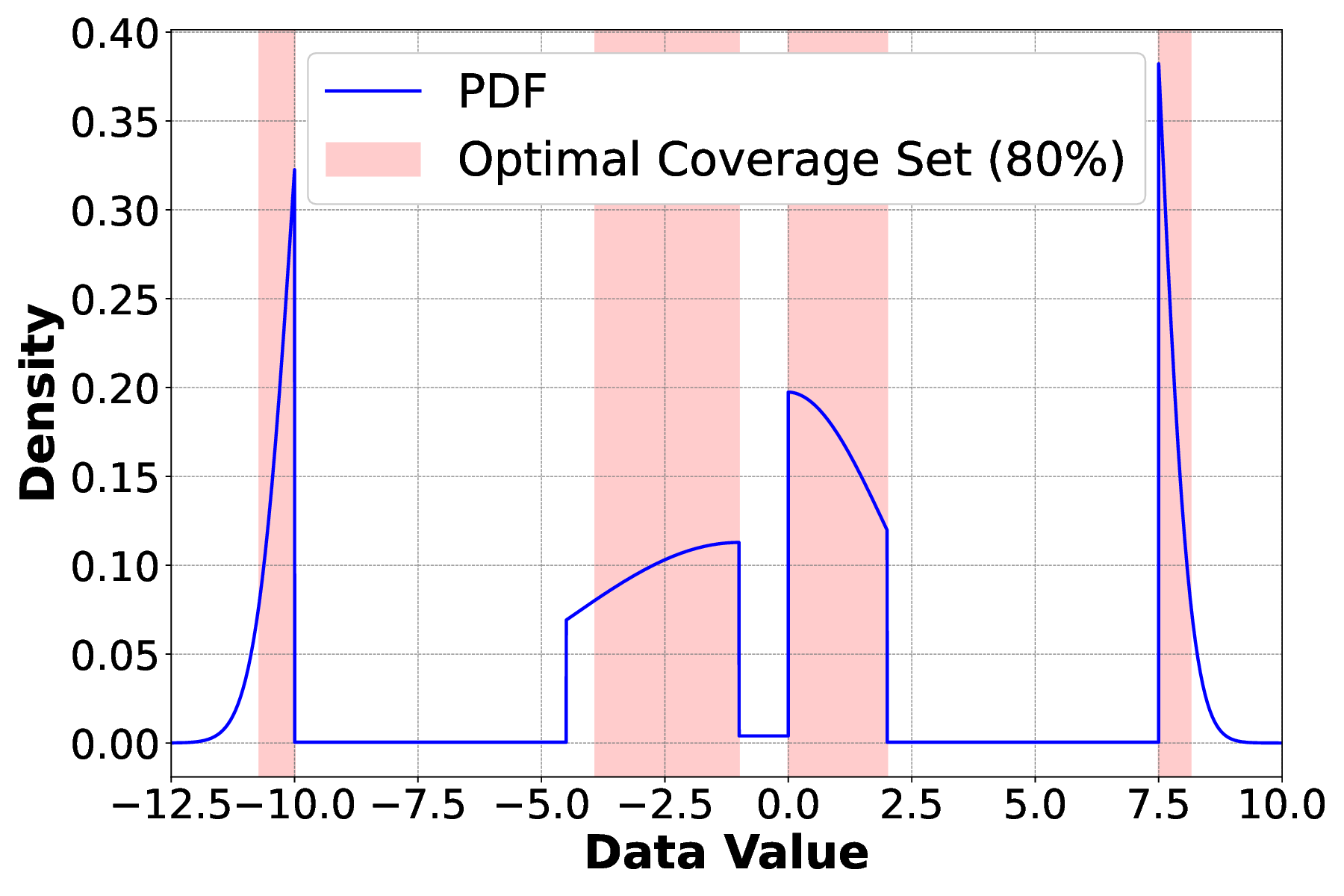}
        \caption{Optimal Coverage Set}
    \end{subfigure}
    \hfill
    \begin{subfigure}{0.32\textwidth}
        \includegraphics[width=\textwidth]{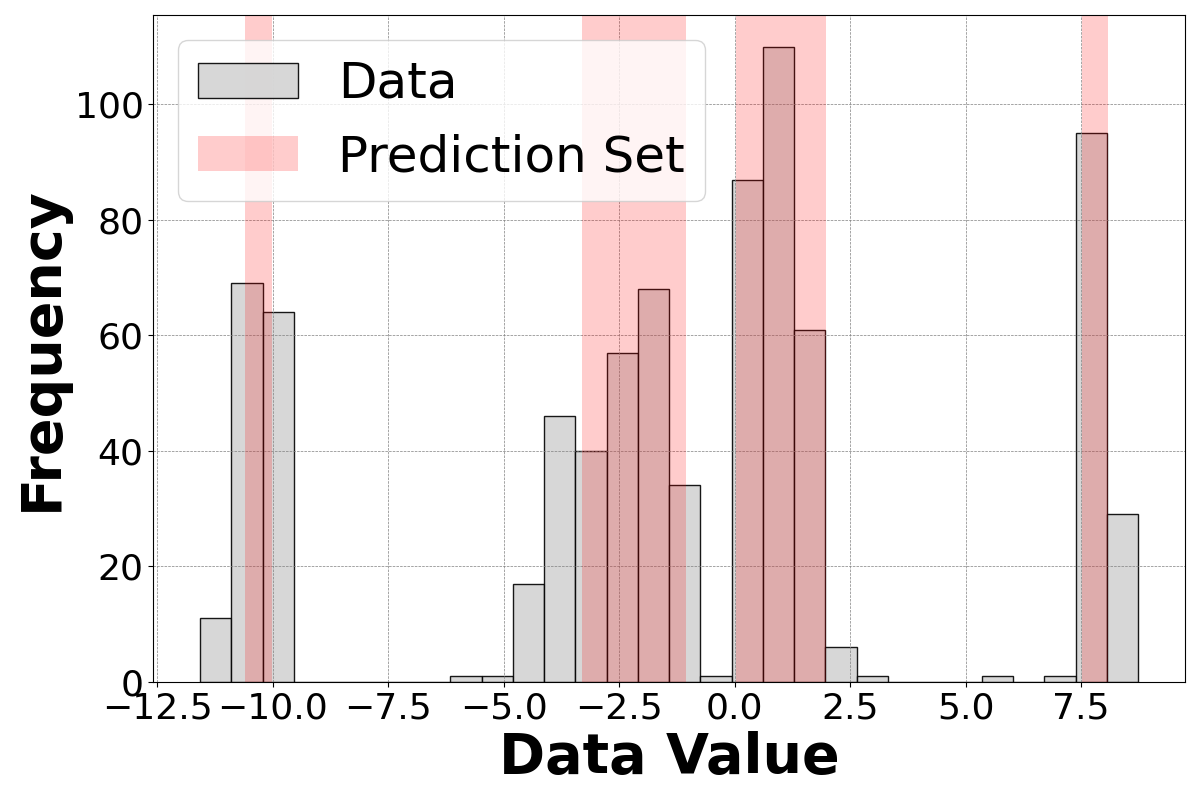}
        \caption{Conformalized DP with $k=4$}
    \end{subfigure}
    \hfill
    \begin{subfigure}{0.32\textwidth}
        \includegraphics[width=\textwidth]{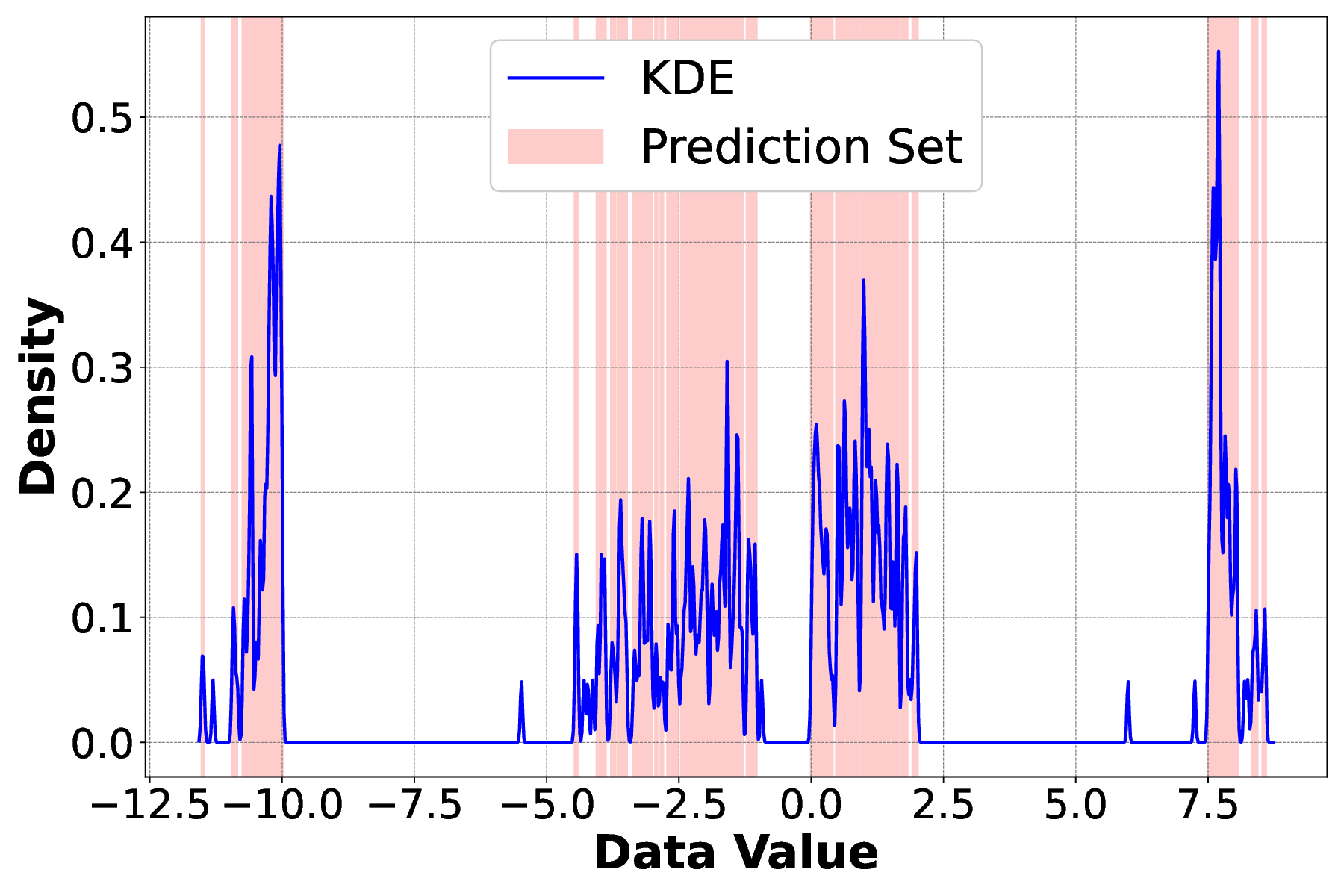}
        \caption{Conformalized KDE with $\rho=0.02$}
    \end{subfigure}
    \caption{(a) Density of The ReLU-Transformed Gaussian and prediction set with optimal volume; (b) Conformalized DP with $k=4$; (c) Conformalized KDE with $\rho=0.02$.}
    \label{fig:ReLU_example}
\end{figure}

\paragraph{ReLU-Transformed Gaussian:} The ReLU-Transformed Gaussian is generated according to $X_i = \sum_{j=1}^t a_j * \sigma(w_j*Z_i + b_j)$ with $Z_i\sim N(0,1)$. It includes the censored Gaussian as a special case. Here, we take $t=7$ and take a randomly generated set of coefficients. The resulting density function is plot in Figure~\ref{fig:ReLU_example} (a). The sample size and coverage probability are taken as $600$ and $1-\alpha=80\%$, respectively.

\begin{figure}[ht]
    \centering
        \includegraphics[width=0.49\columnwidth]{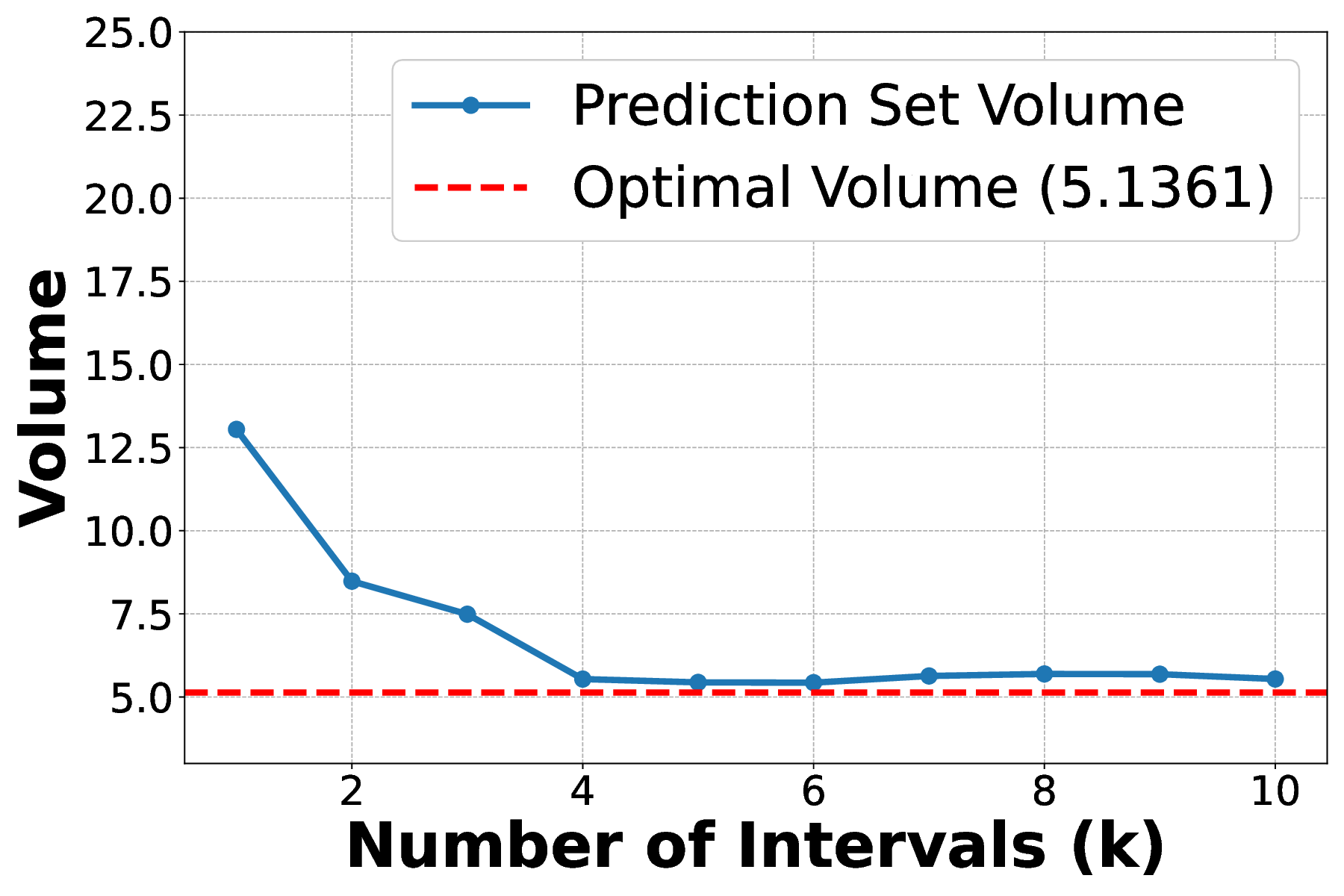}
    \hfill
        \includegraphics[width=0.49\columnwidth]{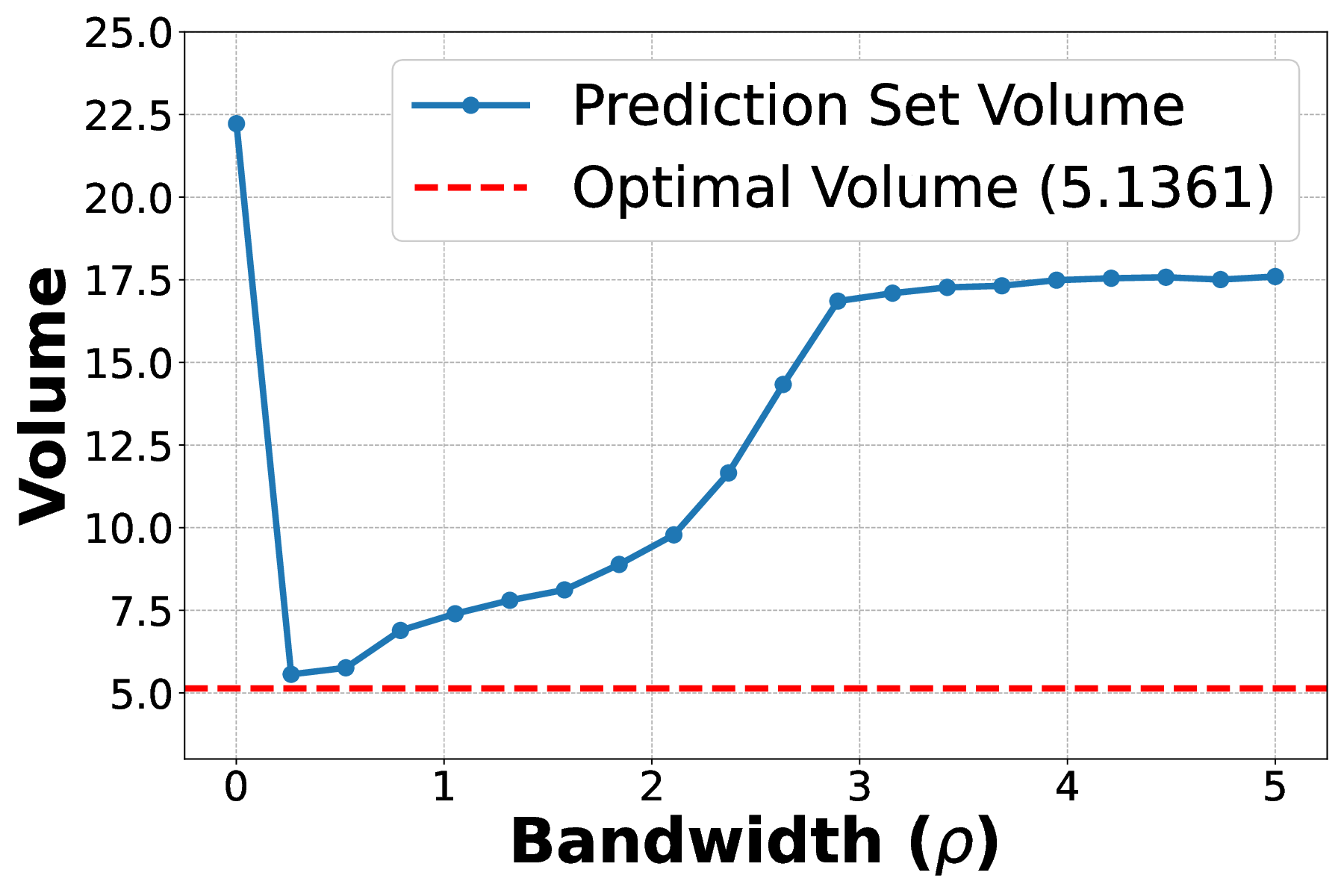}
    \caption{Volumes of prediction sets of the two methods on the ReLU-transformed Gaussian dataset (blue) and the optimal volume $\opt(P,0.8)=5.1361$ (red). The blue curves are computed by averaging $100$ independent experiments.}
    \label{fig:ReLU_length}
\end{figure}

Figure~\ref{fig:ReLU_example} also shows a typical prediction set produced by conformalized DP with $k=4$ and one produced by conformalized KDE with bandwidth $\rho=0.02$. Figure~\ref{fig:ReLU_length} gives a more thorough comparison. The proposed conformalized DP achieves near optimality when $k\geq 4$, since the distribution has $4$ modes. The KDE solutions are sensitive to the choice of bandwidth for this complicated distribution. 

\paragraph{Effects of Sample Sizes and Coverage Probabilites:}
Finally, we study the effects of sample sizes and coverage probabilities for the two methods. Specifically, we examine how the volume of the prediction set decays as the sample size increases and how it varies with different coverage probabilities. The experiments will be conducted with data generated from the following two distributions:
\begin{enumerate}
\item $\frac{1}{3}N(-6,0.0001)+\frac{1}{3}N(0,1)+\frac{1}{3}N(8,0.25)$.
\item The ReLU-Transformed Gaussian $Y_i = \sum_{j=1}^t a_j * \sigma(w_j*Z_i + b_j)$ with $Z_i\sim N(0,1)$, $t=7$ and coefficients are the same as in the previous experiment.
\end{enumerate}
For conformalized DP, we will set $k=3$ for the first distribution and $k=4$ for the second one to match the number of modes in the two cases. For conformalized KDE, since the method is sensitive to the choice of bandwidth, we will scan the bandwidth $\rho$ from $0.001$ to $0.2$, and only report the one with the smallest volume. We also benchmark the performances of the two methods by the optimal volume and a standard split conformal procedure with conformity score $q_{\rm standard}(y)=-\left|y-\frac{1}{n}\sum_{i=1}^nY_i\right|$.

\begin{figure}[htbp]
    \centering
        \includegraphics[width=0.45\textwidth]{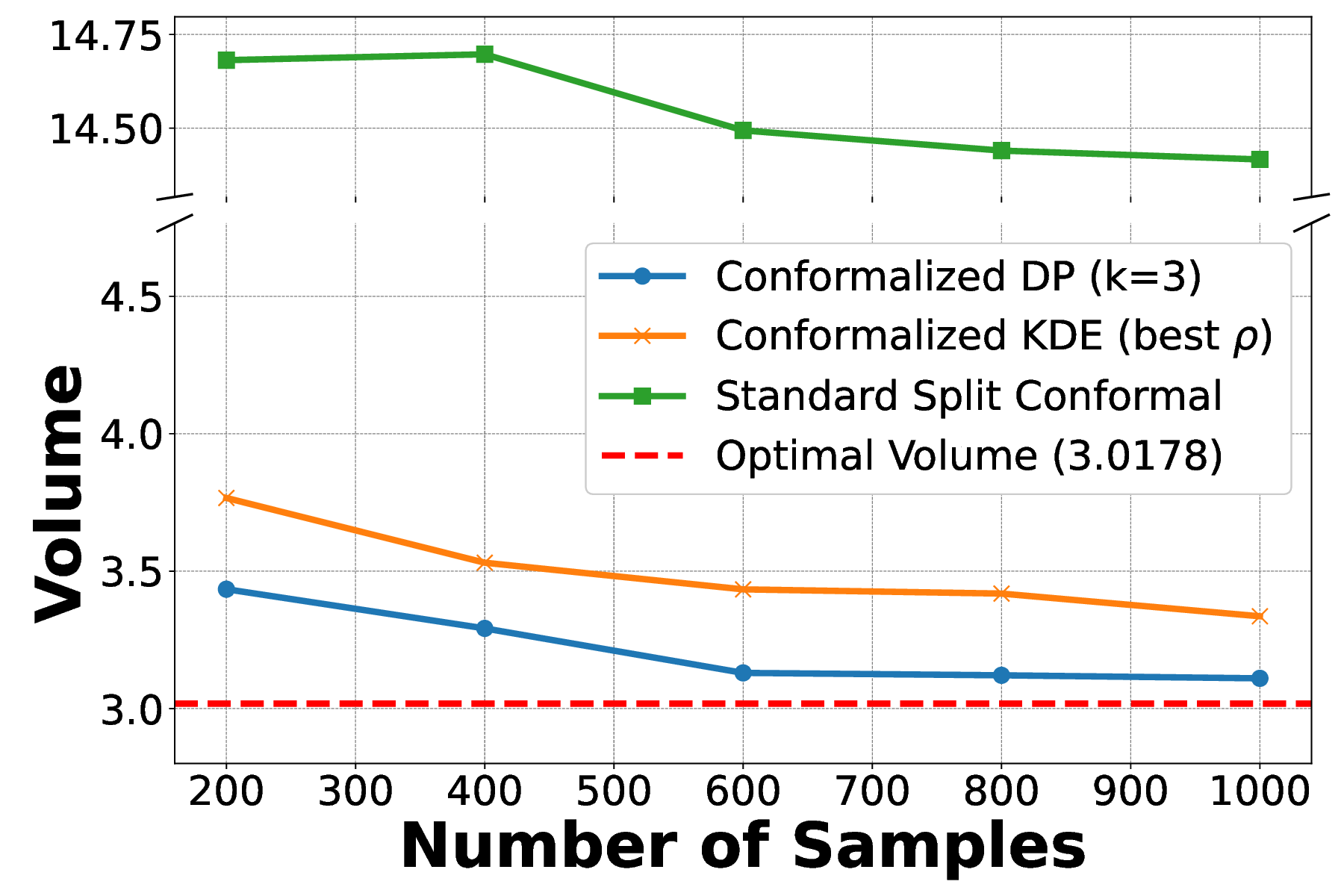}
    \hfill
        \includegraphics[width=0.45\textwidth]{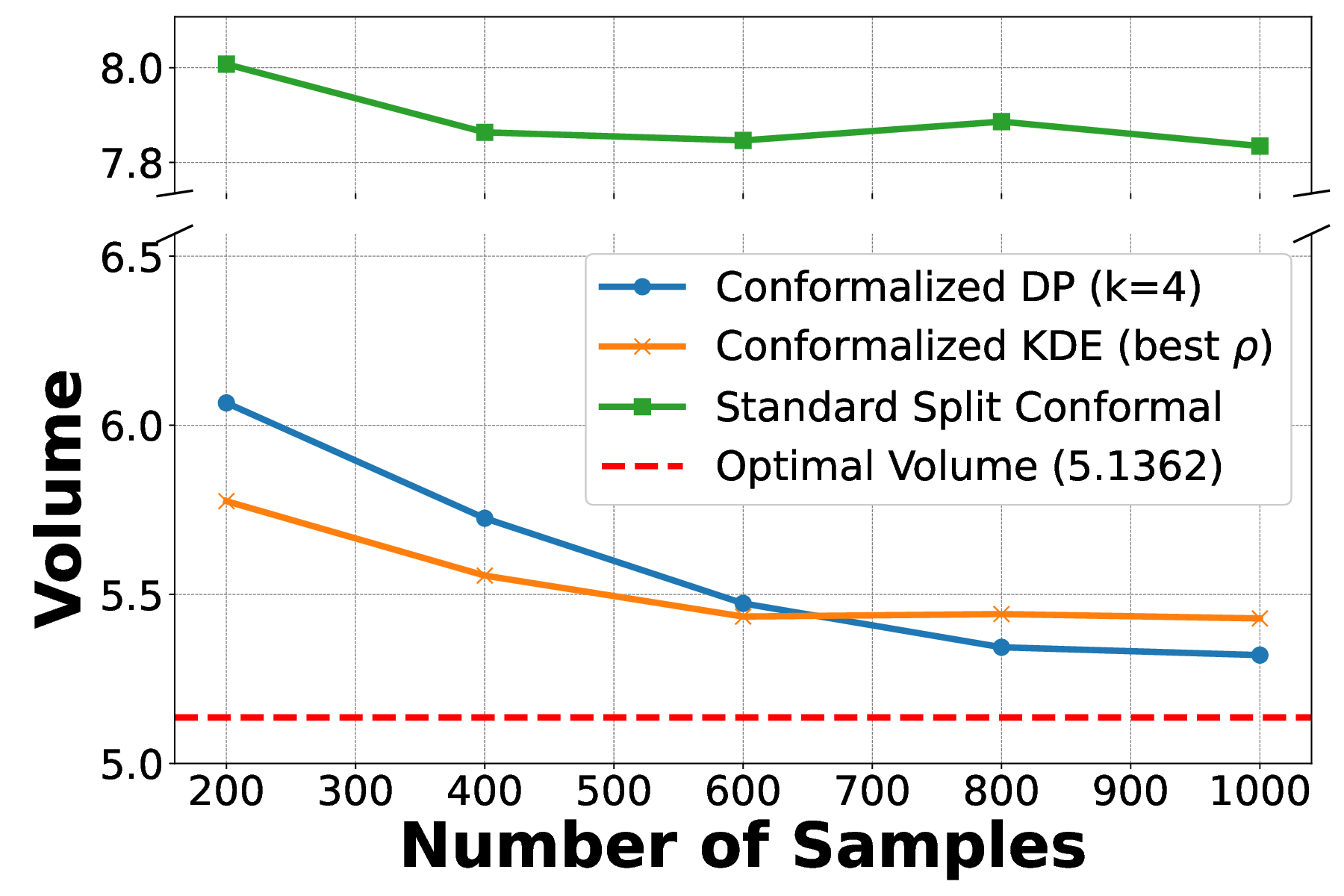}
    \caption{Volume of prediction set against sample size. Left: $\frac{1}{3}N(-6,0.0001)+\frac{1}{3}N(0,1)+\frac{1}{3}N(8,0.25)$. Right: ReLU-Transformed Gaussian. All curves are plotted by averaging results from $100$ independent experiments.}
    \label{fig:sample_size}
\end{figure}

\begin{figure}[H]
    \centering
        \includegraphics[width=0.45\textwidth]{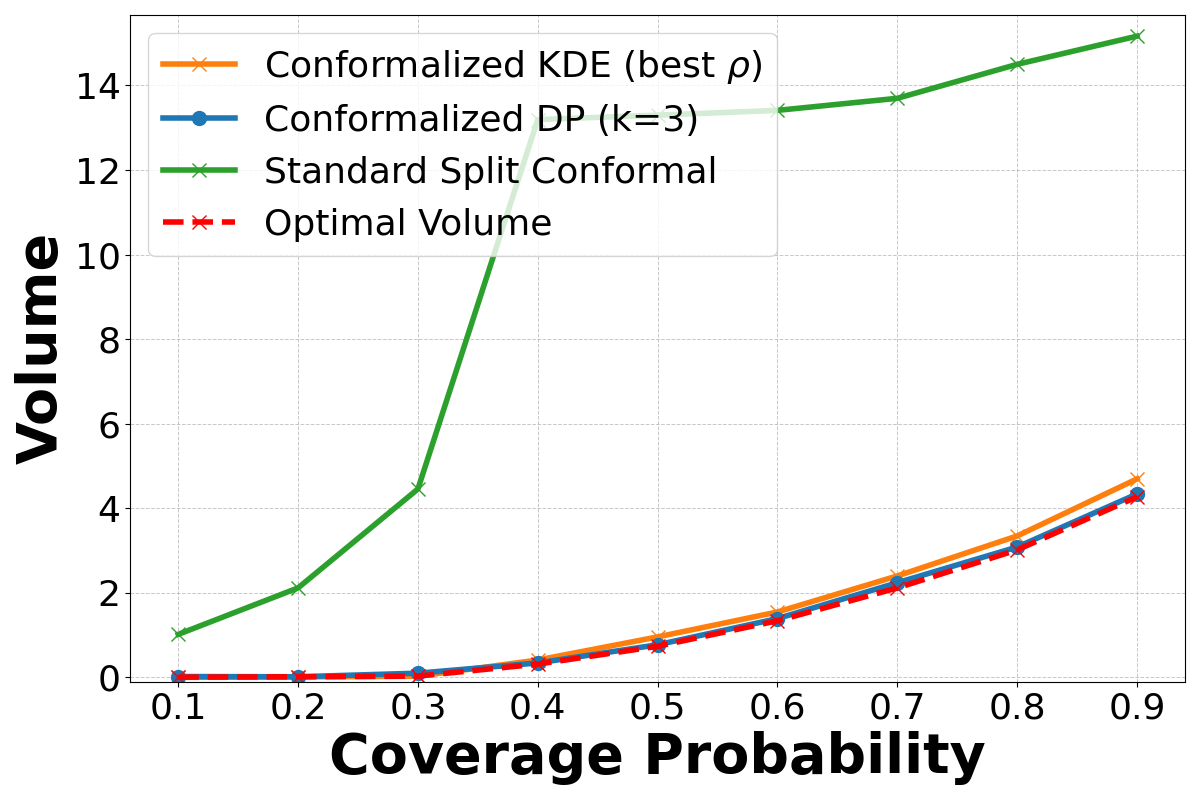}
    \hfill
        \includegraphics[width=0.45\textwidth]{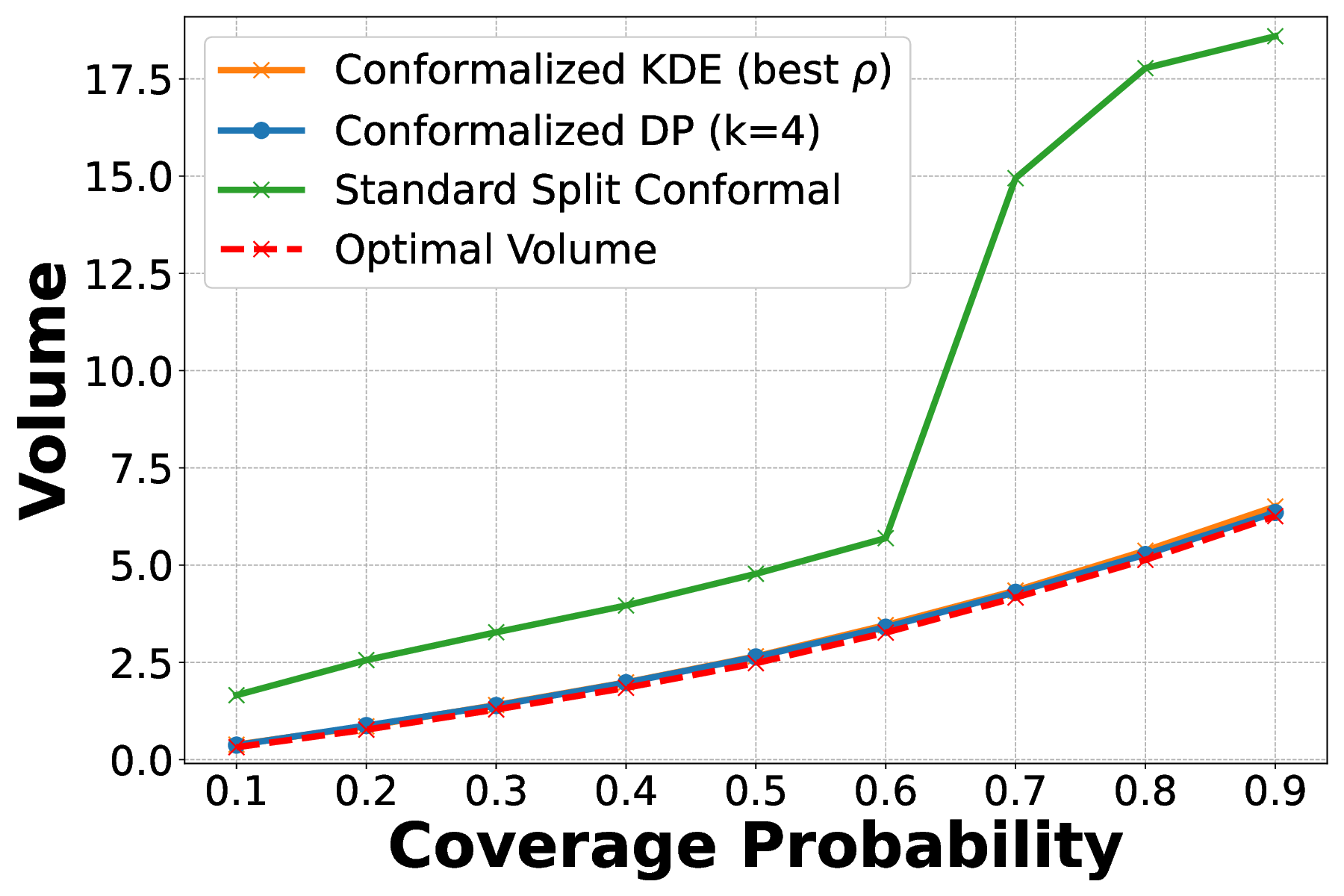}
    \caption{Volume of prediction set against coverage probability. Left: $\frac{1}{3}N(-6,0.0001)+\frac{1}{3}N(0,1)+\frac{1}{3}N(8,0.25)$. Right: ReLU-Transformed Gaussian. All curves are plotted by averaging results from $100$ independent experiments.}
    \label{fig:coverage}
\end{figure}

Figure \ref{fig:sample_size} shows the results with sample size ranging from $200$ to $1000$ with the coverage probability fixed by $1-\alpha=80\%$. Both conformalized DP and conformalized KDE produce smaller prediction sets as sample size increase. Even with the bandwidth optimally tuned for conformalized KDE, which is not feasible in practice, the proposed conformalized DP tends to achieve smaller volumes in most cases. In setting of the ReLU-Transformed Gaussian, we observe that the volume of conformalized KDE prediction set barely decreases after sample size $600$, since in this case density estimation is very hard for KDE.

Figure~\ref{fig:coverage} considers coverage probability ranging from $0.1$ to $0.9$, with sample size fixed at $600$. The conformalized DP constantly achieves smaller volume than the conformalized KDE even though the later is computed with optimally tuned bandwidth. This demonstrates the robustness of the conformalized DP in handling varying coverage requirements while maintaining efficiency in volume.

\subsection{Supervised Setting}
\label{sec:app-labeled-data}

In the supervised setting, we validate our results on the simulated datasets in \citet{romano2019conformalized} and \citet{izbicki2020flexible}.  We compare against the methods of Conformalized Quantile Regression (CQR) of \citet{romano2019conformalized} and Distributional Conformal Prediction (DCP) of \citet{chernozhukov2021distributional} and CD-split and HPD-split of \citet{izbicki2022cd}.

\begin{figure}
    \centering
    \includegraphics[width=0.49\linewidth]{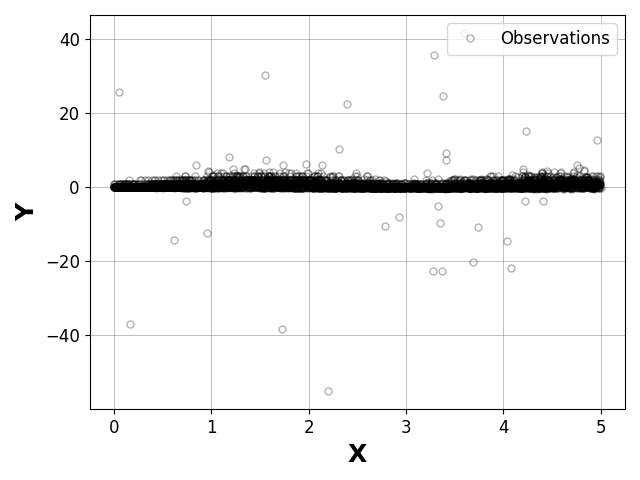}
    \caption{Simulated data of \citet{romano2019conformalized}, including outliers.}
    \label{fig:synthetic-zoom-out}
\end{figure}

\paragraph{Simulated Dataset \citep{romano2019conformalized}.} We first describe the simulated dataset in \citet{romano2019conformalized}. In this data, each one-dimensional predictor variable \(X_i\) is sampled uniformly from the range \([0, 5]\).  The response variable is then sampled according to 
\[Y_i \sim \mathrm{Pois}(\sin^2(X_i) + 0.1) + 0.03 ~X_i ~\varepsilon_{1, i} + 25 ~\mathbf{1}\{U_i < 0.01\} ~\varepsilon_{2, i},\]
where \(\mathrm{Pois}(\lambda)\) is the Poisson distribution with mean \(\lambda\), \(\epsilon_{1, i}\) and \(\epsilon_{2, i}\) are independent standard Gaussian noise, and \(U_i\) is drawn uniformly on the interval \([0, 1]\).  The first component of the distribution, \(\mathrm{Pois}(\sin^2(X_i) + 0.1)\), generates a distribution that is clustered around positive integer values of \(Y\), with variance that changes periodically in \(X\).  The second component of the distribution, \(0.03 ~X_i ~\varepsilon_{1, i}\), adds some additional variance to each of the integer centered clusters, where the magnitude of the variance increases with \(X\).  The final component, \(25 ~\mathbf{1}\{U_i < 0.01\} ~\varepsilon_{2, i}\), adds a small fraction of outliers to the distribution.   
We generate 2000 training examples, and 5000 test examples, as in the work of \cite{romano2019conformalized}.  The same subset of training and test examples are used in the illustration of each of these methods.  The set of test examples is visualized in Figure \ref{fig:synthetic-zoom-out}, with the full range of \(Y\) values including the outliers.  In the plots associated with our conformal output, we zoom in on the \(Y\) axis for readability, leaving the outliers off the chart.   

\paragraph{Simulated Dataset \citep{izbicki2020flexible}.}
We now describe the simulated dataset in \citet{izbicki2020flexible}. In this data, the predictor variables \( X = (X_1, \dots, X_d) \) with $d = 20$ dimensions are independently and uniformly sampled from the range \([-1.5, 1.5]\). The response variable \( Y \) is then generated according to the following bimodal conditional distribution:

$$
Y \mid X \sim 0.5 \mathcal{N}( f (X) - g (X), \sigma^2(X) ) + 0.5 \mathcal{N}( f (X) + g (X), \sigma^2(X) ).
$$

where the functions \( f(X) \), \( g(X) \), and \( \sigma^2(X) \) are defined as:
$$
f(X) = (X_1 - 1)^2 (X_1 + 1), \quad g(X) = 2 \mathbb{I}(X_1 \geq -0.5) \sqrt{X_1 + 0.5}, \quad \sigma^2(X) = \frac{1}{4} + |X_1|.
$$

Here, \( \mathcal{N}(\mu, \sigma^2) \) denotes a normal distribution with mean \( \mu \) and variance \( \sigma^2 \), and the indicator function \( \mathbb{I}(X_1 \geq -0.5) \) accounts for the bimodal nature of the data, introducing a piecewise behavior in the response variable. The first term \( f(X) \) captures a polynomial relationship with \( X_1 \), while the second term \( g(X) \) introduces an asymmetric bimodal effect depending on the value of \( X_1 \). The variance \( \sigma^2(X) \) increases linearly with \( |X_1| \), adding heteroscedasticity to the distribution.

\begin{figure}
    \centering
    \includegraphics[width=0.49\linewidth]{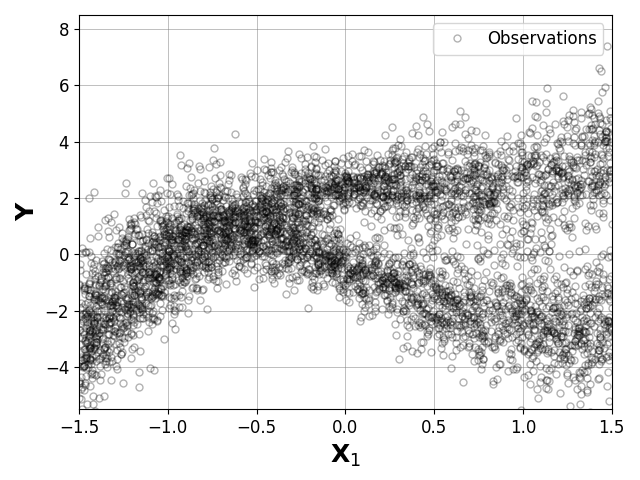}
    \caption{Simulated data from \citet{izbicki2020flexible}, illustrating the bimodal distribution of the response variable.}
    \label{fig:bimodal-synthetic}
\end{figure}

We generate $2000$ training examples and $5000$ test examples. The same training and test sets are used consistently across all experiments to ensure reproducibility. The test set is visualized in Figure~\ref{fig:bimodal-synthetic}, showcasing the full range of \( Y \) values, including the effects of bimodality and variance heterogeneity.

\paragraph{Methods.}  We compare our conformalized DP with the following methods: Conformalized Quantile Regression (CQR) of \citet{romano2019conformalized} and Distributional Conformal Prediction via Quantile Regression (DCP-QR) and Optimal Distributional Conformal Prediction via Quantile Regression (DCP-QR*) of \citet{chernozhukov2021distributional} and CD-split and HPD-split of \citet{izbicki2022cd}.  

We now describe the implementation of these methods. The compared methods, CQR, DCP-QR, DCP-QR*, and our conformalized DP rely on quantile regression. 
For simulated dataset~\citep{romano2019conformalized} with single dimensional predictor variable, we use the package \texttt{sklearn-quantile} \citep{sklearn-quantile} to implement the quantile regression, which implements the method of Quantile Regression Forests, due to \citet{Meinshausen06Quantile}.  
The CD-Split and HPD-Split methods require the conditional density estimation, which is achieved by the R package \texttt{FlexCoDE} \citep{izbicki2017converting}.
For simulated dataset \citep{izbicki2020flexible} with high dimensional predictor variables, the quantile regression by \texttt{sklearn-quantile} is not informative.
For CQR, DCP-QR, DCP-QR*, and our conformalized DP, we first use the R package \texttt{FlexCoDE} to generate the conditional density estimation and then integrate the conditional density estimation to get quantile regression and conditional CDF. 
The CD-Split and HPD-Split methods again use the conditional density estimation provided by \texttt{FlexCoDE}.
All methods are implemented within the split conformal framework, where the training data is randomly divided into two equal parts. Specifically, half of the data is allocated for model training, while the remaining half is used as the calibration set to ensure valid coverage guarantees.

For convenience, we will refer to the \(q\)th estimated quantile of \(Y\) given \(X = x\) as \(\widehat{Q}(q, x)\).  Some of the following methods use quantile regression to estimate the whole conditional c.d.f.\ of \(Y\) given \(X\), by estimating a set of quantiles from a fine grid.  This gives us an estimate of the conditional c.d.f., which gives us access to \(\widehat{F}(y\mid x)\), the inverse of \(\widehat{Q}\).  (That is, \(\widehat{F}(y\mid x) = q\), such that \(y = \widehat{Q}(q, x)\).  Since we only have \(\widehat{Q}\) for values of \(q\) in the grid, we set \(\widehat{F}(y\mid x)\) to be the smallest \(q\) in the grid such that \(y \le \widehat{Q}(q, x)\).)

\begin{itemize}
    \item Conformalized Quantile Regression (CQR), \cite{romano2019conformalized}:  This method fits a model to two quantiles of the data, \(q_{\mathrm{low}} = \frac{\alpha}{2}\) and \(q_{\mathrm{high}} = 1 - \frac{\alpha}{2}\).   On a new test example \(X_\mathrm{test}\), CQR uses the model to estimate the low and high quantile, and  the conformal procedure will output the interval 
    \[\left[\widehat{Q}(q_{\mathrm{low}}, X_\mathrm{test}) - b, ~\widehat{Q}(q_{\mathrm{high}}, X_\mathrm{test}) + b \right],\] 
    where \(b\) is a buffer value chosen in the calibration step of the conformal procedure to guarantee coverage.  

    \item Distributional Conformal Prediction via Quantile Regression (DCP-QR), \cite{chernozhukov2021distributional}:  In this framework, we assume access to a model \(\widehat{F}\) that can estimate the conditional c.d.f. of the distribution of \(Y\) given \(X\), which we estimate via quantile regression. Similar to CQR, we start with \(q_{\mathrm{low}} = \frac{\alpha}{2}\) and \(q_{\mathrm{high}} = 1 - \frac{\alpha}{2}\).  In DCP, instead of adding the buffer in the \(Y\) space, the buffer is added in the quantile space.  That is, on a new test example \(X_\mathrm{test}\), DCP will output the interval 
    \[\left[\widehat{Q}(q_{\mathrm{low}} - b, X_\mathrm{test}),  ~\widehat{Q}(q_{\mathrm{high}} + b, X_\mathrm{test})\right],\]
    where \(b\) is a buffer value chosen in the calibration step of the conformal procedure to guarantee coverage.  

    \item Optimal Distributional Conformal Prediction via Quantile Regression (DCP-QR*), \cite{chernozhukov2021distributional}:  The optimal DCP is very similar to DCP, except that \(q_{\mathrm{low}}\) and \(q_{\mathrm{high}}\) need not be symmetric around the median (\(q = \frac{1}{2}\)).  Instead, they are chosen to provide the minimum volume interval that achieves the desired coverage.  We note that the buffer is still applied symmetrically in the quantile space. That is, the lower quantile is lowered by some value \(b\), and the upper quantile is raised by the same value \(b\).

    \item CD-Split \citep{izbicki2020flexible, izbicki2022cd}: 
    This method provides prediction sets based on the conditional density estimation and a partitioning of the feature space. 
    The conformity score in CD-split is based on a conditional density estimator, which allows the method to approximate the highest predictive density (HPD) set. The feature space is partitioned based on the profile of the conditional density estimator, and the cut-off values are computed locally within each partition. This approach enables CD-split to achieve local and asymptotic conditional validity while providing more informative prediction sets, especially for multimodal distributions.

    \item HPD-Split \cite{izbicki2022cd}:
    The HPD-split method outputs prediction sets based on the highest predictive density (HPD) sets of the conditional density estimation. Unlike CD-split, which partitions the feature space, HPD-Split uses the conformity score based on the conditional CDF of the condition density estimator. Since this conditional CDF is independent of the feature variable, HPD-Split does not require the partition of the feature space and tuning parameters for that as in CD-Split. When the conditional density estimation is accurate, HPD-Split converges to the highest predictive density (HPD) sets.

    \item Conformalized Dynamic Programming, \(k = 1\) and \(k = 5\):  We implement a modification of the procedure described in this work.  In the unsupervised setting, we described the dynamic programming procedure that outputs the minimum volume set of \(k\) intervals that contain a desired fraction of samples.  In this setting, given a new test example \(X\), we do not have access to samples.  Instead, we have access to a grid of estimated quantiles of \(Y\) given \(X\).  We implement a version of the dynamic programming procedure that operates on this quantile grid instead of a set of points, to output the minimum volume set of \(k\) intervals that cover at least the desired probability mass.  We can also modify our greedy contraction and expansion procedures to provide a nested system of sets for different coverage levels.  
\end{itemize}

\paragraph{Discussion.}  The results of our experiments are illustrated in Figure \ref{fig:supervised_experiments}.  Our experiments show that Conformalized Quantile Regression (CQR) and Distributional Conformal Prediction via Quantile Regression (DCP-QR) perform approximately as well as each other on this dataset, achieving average volume 1.42 and 1.48 respectively (see Figures \ref{subfig:cqr}, \ref{subfig:dcp-qr}).  

Optimal Distributional Conformal Prediction via Quantile Regression (DCP-QR*) achieves a significant improvement over DCP-QR on this data, achieving average volume 1.29 (see Figure \ref{subfig:dcp-qrstar}).  This is due to the fact that the distribution of \(Y\) values is not symmetric around, or peaked at the median \(Y\) value.  Thus, DCP-QR suffers a disadvantage, because it outputs intervals that are centered around the median in quantile space, and does not take into account the relative volumes of the quantiles in \(Y\) space.  DCP-QR* on the other hand, is able to take advantage of the fact that, for this data, quantiles close to 0 have very low volume, and output intervals that use these quantiles.   

While DCP-QR* uses information about the relative volume of the quantiles to choose \(q_{\mathrm{low}}\) and \(q_{\mathrm{high}}\), which define the output intervals before conformalization, it does not take the volume into account during the conformalization step.  Expanding the interval by a buffer value \(b\) that is small in quantile space, can lead to a large difference in \(Y\) space, increasing the volume of the output interval significantly.  For example in Figure \ref{subfig:dcp-qrstar}, the intervals for \(X\) just larger than 4 stretch very far into the negative \(Y\) region, as a small adjustment in quantile space is a large adjustment in \(Y\) space.  

This issue is avoided by our Conformalized Dynamic Programming (Conformalized DP) method with greedy expansion and contraction for \(k = 1\) interval.  Before conformalization, the interval output by Conformalized DP and DCP-QR* is the same: it is the volume optimal interval that achieves a given coverage according to the estimated c.d.f..  However, our method takes the relative volume of different quantiles into account in the conformalization step, and avoids the issue of expanding the interval in quantile space in directions that add too much volume in \(Y\) space.  This allows the method to achieve an improved average volume of 1.14 (see Figure \ref{subfig:conformalizeddpk-1}).  

\begin{figure}[H]
    \centering
    \begin{subfigure}{0.49\textwidth}
        \includegraphics[width=\textwidth]{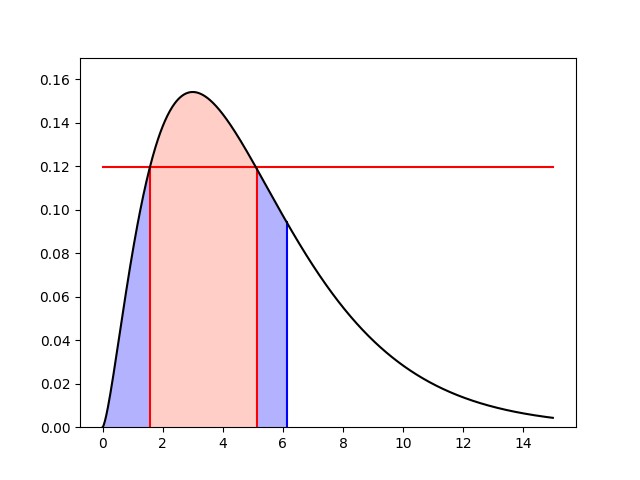}
        \caption{When expanding from the red region, coverage 0.5, to a region of coverage 0.7, DCP-QR* chooses the blue region with additional volume 2.56.}
    \end{subfigure}
    \hfill
    \begin{subfigure}{0.49\textwidth}
        \includegraphics[width=\textwidth]{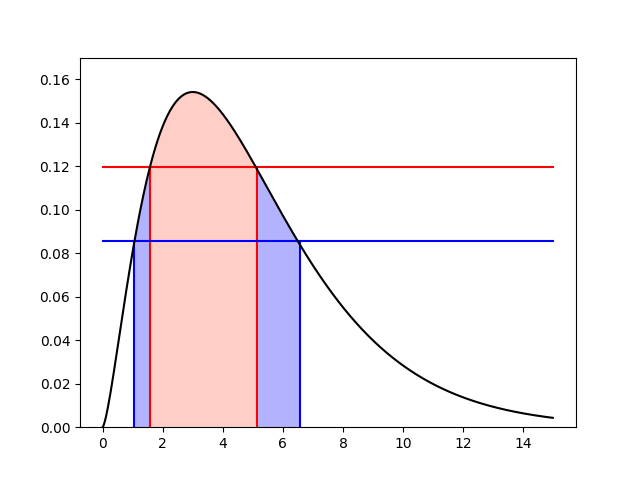}
        \caption{When expanding from the red region, coverage 0.5, to a region of coverage 0.7, Conformalized DP chooses the blue region with additional volume 1.96.}
    \end{subfigure}
    \caption{We illustrate the difference between DCP-QR* and Conformalized DP for \(k = 1\), on the example where the estimated conditional distribution of \(Y\) for a new \(X_\mathrm{test}\) is \(\chi^2(5)\).  We plot the intervals that are chosen by the methods against the p.d.f. of the estimated distribution.}
    \label{fig:volume-aware-example}
\end{figure}

An illustration of this issue is given in Figure \ref{fig:volume-aware-example}.  Suppose that for a new test example \(X\), the estimated conditional distribution of \(Y\) is skewed.  (In this illustration it is \(\chi^2(5)\).)  Suppose that our target coverage was 0.5, and in the calibration phase we are required to expand coverage to 0.7.  Both ConformalizedDP and DCP-QR* will start by calculating the minimum volume interval that captures 0.5 of the probability mass.  In this case it is the red region from \(x = 1.58\) to \(x = 5.14\) (i.e., the set of \(x\) such that \(f(x) > 0.12\), where \(f(x)\) is the p.d.f. of the distribution).  Then, each method must expand this interval to capture 0.7 of the probability mass.  DCP-QR* does this by adding two blue regions, each of which capture an additional 0.1 probability mass.  This results in expanding the interval significantly to the right, even though the density is low.  ConformalizedDP takes the volume (i.e., density) into account when expanding the interval, and produces the minimum volume interval that captures 0.7 of the distribution (i.e., the set of \(x\) such that \(f(x) > 0.085\)), in this example.  (We note that the expansion and contraction procedure of ConformalizedDP does not always result in the volume optimal prediction set for the adjusted coverage, only the original target coverage.  However, in this case, since the distribution is unimodal and \(k = 1\), we do indeed recover the volume optimal set even for the adjusted coverage.)  

Finally, we also implement Conformalized DP with \(k = 5\) intervals.  This allows us to fit to the multimodal shape of the \(Y\) data, and achieve a much lower average volume of 0.45 (see Figure \ref{subfig:conformalizeddpk-5}). 

\begin{figure}[H]
    \centering
    \begin{subfigure}{0.49\textwidth}
        \includegraphics[width=\textwidth]{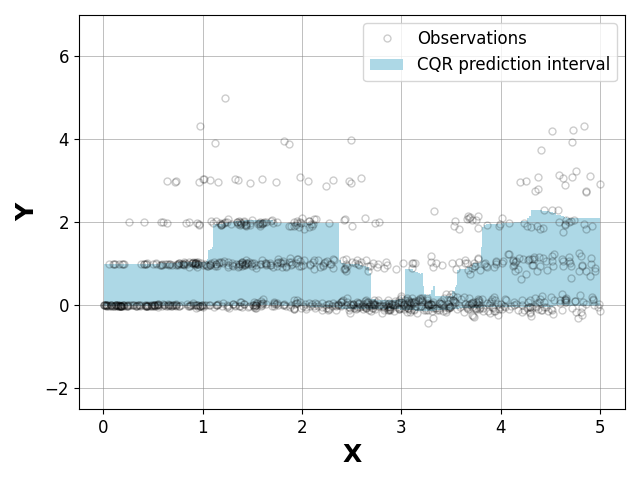}
        \caption{Conformalized Quantile Regression (CQR), \cite{romano2019conformalized}, achieves average volume 1.42 and empirical coverage 70.62\%.}
        \label{subfig:cqr}
    \end{subfigure}
\end{figure}

\begin{figure}[H]\ContinuedFloat
    \begin{subfigure}{0.49\textwidth}
        \includegraphics[width=\textwidth]{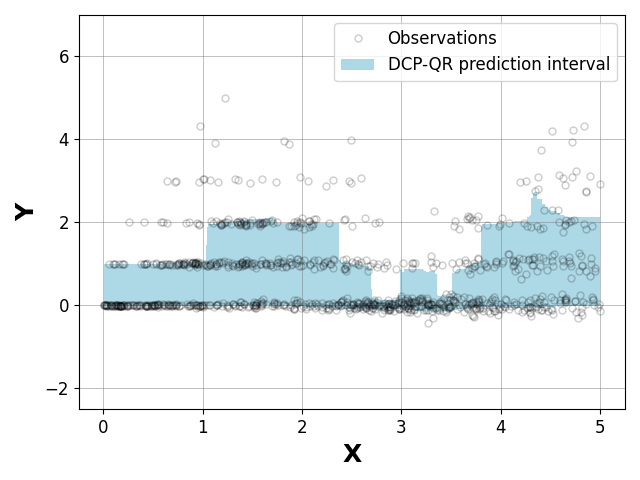}
        \caption{Distributional Conformal Prediction (DCP), \cite{chernozhukov2021distributional}, achieves average volume 1.48 and empirical coverage 71.6\%.}
        \label{subfig:dcp-qr}
    \end{subfigure}
    \hfill
    \begin{subfigure}{0.49\textwidth}
        \includegraphics[width=\textwidth]{figures/dcp-qrstar-synthetic.png}
        \caption{Optimal Distributional Conformal Prediction (DCP-QR*), \cite{chernozhukov2021distributional}, achieves average volume 1.29 and empirical coverage 71.06\%.}
        \label{subfig:dcp-qrstar}
    \end{subfigure}
    \hfill
    \begin{subfigure}{0.49\textwidth}
    \includegraphics[width=\textwidth]{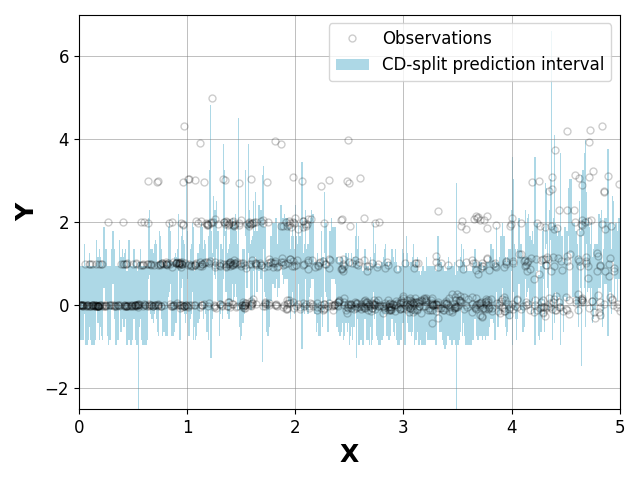}
        \caption{CD-split Conformal Prediction, \cite{izbicki2022cd}, achieves average volume 1.83 and empirical coverage 69.94\%.}
        \label{subfig:CD-split}
    \end{subfigure}
    \hfill
    \begin{subfigure}{0.49\textwidth}
    \includegraphics[width=\textwidth]{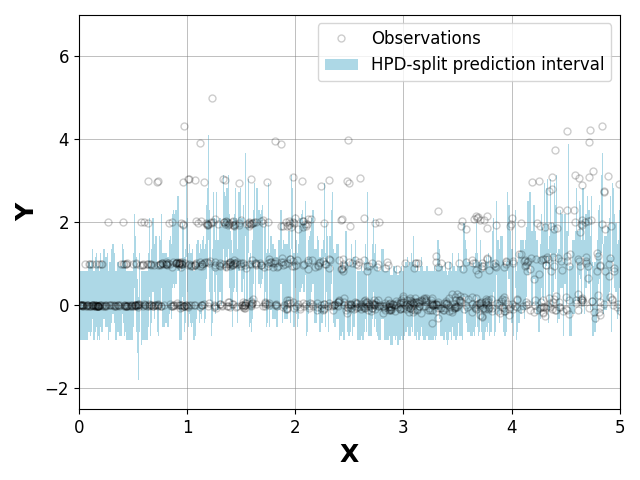}
        \caption{HPD-split Conformal Prediction, \cite{izbicki2022cd}, achieves average volume 1.75 and empirical coverage 69.44\%.}
        \label{subfig:HPD-split}
    \end{subfigure}
    \hfill
    \begin{subfigure}{0.49\textwidth}
    \includegraphics[width=\textwidth]{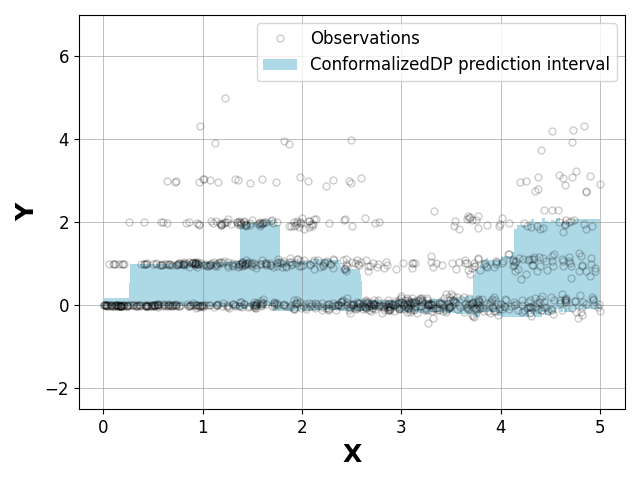}
        \caption{Conformalized Dynamic Programming ($k = 1$), achieves average volume 1.14 and empirical coverage 74.04\%.}
        \label{subfig:conformalizeddpk-1}
    \end{subfigure}
    \hfill
    \begin{subfigure}{0.49\textwidth}
        \includegraphics[width=\textwidth]{figures/conformalizedDP-k5-synthetic.png}
        \caption{Conformalized Dynamic Programming ($k = 5$), achieves average volume 0.45 and empirical coverage 72.36\%.} 
        \label{subfig:conformalizeddpk-5}
    \end{subfigure}
    \caption{Comparison of supervised conformal prediction methods on simulated data from \cite{romano2019conformalized}.  All results are for a target coverage of 0.70.}
    \label{fig:supervised_experiments}
\end{figure}

\begin{figure}[H] 
    \centering
    \begin{subfigure}{0.49\textwidth}
    \includegraphics[width=\textwidth]{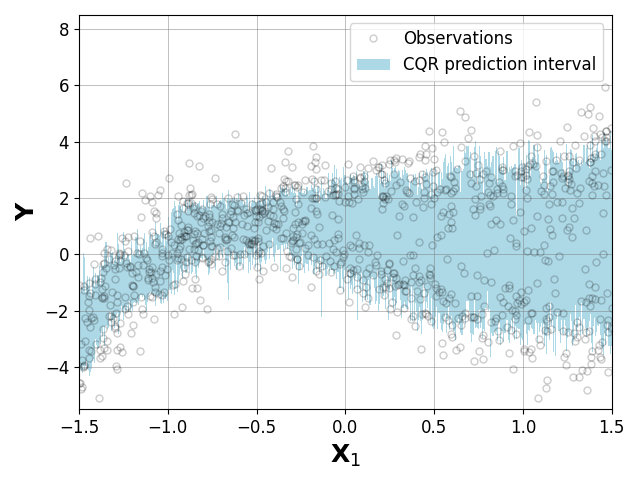}
        \caption{CQR, achieves average volume 4.10 and empirical coverage 71.54\%.}
        \label{subfig:bimodal_cqr}
    \end{subfigure}\\
    \hfill
    \begin{subfigure}{0.49\textwidth}
    \includegraphics[width=\textwidth]{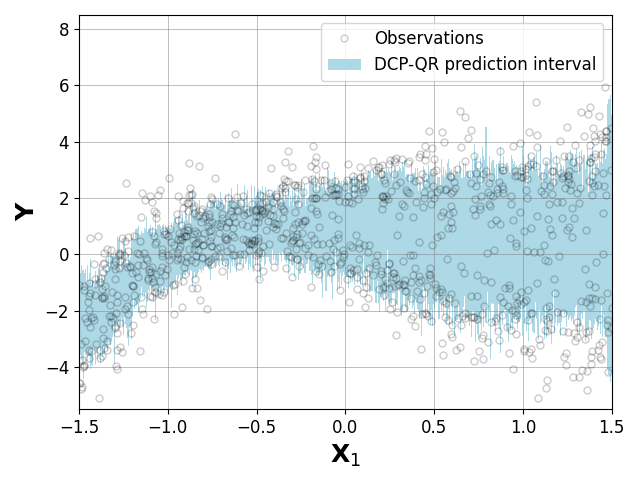}
        \caption{DCP-QR, achieves average volume 4.04 and empirical coverage 70.85\%.}
        \label{subfig:bimodal_dcp_qr}
    \end{subfigure}
    \hfill
    \begin{subfigure}{0.49\textwidth}
        \includegraphics[width=\textwidth]{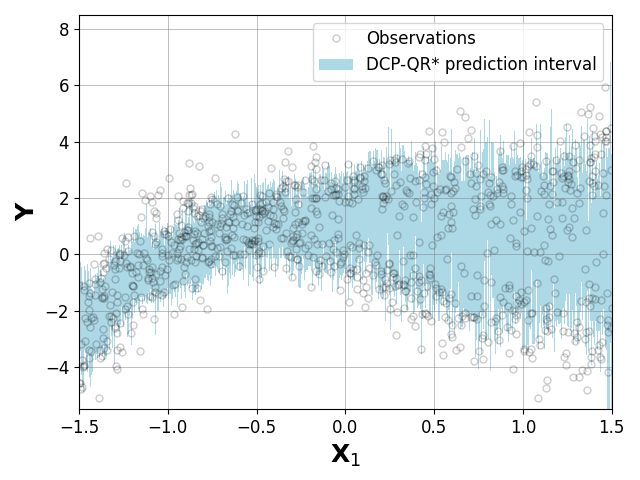}
        \caption{DCP-QR*, achieves average volume 4.05 and empirical coverage 69.66\%.} 
        \label{subfig:bimodal_dcp_qrstar}
    \end{subfigure}
    \hfill
    \begin{subfigure}{0.49\textwidth}
    \includegraphics[width=\textwidth]{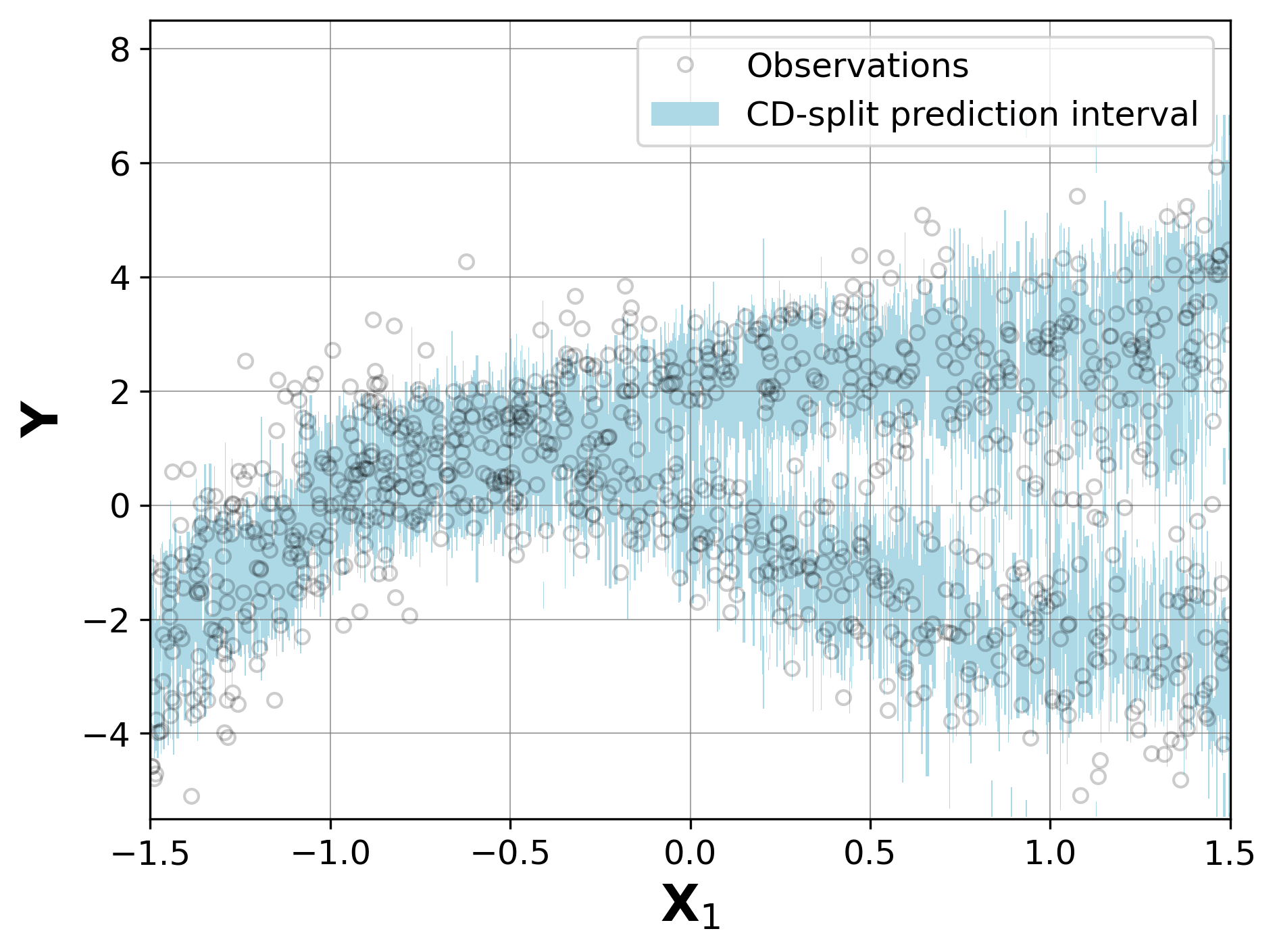}
        \caption{CD-Split, achieves average volume 3.69 and empirical coverage 69.86\%.}
        \label{subfig:bimodal_cd_split}
    \end{subfigure}
    \hfill
    \begin{subfigure}{0.49\textwidth}
        \includegraphics[width=\textwidth]{figures/bimodal_hpd_split.png}
        \caption{HPD-Split, achieves average volume 3.60 and empirical coverage 69.64\%.} 
        \label{subfig:bimodal_hpd_split}
    \end{subfigure}
\end{figure}

\begin{figure}[H]\ContinuedFloat
    \begin{subfigure}{0.49\textwidth}
    \includegraphics[width=\textwidth]{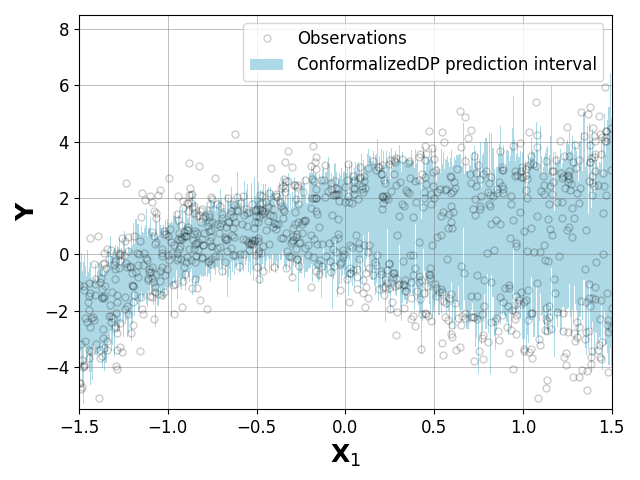}
        \caption{Conformalized Dynamic Programming ($k = 1$), achieves average volume 4.00 and empirical coverage 68.98\%.}
        \label{subfig:bimodal_dp_k1}
    \end{subfigure}
    \hfill
    \begin{subfigure}{0.49\textwidth}
        \includegraphics[width=\textwidth]{figures/bimodal_dp_k2_c70.png}
        \caption{Conformalized Dynamic Programming ($k = 2$), achieves average volume 3.55 and empirical coverage 69.42\%.} 
        \label{subfig:bimodal_dp_k2}
    \end{subfigure}
    \caption{Comparison of supervised conformal prediction methods on simulated data from \cite{izbicki2020flexible}.  All results are for a target coverage of 0.70.}
    \label{fig:supervised_experiments_bimodal}
\end{figure}

\section{KDE Optimality Implies DP Optimality}\label{sec:DPvsKDE}

Suppose a distribution $P$ on $\mathbb{R}$ admits a density function $p$. The kernel density estimator depending on $k$ i.i.d. samples $Z_1,\cdots,Z_k$ is defined by
\begin{equation}
p_k(y)=\frac{1}{k\rho}\sum_{j=1}^kK\left(\frac{y-Z_j}{\rho}\right),\label{eq:KDE-k}
\end{equation}
where $K(\cdot)$ is a standard Gaussian kernel and $\rho$ is a bandwidth parameter. The conformal prediction method by \cite{Lei2013DistributionFreePS} is based on the idea that the level set of $p_k$ is close to that of $p$ as long as $p_k$ is close to $p$. In this section, we will show that as long as $p_k$ is close to $p$, the dynamic programming also finds a prediction set whose volume is nearly optimal compared with the level set of $p$. This implies that DP always requires no stronger assumption to achieve volume optimality.

The existence of a KDE close to $p$ can be even weakened into the following assumption.
\begin{assumption}\label{as:kde}
For any positive integer $k$, there exists some $\epsilon_k>0$ and some Gaussian mixture $P_k=\sum_{j=1}^k w_jN(\mu_j,\sigma_j^2)$ such that $\TV(P_k,P)\leq\epsilon_k$.
\end{assumption}
In particular, the KDE (\ref{eq:KDE-k}) based on $k$ samples is a special case of the Gaussian mixture, given that Gaussian kernel is used. Though the characterization of the closeness between $p_k$ and $p$ is through $\ell_{\infty}$ norm by \cite{Lei2013DistributionFreePS}, similar error bounds also apply to the $\ell_1$ norm, which is the total variation distance. For example, suppose $P$ has bounded support and the Hölder smoothness is $\beta\in (0,2]$. Then, one can take $\epsilon_k=\widetilde{\Theta}\left(k^{-\frac{\beta}{2\beta+1}}\right)$ with an appropriate choice of the bandwidth, where $\widetilde{\Theta}$ hides some logarithmic factor of $k$.

\begin{theorem}\label{thm:DPvsKDE}
Consider i.i.d. observations $Y_1,\cdots,Y_n,Y_{n+1}$ generated by some distribution $P$ on $\mathbb{R}$ that satisfies Assumption \ref{as:kde}. For any $\alpha,\delta,\gamma\in(0,1)$ such that $\delta\gg\sqrt{\frac{k+\log n}{n}}$ and $\gamma+2\delta+2\epsilon_k<\alpha$, let $\widehat{C}_{\rm DP}\in\calC_k$ be the output of Algorithm \ref{alg:dp} with coverage level $1-\alpha+\delta$ and slack $\gamma$. Then, with probability at least $1-\delta$, we have
\begin{enumerate}
\item $\mathbb{P}\left(Y_{n+1}\in \widehat{C}_{\rm DP}\mid Y_1,\cdots,Y_n\right)\geq 1-\alpha$;
\item $\vol(\widehat{C}_{\rm DP})\leq \opt(P,1-\alpha+\gamma+2\delta+2\epsilon_k)$.
\end{enumerate}
\end{theorem}

The result of Theorem \ref{thm:DPvsKDE} can also be conformalized as in Section \ref{sec:cdp}, so that the restricted volume optimality $\opt_k(P,\cdot)$ in Theorem \ref{thm:unsupervised} can be strengthened to $\opt(P,\cdot)$ without restriction whenever $P$ satisfies Assumption \ref{as:kde}, which, in particular, includes the situation where the density of $P$ can be well estimated by KDE.

The volume sub-optimality given by Theorem \ref{thm:DPvsKDE} is $\gamma+2\delta+2\epsilon_k$. When the distribution $P$ is $\beta$-smooth, the sub-optimality is of order $\widetilde{\Theta}\left(\sqrt{\frac{k}{n}}+k^{-\frac{\beta}{2\beta+1}}\right)$ by taking $\epsilon_k=\widetilde{\Theta}\left(k^{-\frac{\beta}{2\beta+1}}\right)$, $\delta=\widetilde{\Theta}\left(\sqrt{\frac{k}{n}}\right)$, and $\gamma$ sufficiently small. Thus, optimizing this bound over $k$ leads to the rate $\widetilde{\Theta}\left(n^{-\frac{\beta}{4\beta+1}}\right)$. In comparison, the KDE achieves a faster rate $\widetilde{\Theta}\left(n^{-\frac{\beta}{2\beta+1}}\right)$ \citep{Lei2013DistributionFreePS} for smooth densities. This is actually a technical artifact by specializing Assumption \ref{as:kde} to the KDE (\ref{eq:KDE-k}). In fact, when the density of $P$ is $\beta$-smooth, it is well known that Assumption \ref{as:kde} is satisfied with a better $\epsilon_k=\widetilde{\Theta}(k^{-\beta})$ \citep{ghosal2007posterior,kruijer2010adaptive}, which then leads to the volume sub-optimality $\widetilde{\Theta}\left(\sqrt{\frac{k}{n}}+k^{-\beta}\right)$ that leads to the near optimal rate $\widetilde{\Theta}\left(n^{-\frac{\beta}{2\beta+1}}\right)$ with $k=\widetilde{\Theta}(n^{\frac{1}{2\beta+1}})$.

\subsection{Proof of Theorem \ref{thm:DPvsKDE}}

We first state a lemma that shows that a level set of a Gaussian mixture with $k$ components must belong to the class $\calC_k$.
\begin{lemma}\label{lem:gmmode}
For a Gaussian mixture $P_k=\sum_{j=1}^kw_jN(\mu_j,\sigma_j^2)$ and any $\alpha\in(0,1)$,
$$\opt_k(P_k,1-\alpha)=\opt(P_k,1-\alpha).$$
\end{lemma}
The proof of the lemma will be given in Appendix~\ref{sec:proof_of_lemmas}.
Now we are ready to state the proof of Theorem \ref{thm:DPvsKDE}.
\begin{proof}[Proof of Theorem \ref{thm:DPvsKDE}]
By Proposition \ref{thm:DP}, we know that $\widehat{C}_{\rm DP}$ satisfies $\mathbb{P}_n(\widehat{C}_{\rm DP})\geq 1-\alpha+\delta$ and $\vol(\widehat{C}_{\rm DP})\leq \opt_k(\mathbb{P}_n,1-\alpha+\delta+\gamma)$. The condition on $\delta$ implies that $\sup_{C \in \calC_k}|\mathbb{P}_n(C)-P(C)|\leq\delta$ with probability at least $1-\delta$ \citep{devroye2001combinatorial}. Therefore, the coverage probability is
$$\mathbb{P}\left(Y_{n+1}\in \widehat{C}_{\rm DP}\mid Y_1,\cdots,Y_n\right)\geq \mathbb{P}_n(\widehat{C}_{\rm DP})-\delta \geq 1-\alpha,$$
and the volume can be bounded by
\begin{eqnarray*}
\vol(\widehat{C}_{\rm DP}) &\leq & \opt_k(\mathbb{P}_n,1-\alpha+\delta+\gamma) \\
&\leq& \opt_k(P,1-\alpha+2\delta+\gamma) \\
&\leq& \opt_k(P_k,1-\alpha+2\delta+\gamma+\epsilon_k) \\
&=& \opt(P_k,1-\alpha+2\delta+\gamma+\epsilon_k) \\
&\leq& \opt(P,1-\alpha+2\delta+\gamma+2\epsilon_k),
\end{eqnarray*}
where the identity above is by Lemma \ref{lem:gmmode}.
\end{proof}

\section{Additional Proofs}\label{sec:proofs}

\subsection{Proof of Theorem \ref{thm:impossibility}}

The proof relies on the following technical lemma, whose proof will be given in Appendix~\ref{sec:proof_of_lemmas}.

\begin{lemma}\label{lem:tv}
For any $\delta,\epsilon>0$ and any integer $n>0$, there exists some distribution $\Pi$ supported on
$$\mathcal{P}_{\epsilon}=\left\{P:\text{supp}(P)\subset[0,1],\TV(P,\lambda)\geq 1-\epsilon\right\},$$
such that $\TV\left(\lambda^n,\int P^n \mathrm{d}\Pi\right)\leq \delta$. 
\end{lemma}

Now we are ready to state the proof of Theorem \ref{thm:impossibility}.
\begin{proof}[Proof of Theorem \ref{thm:impossibility}]
By Lemma \ref{lem:tv}, there exist $\Pi_{n,\delta}$ and $\Pi_{n+1,\delta}$ supported on $\mathcal{P}_{\delta}$, such that $\TV\left(\lambda^n, \int P^n \mathrm{d}\Pi_{n,\delta}\right)\leq\delta$ and $\TV\left(\lambda^{n+1}, \int P^{n+1} \mathrm{d}\Pi_{n+1,\delta}\right)\leq\delta$. 

Since $P^{n+1}(Y_{n+1}\in \widehat{C}(Y_1,\cdots,Y_n))\geq 1-\alpha$ for all $P$, we have
$$\int P^{n+1}(Y_{n+1}\in \widehat{C}(Y_1,\cdots,Y_n))\mathrm{d}\Pi_{n+1,\delta}\geq 1-\alpha.$$
By $\TV\left(\lambda^{n+1}, \int P^{n+1} \mathrm{d}\Pi_{n+1,\delta}\right)\leq\delta$, we have
\begin{align*}
    \E_{Y_1,\dots,Y_n \sim \lambda^n}(\lambda(\widehat{C}(Y_1,\cdots,Y_n)))
    =\lambda^{n+1}(Y_{n+1}\in \widehat{C}(Y_1,\cdots,Y_n))\geq 1-\alpha-\delta.
\end{align*}
By $\TV\left(\lambda^n, \int P^n \mathrm{d}\Pi_{n,\delta}\right)\leq\delta$, we have
$$\int \E_{Y_1,\dots,Y_n \sim P^n}\lambda(\widehat{C}(Y_1,\cdots,Y_n)) \mathrm{d}\Pi_{n,\delta}\geq 1-\alpha-2\delta.$$
Then, there must exists some $P\in\text{supp}(\Pi_{n,\delta})\subset\mathcal{P}_{\delta}$, such that
$$\E_{Y_1,\dots,Y_n \sim P^n}\lambda(\widehat{C}(Y_1,\cdots,Y_n))\geq 1-\alpha-2\delta.$$
The fact that $P\in \mathcal{P}_{\delta}$ implies $\TV(P,\lambda)\geq 1-\delta$. 
By the definition of total variation, there exists some set $B$ such that $P(B)-\lambda(B)\geq 1-\delta$, which implies $P(B)\geq 1-\delta$ and $\lambda(B)\leq\delta$. Therefore, $\opt(P,1-\delta)\leq \delta$. 

We finally have for any $\varepsilon \in (0,\alpha)$, the expected volume of prediction set is
\begin{eqnarray*}
\E_{Y_1,\dots,Y_n \sim P^n}\vol(\widehat{C}(Y_1,\cdots,Y_n)) 
&\geq& 1-\alpha-2\delta \\
&\geq& \delta \\
&\geq& \opt(P,1-\delta) \\
&\geq& \opt(P,1-\alpha + \varepsilon),
\end{eqnarray*}
as long as $\delta$ is sufficiently small so that $\delta<\min\{(1-\alpha)/3, \alpha-\epsilon\}$. The proof is complete.
\end{proof}

\subsection{Proof of Theorem \ref{thm:unsupervised}}

Theorem \ref{thm:unsupervised} is a special case of Theorem \ref{thm:supervised} in the setting where $F(\cdot\mid x)$ does not depend on $x$ and $\widehat{F}(\cdot\mid x)$ is defined as the empirical CDF of $Y_1,\cdots,Y_n$. Then, Assumption \ref{ass:F} is automatically satisfied by a standard VC dimension bound \citep{devroye2001combinatorial}.

\subsection{Proof of Theorem \ref{thm:supervised}}

We will prove the three properties of Theorem \ref{thm:supervised} separately. We note that the marginal coverage property holds without Assumptions \ref{as:ne-su} and \ref{ass:F}. It is a standard consequence of applying the split conformal framework, but we still include a proof here for completeness.

\begin{proof}[Proof of Theorem \ref{thm:supervised} (marginal coverage)]
By the construction of $\widehat{C}_{\rm DCP-DP}(X_{2n+1})$, we have
\begin{align*}
    \Pr\left(Y_{2n+1} \in \widehat{C}_{\rm DCP-DP}(X_{2n+1})\right) = \Pr \left(q(Y_{2n+1},X_{2n+1})\geq q_{\lfloor(n+1)\alpha\rfloor}\right).
\end{align*}
Since the conformity score $q$ is constructed from $\widehat{F}(\cdot \mid \cdot)$, it is independent from the second half of the data. 
Thus, $q(Y_{2n+1},X_{2n+1})$ is exchangeable with $q(Y_{n+1},X_{n+1}),\cdots, q(Y_{2n},X_{2n})$, which implies the desired conclusion by the definition of $q_{\lfloor(n+1)\alpha\rfloor}$.
\end{proof}

Next, we establish the conditional coverage property. We need the following property of the conformity score that is computed based on a nested system.
\begin{lemma}\label{lem:nest}
For any $j\in[m+1]$, $y\in S_j(x)$ if and only if $q(y,x)\geq m-j+1$, where the set $S_{m+1}(x)$ is defined as $\mathbb{R}$.
\end{lemma}

The proof of the lemma will be given in Appendix~\ref{sec:proof_of_lemmas}.

\begin{proof}[Proof of Theorem \ref{thm:supervised} (approximate conditional coverage)]
We first note that Assumption \ref{ass:F} implies
\begin{equation}
\mathbb{E}\|\widehat{F}(\cdot \mid X_{2n+1})-F(\cdot \mid X_{2n+1})\|_{k,\infty}\leq 2\delta.\label{eq:f-expectation}
\end{equation}
We use $\mathcal{F}_{2n}$ to denote the $\sigma$-field generated by the random variables $(X_1,Y_1),\cdots, (X_{2n},Y_{2n})$. Let $\mathbb{E}_{X_{2n+1}}$ and $\mathbb{E}_{\mathcal{F}_{2n}}$ be the expectation operators under the marginal distributions of $X_{2n+1}$, and of $(X_1,Y_1),\cdots, (X_{2n},Y_{2n})$, respectively. Then, we have
\begin{align*}
\mathbb{P}\left(q(Y_{2n+1},X_{2n+1})\geq q_{\lfloor(n+1)\alpha\rfloor}\right) = \E_{\mathcal{F}_{2n}}\E_{X_{2n+1}}\mathbb{P}\left(q(Y_{2n+1},X_{2n+1}) 
\geq q_{\lfloor(n+1)\alpha\rfloor}| X_{2n+1},\mathcal{F}_{2n}\right). 
\end{align*}
By Lemma \ref{lem:nest}, $q(Y_{2n+1},X_{2n+1})\geq q_{\lfloor(n+1)\alpha\rfloor}$ is equivalent to $Y_{2n+1}\in S_{\widehat{j}}(X_{2n+1})$ for some $\widehat{j}$ measurable with respect to $\mathcal{F}_{2n}$. Then, we have
\begin{align*}
\E_{\mathcal{F}_{2n}}\E_{X_{2n+1}}\mathbb{P}\left(q(Y_{2n+1},X_{2n+1})\geq q_{\lfloor(n+1)\alpha\rfloor}|X_{2n+1},\mathcal{F}_{2n}\right) &=\mathbb{E}_{\mathcal{F}_{2n}}\mathbb{E}_{X_{2n+1}}\mathbb{P}\left(Y_{2n+1}\in S_{\widehat{j}}(X_{2n+1})|X_{2n+1},\mathcal{F}_{2n}\right) \\
&=\mathbb{E}_{\mathcal{F}_{2n}}\mathbb{E}_{X_{2n+1}}\int_{S_{\widehat{j}}(X_{2n+1})} \mathrm{d} F(y\mid X_{2n+1}). 
\end{align*}
Since $S_{\widehat{j}}(X_{2n+1}) \in \calC_k$, by (\ref{eq:f-expectation}), we have 
\begin{align*}
\mathbb{E}_{\mathcal{F}_{2n}}\mathbb{E}_{X_{2n+1}}\int_{S_{\widehat{j}}(X_{2n+1})} \mathrm{d} F(y\mid X_{2n+1}) 
\leq& \mathbb{E}_{\mathcal{F}_{2n}}\mathbb{E}_{X_{2n+1}}\int_{S_{\widehat{j}}(X_{2n+1})} \mathrm{d}\widehat{F}(y \mid X_{2n+1}) \\
&+ \mathbb{E}\|\widehat{F}(\cdot \mid X_{2n+1})-F(\cdot \mid X_{2n+1})\|_{k,\infty} \\
\leq& \mathbb{E}_{\mathcal{F}_{2n}}\mathbb{E}_{X_{2n+1}}\int_{S_{\widehat{j}}(X_{2n+1})} \mathrm{d}\widehat{F}(y \mid X_{2n+1}) + 2\delta.
\end{align*}
By Assumption \ref{as:ne-su}, we have $\int_{S_{\widehat{j}}(X_{2n+1})}\mathrm{d}\widehat{F}(y \mid X_{2n+1})=\frac{\widehat{j}}{m}$, which is independent of $X_{2n+1}$, since $\widehat{j}$ measurable with respect to $\mathcal{F}_{2n}$. Thus, we have 
\begin{align*}
\mathbb{E}_{\mathcal{F}_{2n}}\mathbb{E}_{X_{2n+1}}\int_{S_{\widehat{j}}(X_{2n+1})} \mathrm{d}\widehat{F}(y \mid X_{2n+1}) + 2\delta 
=\mathbb{E}_{\mathcal{F}_{2n}}\int_{S_{\widehat{j}}(X_{2n+1})} \mathrm{d}\widehat{F}(y \mid X_{2n+1}) + 2\delta.
\end{align*}
By Assumption~\ref{ass:F}, we have with probability at least $1-\delta$,
\begin{align*}
\mathbb{E}_{\mathcal{F}_{2n}}\int_{S_{\widehat{j}}(X_{2n+1})} \mathrm{d}\widehat{F}(y \mid X_{2n+1}) + 2\delta 
\leq& \mathbb{E}_{\mathcal{F}_{2n}}\int_{S_{\widehat{j}}(X_{2n+1})} \mathrm{d}F(y \mid X_{2n+1}) +3\delta \\
=& \mathbb{P}\left(q(Y_{2n+1},X_{2n+1})\geq q_{\lfloor(n+1)\alpha\rfloor} \mid X_{2n+1}\right) + 3\delta.
\end{align*}
Therefore, with probability at least $1-\delta$, the approximate conditional coverage holds
\begin{align*}
\mathbb{P}\left(q(Y_{2n+1},X_{2n+1})\geq q_{\lfloor(n+1)\alpha\rfloor} \mid X_{2n+1}\right) 
\geq \mathbb{P}\left(q(Y_{2n+1},X_{2n+1})\geq q_{\lfloor(n+1)\alpha\rfloor}\right) - 3\delta \geq 1-\alpha -3\delta.
\end{align*}
\end{proof}

Finally, we prove the last property on volume optimality.
\begin{proof}[Proof of Theorem \ref{thm:supervised} (conditional restricted volume optimality)]
By Assumption \ref{as:ne-su}, there exists some $j^*\in[m]$, such that
$$\mathbb{E}\int_{S_{j^*}(X_{2n+1})} \mathrm{d}\widehat{F}(y \mid X_{2n+1})\geq 1-\alpha+\frac{1}{n}+3\delta.$$
Assumption \ref{ass:F} implies that 
\begin{eqnarray*}
\mathbb{P}\left(Y_{2n+1}\in S_{j^*}(X_{2n+1})\right) 
&=& \mathbb{E}\int_{S_{j^*}(X_{2n+1})} \mathrm{d}F(y\mid X_{2n+1}) \\
&\geq& \mathbb{E}\int_{S_{j^*}(X_{2n+1})} \mathrm{d}\widehat{F}(y\mid X_{2n+1})\\
&& - \mathbb{E}\|\widehat{F}(\cdot \mid X_{2n+1})-F(\cdot\mid X_{2n+1})\|_{k,\infty}\\
&\geq& 1-\alpha+\frac{1}{n}+\delta.
\end{eqnarray*}
By Hoeffding's inequality and the condition on $\delta$, we have
$$\frac{1}{n}\sum_{i=n+1}^{2n}\mathbb{I}\{Y_i\in S_{j^*}(X_i)\}\geq \mathbb{P}\left(Y_{2n+1}\in S_{j^*}(X_{2n+1})\right)-\delta,$$
with probability at least $1-\delta$.
Combining the two inequalities above and Lemma \ref{lem:nest}, we get
$$\frac{1}{n}\sum_{i=n+1}^{2n}\mathbb{I}\{q(Y_i,X_i)\geq m-j^*+1\}\geq 1-\alpha+\frac{1}{n},$$
with probability at least $1-\delta$.
This immediately implies $q_{\lfloor n\alpha\rfloor}=q_{\lfloor n(\alpha-n^{-1})+1\rfloor}\geq m-j^*+1$ by the definition of order statistics. Therefore, the volume of the prediction set $\widehat{C}_{\rm DCP-DP}(X_{2n+1})$ is at most
\begin{align*}
&\vol\left(\left\{y\in\mathbb{R}: q(y,X_{2n+1})\geq q_{\lfloor(n+1)\alpha\rfloor}\right\}\right) \\
\leq& \vol\left(\left\{y\in\mathbb{R}: q(y,X_{2n+1})\geq q_{\lfloor n\alpha\rfloor}\right\}\right) \\
\leq& \vol\left(\left\{y\in\mathbb{R}: q(y,X_{2n+1})\geq m-j^*+1\right\}\right) \\
=& \vol(S_{j^*}(X_{2n+1})),
\end{align*}
where the last identiy is by Lemma \ref{lem:nest}.
The volume of $S_{j^*}(X_{2n+1})$ can be controlled by Assumption \ref{as:ne-su},
\begin{align*}
    \vol(S_{j^*}(X_{2n+1})) \leq& \opt_k\left(\widehat{F}(\cdot\mid X_{2n+1}),1-\alpha+\frac{1}{n}+3\delta+\gamma\right) \\
\leq& \opt_k\left(F(\cdot\mid X_{2n+1}),1-\alpha+\frac{1}{n}+4\delta+\gamma\right),
\end{align*}
where the last inequality, which holds with probability at least $1-\delta$, is by Assumption~\ref{ass:F}. Combining the inequalities above with union bound, we get the conclusion.
\end{proof}

\subsection{Proofs of Proposition \ref{thm:DP}, Lemma \ref{lem:gmmode}, Lemma \ref{lem:tv} and Lemma \ref{lem:nest}} \label{sec:proof_of_lemmas}

\textbf{Dynamic Programming Algorithm.} 
The dynamic programming table $DP(i,j,l)$ stores the minimum volume of $i$ intervals that collectively cover $l \gamma n$ points from the sorted training data $Y_{(1)}, \dots, Y_{(j)}$, where the right endpoint of the rightmost interval is fixed at $Y_{(j)}$. Here, $Y_{(1)}, \dots, Y_{(n)}$ are the training data points $Y_1,\dots, Y_n$ sorted in non-decreasing order.
For each state in the DP table, we iterate over all possible left endpoints of the rightmost interval, as well as the right endpoint of the preceding interval (if it exists). This allows us to systematically compute the optimal solution for each state by the following formula:
\begin{align*}
&\text{If $i = 1$,} \quad DP(i,j,l) = Y_{(j)} - Y_{(j-\lceil l\gamma n \rceil +1)},\\
&\text{If $i > 1$,} \quad DP(i,j,l) = \min_{i-1\leq j''< j' \leq j}\{Y_{(j)} - Y_{(j')} + DP(i-1, j'', l')\},
\end{align*}
where $l' = l - \lfloor (j - j' + 1)/(\gamma n) \rfloor$.
Finally, we find the minimum volume solution among all entries $DP(k,j,\lceil(1-\alpha)/ \gamma\rceil)$ for all $1\leq j\leq n$.
Then, we use the standard backtrack approach on the DP table to find the prediction set $\widehat C_{DP}$.

\begin{proof}[Proof of Proposition \ref{thm:DP}.]
    Without loss of generality, we assume that $1/\gamma$ is an integer, otherwise, we can decrease $\gamma$ to make this hold.
    For any $i \in [k], j \in [n], l \in [ 1/\gamma ]$, we use the dynamic programming table entry $DP(i,j,l)$ to store the minimum volume of $i$ intervals that cover $ \lceil l\cdot \gamma n \rceil$ points in $Y_{(1)},\dots, Y_{(j)}$ and the right endpoint of the rightmost interval is at $Y_{(j)}$. 
    If there is no feasible solution for this subproblem, we set $DP(i,j,l) = \infty$. 
    This dynamic programming is shown in Algorithm~\ref{alg:dp}.
    This dynamic programming runs in time $O(n^3k /\gamma)$.

    We then find the solution with the minimum volume among all subproblems $DP(k,j, \lceil (1-\alpha)/\gamma \rceil )$ for $1\leq j \leq n$. It is easy to see that there exists a feasible solution. Let $\widehat{C}_{\rm DP} \in \calC_k$ be a union of $k$ intervals in this solution. This solution covers at least $ \lceil (1-\alpha)/\gamma \cdot (n\gamma) \rceil  = \lceil (1-\alpha) n \rceil$ points in $X_1,\dots, X_n$. Thus, we have 
    \begin{align*}
        \mathbb{P}_n(\widehat{C}_{\rm DP})\geq 1-\alpha.
    \end{align*}
    If the restricted optimal volume $\opt_k(\mathbb{P}_n, ((1-\alpha)/\gamma+1) \cdot (n\gamma)/n )$ is smaller than the volume of $\widehat{C}_{\rm DP}$, then this solution cannot have the minimum volume among all subproblems $DP(k,j, \lceil (1-\alpha)/\gamma \rceil )$ for $1\leq j \leq n$. Thus, the volume of this solution must satisfy
    \begin{align*}
        \vol(\widehat{C}_{\rm DP}) \leq& \opt_k(\mathbb{P}_n, ((1-\alpha)/\gamma+1) \cdot (n\gamma) / n) \\
        =& \opt_k(\mathbb{P}_n,1-\alpha + \gamma).
    \end{align*}
    The proof is thus complete.
\end{proof}

\begin{proof}[Proof of Lemma \ref{lem:gmmode}]
By \cite{carreira2003number}, there are $k' \le k$ local maxima for the density function $p_k$. We will use $k'$ intervals and define the rest of the intervals to be empty. Suppose $u_1 \le u_2 \le  \dots \le u_{k'} \in \R$ are the local maxima of $p_k$. The density $p_k(y)$ is differentiable, and its local minima and local maxima have to alternate. Hence there are exactly $k'-1$ local minima, denoted by $\ell_1, \ell_2, \dots, \ell_{k'-1}$ with $\ell_j \in [u_{j}, u_{j+1}]$ for all $ j \in \{1,2,\dots, k'-1\}$. (Note that there are no local minima less than $u_1$ or greater than $u_{k'}$ since $p_k(y) \to 0$ as $y \to \pm \infty$). For notational convenience let $\ell_0 = -\infty, \ell_{k'}=\infty$. Let $C^* \subset \R$  satisfies $P_k(C^*)\geq 1-\alpha$ and $\vol(C^*)=\opt(P_k,1-\alpha)$.

We now show that there exist $I_1,\cdots,I_{k'}\in\calC_1$ such that $u_j\in I_j\subset [\ell_{j-1},\ell_j]$ for all $j\in[k']$, $P_k(\cup_{j=1}^{k'}I_j)\geq 1-\alpha$, and $\vol(\cup_{j=1}^{k'}I_j)\leq \vol(C^*)$. This would imply the desired conclusion. Consider $S_j = S^* \cap [\ell_{j-1}, \ell_j]$ for all $j \in [k']$. Next we observe for all $j \in [k']$, $p_k$ is monotonically increasing in the interval $[\ell_{j-1}, u_{j}]$ and is monotonically decreasing in the intervals $[u_j, \ell_j]$, with a local maximum at $u_j$. Hence if $S_j$ comprises multiple disjoint intervals within $[\ell_{j-1}, \ell_{j}]$, we can pick one interval with the same volume within $[\ell_{j-1}, \ell_j]$ that also includes $u_j$ and covers at least as much probability mass. This establishes the property of $\cup_{j=1}^{k'}I_j$, and hence the lemma.
\end{proof}

\begin{proof}[Proof of Lemma \ref{lem:tv}]
First, we construct a family of distributions supported on subsets of $[0,1]$. Consider an integer $m$. We partition the interval $[0,1]$ into $m$ intervals with length $1/m$ each and define the subinterval $A_j=\left[\frac{j-1}{m},\frac{j}{m}\right)$ for $j \in [m]$. 
We next define a family of distributions supported on these subintervals. 
For any vector $Z\in\{0,1\}^m$, let $A_Z=\bigcup_{j:Z_j=1}A_j$ denote the union of intervals corresponding to the indices where $Z_j = 1$. Then, we define the density function $$p_Z(y)=\frac{\mathbb{I}\{y\in A_Z\}}{\frac{1}{m}\sum_{j=1}^mZ_j},$$
where $\mathbb{I}$ is the indicator function.
Let $P_Z$ be the corresponding distribution.
 
We then construct the weight distribution $\Pi$. 
Given any $\varepsilon > 0$, we now provide a restricted set of vectors $Z \in \{0,1\}^m$ such that the distribution $P_Z$ has at least $1-\varepsilon$ total variation distance to the uniform distribution $\lambda$. We pick a parameter $\beta \in (0,1)$ depending on $\varepsilon$ and $m$. Then, we define a set of vectors $Z$ with $A_Z$ covering approximately $\beta$ fraction of $[0,1]$, 
$$\mathcal{Z}=\left\{Z\in\{0,1\}^m:\left|\frac{1}{m}\sum_{j=1}^mZ_j-\beta\right|\leq\left(\frac{\beta}{m}\right)^{1/3}\right\}.$$
For any $Z\in\mathcal{Z}$, we have the total variation distance 
$$\TV(P_Z,\lambda) \geq \lambda(A_Z^c)=1-\frac{1}{m}\sum_{j=1}^mZ_j\geq 1-\beta-\left(\frac{\beta}{m}\right)^{1/3}.$$ 
Therefore, as long as $\beta+\left(\frac{\beta}{m}\right)^{1/3}\leq\epsilon$, we have $P_Z\in\mathcal{P}_{\epsilon}$ for all $Z\in\mathcal{Z}$.
We construct the weight distribution $\Pi$ supported on the $\{P_Z: Z \in \calZ\}$. Let $\tilde{\Pi}=\otimes_{j=1}^m\text{Bernoulli}(\beta)$ be the product distribution on $\{0,1\}^m$ such that each coordinate is $1$ with probability $\beta$. Then, we define $\Pi$ to be the distribution $\tilde{\Pi}$ conditioning on $\mathcal{Z}$, $\Pi(Z)=\frac{\tilde{\Pi}(Z\cap\mathcal{Z})}{\tilde{\Pi}(\mathcal{Z})}$.

We bound $\TV(\lambda^n,\int P_Z^n d\Pi(Z))$ by chi-squared divergence,
\begin{align*}
    \frac{1}{2}\TV\left(\lambda^n,\int P_Z^n \mathrm{d}\Pi(Z)\right)^2 \leq \int_{[0,1]^n} \left(\int_Z p_Z^n \mathrm{d}\Pi(Z)\right)^2 \mathrm{d}x -1.
\end{align*}
Since $\int P_Z^n d\Pi(Z)$ is a mixture of product distributions over $\{P_Z^n: Z \in \calZ\}$ with weights $\Pi$, we can expand the first term in the right-hand side as
\begin{align*}
    \int_{[0,1]^n} \left(\int_Z p_Z^n \mathrm{d}\Pi(Z)\right)^2 \mathrm{d}x
= \mathbb{E}_{Z,Z'\stackrel{iid}{\sim} \Pi}\left(\int p_Z(y)p_{Z'}(y) \mathrm{d}x\right)^n.
\end{align*}
By taking the density $p_Z(y) = \frac{\mathbb{I}\{x\in A_Z\}}{\frac{1}{m}\sum_{j=1}^mZ_j}$ into the equation, we have
\begin{align*}
    \mathbb{E}_{Z,Z'\stackrel{iid}{\sim} \Pi}\left(\int p_Z(y)p_{Z'}(y) \mathrm{d}x\right)^n
= \mathbb{E}_{Z,Z'\stackrel{iid}{\sim} \Pi}\left(\frac{1}{\frac{1}{m}\sum_{j=1}^mZ_j}\cdot \frac{1}{\frac{1}{m}\sum_{j=1}^mZ'_j}\cdot \frac{1}{m}\sum_{j=1}^mZ_jZ_j'\right)^n.
\end{align*}
According to the construction of $\calZ$, for any vector $Z \in \calZ$, we have $\frac{1}{m}\sum_{j=1}^mZ_j \geq \beta - \left(\frac{\beta}{m}\right)^{1/3}$. Since $\tilde\Pi$ is the product of Bernoulli distribution with probability $\beta$, by the Chernoff bound, we have $\tilde\Pi(Z \in \calZ) \geq 1 - (\beta/m)^{1/3}$. Thus, we have
\begin{eqnarray*}
&&\mathbb{E}_{Z,Z'\stackrel{iid}{\sim} \Pi}\left(\int p_Z(y)p_{Z'}(y) \mathrm{d}x\right)^n\\
&\leq& \left(\beta-\left(\frac{\beta}{m}\right)^{1/3}\right)^{-2n}\mathbb{E}_{Z,Z'\stackrel{iid}{\sim} \Pi}\left(\frac{1}{m}\sum_{j=1}^mZ_jZ_j'\right)^n \\
&=& \left(\beta-\left(\frac{\beta}{m}\right)^{1/3}\right)^{-2n}\frac{\mathbb{E}_{Z,Z'\stackrel{iid}{\sim} \tilde{\Pi}}\left(\frac{1}{m}\sum_{j=1}^mZ_jZ_j'\right)^n\mathbb{I}\{Z\in\mathcal{Z}\}\mathbb{I}\{Z'\in\mathcal{Z}\}}{\tilde{\Pi}(Z\in \mathcal{Z})^2} \\
&\leq& \left(\beta-\left(\frac{\beta}{m}\right)^{1/3}\right)^{-2n}\left(1-\left(\frac{\beta}{m}\right)^{1/3}\right)^{-2}\mathbb{E}_{Z,Z'\stackrel{iid}{\sim} \tilde{\Pi}}\left(\frac{1}{m}\sum_{j=1}^mZ_jZ_j'\right)^n.
\end{eqnarray*}
The last term on the right-hand side can be bounded by
\begin{eqnarray*}
 \mathbb{E}_{Z,Z'\stackrel{iid}{\sim} \tilde{\Pi}}\left(\frac{1}{m}\sum_{j=1}^mZ_jZ_j'\right)^n 
&\leq& \left(\beta^2 + \left(\frac{\beta^2}{m}\right)^{1/3}\right)^n + \tilde{\Pi}\left(\frac{1}{m}\sum_{j=1}^mZ_jZ_j'>\beta^2 + \left(\frac{\beta^2}{m}\right)^{1/3}\right) \\
&\leq& \left(\beta^2 + \left(\frac{\beta^2}{m}\right)^{1/3}\right)^n + \left(\frac{\beta^2}{m}\right)^{1/3},
\end{eqnarray*}
where the first inequality is using $\frac{1}{m}\sum_{j=1}^mZ_jZ_j' \leq 1$ for $\frac{1}{m}\sum_{j=1}^mZ_jZ_j' > \beta^2 + \left(\frac{\beta^2}{m}\right)^{1/3}$ and the second inequality is from the Chernoff bound on the sum of $m$ Bernoulli variables with probability $\beta^2$.
Combining all bounds above, we have
$$\frac{1}{2}\TV\left(\lambda^n,\int P_Z^n \mathrm{d}\Pi(Z)\right)^2\leq \frac{\left(\beta^2 + \left(\frac{\beta^2}{m}\right)^{1/3}\right)^n + \left(\frac{\beta^2}{m}\right)^{1/3}}{\left(\beta-\left(\frac{\beta}{m}\right)^{1/3}\right)^{2n}\left(1-\left(\frac{\beta}{m}\right)^{1/3}\right)^{2}}-1.$$
It is clear that the above bound tends to zero when $m\rightarrow\infty$. Therefore, for any $\delta > 0$, we have
 $\TV\left(\lambda^n,\int P_Z^n \mathrm{d}\Pi(Z)\right)\leq\delta$ for a sufficiently large $m$.
\end{proof}

\begin{proof}[Proof of Lemma \ref{lem:nest}]
The result is a direct consequence of the nested property of $\{S_j(x)\}_{j\in[m]}$.
\end{proof}

\end{document}